%% file: main.tex
\newif\ifdraft
\draftfalse

\documentclass{article} 

\usepackage{iclr2027_conference,times}
\iclrpreprintcopy

\input{math_commands.tex}

\usepackage{hyperref}
\usepackage{url}
\usepackage{graphicx}
\usepackage{placeins}
\usepackage[normalem]{ulem}
\usepackage{wrapfig}
\usepackage{subcaption}
\usepackage{booktabs}

\usepackage[normalem]{ulem} 
\newcommand{\stkout}[1]{\ifmmode\text{\sout{\ensuremath{#1}}}\else\sout{#1}\fi}

\ifdraft

\newcommand{\deleted}[1]{\textcolor{red}{\stkout{#1}}}

\newcommand{\deletedfloat}[1]{}
\newcommand{\commented}[1]{\textcolor{blue}{#1}}
\else

\newcommand{\deleted}[1]{}

\newcommand{\deletedfloat}[1]{}
\newcommand{\commented}[1]{}
\fi

\title{How Far Does a Shared Linear Map Go? Probing Feature-Space Manipulability for Image Editing}

\author{Elias Krey \\
  Division AI4Health\\
  Carl von Ossietzky Universität Oldenburg\\
  Oldenburg, Germany \\
  \texttt{elias.krey@uol.de} \\
  \And
  Nils Neukirch \\
  Division AI4Health \\
  Carl von Ossietzky Universität Oldenburg\\
  Oldenburg, Germany \\
  \texttt{nils.neukirch@uol.de} \\
  \AND
  Nils Strodthoff \\
  Division AI4Health \\
  Carl von Ossietzky Universität Oldenburg\\
  Oldenburg, Germany \\
  \texttt{nils.strodthoff@uol.de} \\
}

\begin{document}

\maketitle

\begin{abstract}
Understanding how image-space transformations manifest in a model's internal feature representations is a longstanding goal in representation analysis. 
Prior work on model stitching and equivariance has shown that certain geometric transformations — rotations, flips — can be captured by learned linear operators between feature maps. It remains unclear whether this holds for a broader and more practically relevant class of transformations, including photometric edits and open-ended, semantically defined manipulations with no a priori linear structure in input space, such as those produced by prompted diffusion-based image editors.
We train probes with increasing capacity, ranging from a single linear map shared across all spatial locations in a feature map to nonlinear per-vector, receptive-field, and global transformer models, to predict the feature-space effects of diverse image manipulations: geometric transforms, photometric edits, local occlusions, and semantic edits (e.g., altering headlights, rim color, body color) generated via diffusion-based editing. We evaluate three vision backbones: two supervised architectures, ConvNeXt and SwinV2, and one self-supervised foundation model, DINOv3. For ConvNeXt and SwinV2, a single shared linear map, without spatial or instance-dependent conditioning, often predicts held-out manipulation outcomes with little loss relative to substantially more expressive nonlinear models. This pattern is less consistently observed for DINOv3. For the supervised backbones, the sufficiency of the shared linear map generally increases with network depth. We further show this holds despite the spatial weight-tying constraint, which is not implied by prior stitching formulations and represents a nontrivial locality claim independent of linearity. We deliberately scope our claims to representational sufficiency for held-out prediction, rather than to intrinsic properties of feature-space geometry. Our results indicate that a remarkably simple, shared linear operator is often sufficient to represent a broad class of image manipulations, with the semantic content of the edits captured by its leading $k$ singular components and higher-rank components primarily refining image details. 
\end{abstract}

\section{Introduction}
\label{intro}

\paragraph{Research question} This work poses a representational question framed in a predictive manner: \textit{Can the effect of an arbitrary image manipulation on a model's internal features be captured by a simple, shared function?} The answer to this question relates to many different domains: First, it connects to mechanistic interpretability, \citep{rai2024practical,bereska2024mechanistic} for exemplary review, which attempts to reveal how information is organized and processed in intermediate model representation. Resolving the question is essential for controllable, counterfactual editing directly in feature space. Second, it provides an empirical entry point into questions of representational symmetry in intermediate layers, see e.g.\ \citep{bordes2022high}, though we emphasize this work does not test equivariance to continuous transformation groups. 

\begin{figure}[t]
    \centering
    \includegraphics[width=0.7\linewidth]{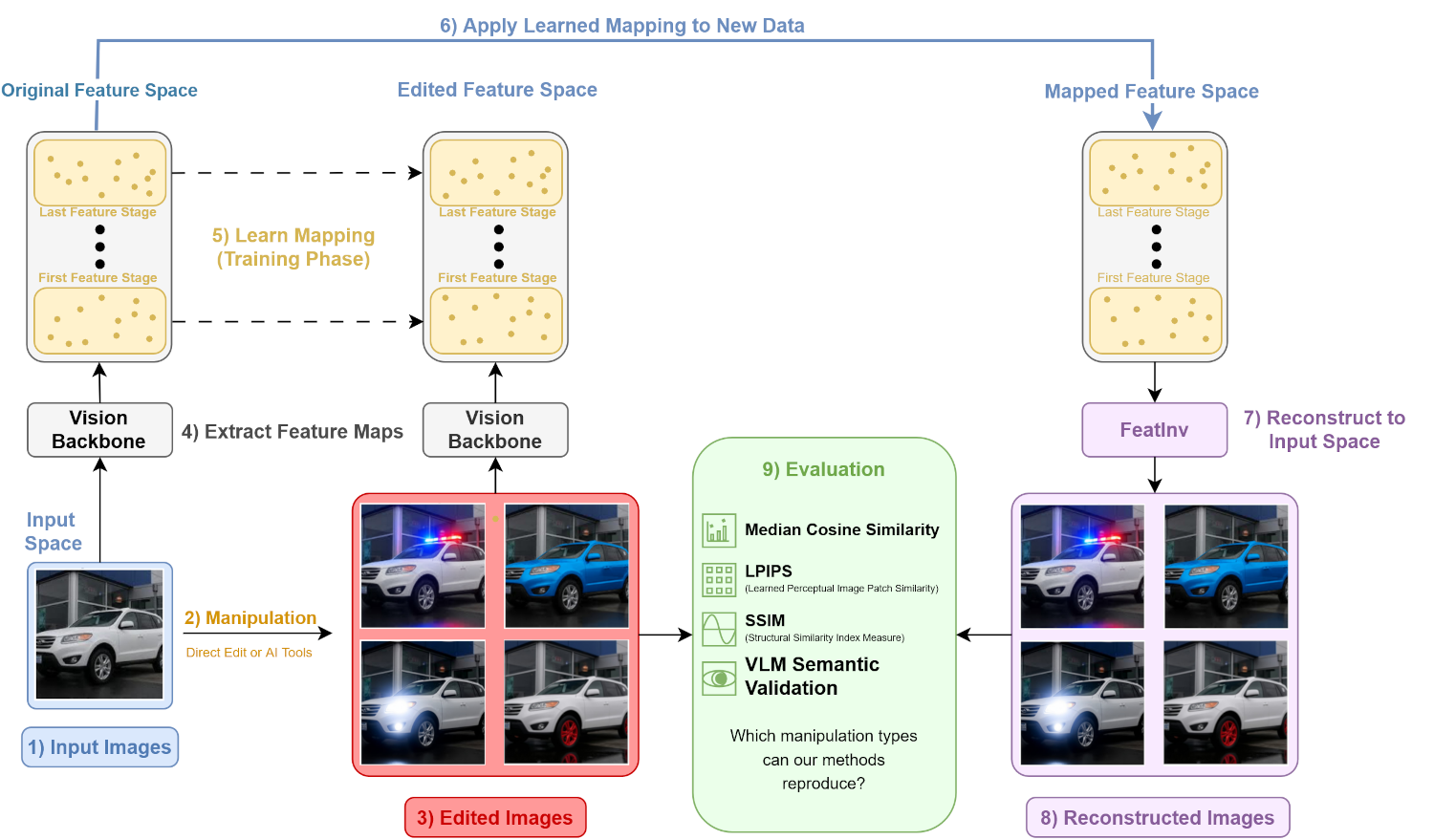}
    \caption{Featmap overview. Given an input image, we apply geometric, photometric, masking, or semantic manipulations and extract feature maps from a pretrained vision backbone (ConvNeXt, SwinV2, or DINOv3). We train models to map the original features to those of the manipulated image. We evaluate the learned mapping by reconstruction quality, using feature-space cosine similarity and SSIM/LPIPS between the manipulated image and its FeatInv \citep{neukirch2025featinv} reconstruction, and by validation through a vision-language model (VLM).}
    \label{fig:abstract}
\end{figure}

\paragraph{State of research} Variants of the above question have been investigated in the image stitching literature. \citet{lenc2019} showed for CNNs that geometric transformations (flips, scaling, rotations) can be realized as linear transformations acting on entire feature maps. Related work on model stitching and cross-representation alignment \citep{NEURIPS2021_01ded425, moschella2023relative} similarly finds that representations are related by simple (often linear) transformations.
This leaves three central gaps: (G1) Does this extend to transformations with no inherent geometric/linear structure in pixel space, such as photometric or diffusion-edit-induced semantic transformations? (G2) Is a single operator spatially shared across all locations in feature space sufficient to perform the mapping? (G3) Does the property apply to CNNs but also to modern ViT backbones (including those trained through self-supervision) alike?

\paragraph{Approach} Overcoming (G1) heavily draws on the recent advances in image editing. Diffusion-based image editing models, such as Qwen Image-Edit \citep{wu2025qwenimagetechnicalreport}, enable a generic, prompt-driven way to induce open-ended semantic transformations and thereby support the study of transformations that were not within reach of prior studies. We address (G2) by studying predictive mappings with different linearity/locality constraints, ranging from shared linear mapping operating on feature vectors to transformer models operating on the entire feature map. Finally to address (G3), we study the question for a CNN and two modern (supervised and self-supervised) ViT-based vision backbones.

\paragraph{Central finding}  Across the considered manipulations, a single shared linear map applied identically across spatial positions often suffices to predict held-out manipulation outcomes with only modest loss relative to substantially more expressive models. Predictive sufficiency generally increases with feature depth, but the strength of this trend varies across backbones. Localized object additions are most reliably predicted in the deepest layers of ConvNeXt and SwinV2, whereas DINOv3 shows greater variability across stages and samples.

\paragraph{Contributions}
We provide the following technical contributions:
(1) We devise a diverse set of geometric, photometric and semantic manipulations (details and examples in App.~\ref{sec:Details_manip}) and apply them to three datasets from different image domains \citep{KrauseStarkDengFei-Fei-3D201, yu2016lsunconstructionlargescaleimage, WahCUB_200_2011} to investigate their impact on the feature maps of ConvNeXt, SwinV2 and DINOv3 as representatives for the three predominantly used vision backbones. 
(2) We evaluate a taxonomy of mapping models, from a shared per-vector linear map to global transformers (see Fig.~\ref{fig:semantic_manipulations_main}, App.~\ref{subsec:appendix_mapped_samples}), using both reconstruction-quality metrics (Figs.~\ref{fig:boxplot_reconstruction_metrics}, App.~\ref{subsec:app_metrics}) and semantic-fidelity metrics via a frozen classifier  (see Fig.~\ref{fig:feat3_classifier_metrics}). (3) We validate our metric choices by using a VLM to assess the semantic correctness of generated samples for each manipulation (Fig.~\ref{fig:VLM_metrics}).
(4) We show that a shared linear model operating on individual feature vectors is often sufficient to achieve strong performance relative to substantially more complex models, including for non-trivial semantic manipulations. The extent of performance degradation in higher layers varies across backbones. We further characterize properties of the learned mappings.
(5) We provide hints that the sufficiency of a local linear mapping  strengthens with network depth, across CNN and ViT backbones for supervised and self-supervised pretrained models.

\section{Related work}
\label{related_work}

\paragraph{Model stitching and representational equivalence}
A parallel line of work on model stitching studies how structured transformations in input space (e.g., flips, rotations) induce approximately linear or affine transformations in feature space. \citet{lenc2015,lenc2019} introduced equivariant transformation (``stitching'') layers that learn linear maps between CNN feature maps to compensate for geometric input transformations such as flips and 90° rotations, showing that representations change in an approximately linear, predictable way. This approach has been extended to compare neural representations across architectures and training regimes via learned affine or linear stitching layers \citep{NEURIPS2021_01ded425}, and to align latent spaces of different models using closed-form affine transformations estimated from parallel anchors \citep{moschella2023relative}. Together, these studies indicate that, in several experimentally studied settings, differences between representations can be captured by simple linear or affine transformations. Our work addresses central gaps in these works: namely, it studies (G1) transformations with no a priori linear or geometric structure, (G2) spatially shared, weight-tied operators as opposed to general (potentially position-dependent) stitching layers, and (G3) generality across CNN, ViT, and self-supervised backbones.

\paragraph{Linear representation hypothesis and concept discovery} 
The linear representation hypothesis \citep{park2025the} motivates a range of concept-based XAI methods (TCAV \citep{kim2018interpretabilityfeatureattributionquantitative}, CRAFT \citep{fel2023craftconceptrecursiveactivation}, MCD \citep{Vielhaben:2022MCD}, sparse autoencoders \citep{bricken2023monosemanticity,huben2024sparse,pmlr-v267-fel25a}) as well as compositional and data-augmentation work \citep{trager2024linearspacesmeaningscompositional,wang2020implicitsemanticdataaugmentation}, and has more recently been qualified by alternative accounts based on convex prototype combinations \citep{fel2026into} or low-dimensional manifolds \citep{bhalla2026sparseautoencoderscaptureconcept,vielhabenbeyond}.
This literature is concerned with identifying static, meaningful directions within a single fixed representation. Our setting differs: we study learned mappings between an original and a manipulated representation, induced by an external image transformation, and ask whether this mapping, rather than any one concept direction, is well-approximated as linear.

\paragraph{Feature-space decoding/reconstruction}
The success of manipulations in feature space can be most reliably assessed in input space. This relies on approaches that allow to reconstruct input space representations from a given feature-space representation. Recently, conditional diffusion models have been proposed for this task using pooled \citep{bordes2022high} or unpooled \citep{neukirch2025featinv} feature maps as input, whereas \citet{allakhverdov2026feature} trained a transformer-based mapping model from scratch. In this work, we rely on FeatInv \citep{neukirch2025featinv}, which has been shown to yield robust reconstructions of high quality. As it represents a key component of our evaluation pipeline, we provide an extended robustness analysis related to random perturbations in the incoming feature map. 

\section{Methods}
\label{methods}

\paragraph{Manipulations}
We consider three categories of input manipulation: (i) geometric and photometric transformations (e.g., rotations, mirroring, noise, color shifts), (ii) local rectangular masking with a solid color, and (iii) semantic edits of dataset-specific high-level visual attributes via prompted generative editing. All transformation parameters, masking configurations, semantic prompts, and generation settings are provided in Appendix~\ref{sec:Details_manip}; semantic edits are generated using Qwen-Image-Edit-2511 \citep{wu2025qwenimagetechnicalreport}.

\paragraph{Considered mappings}
We examine mappings of increasing complexity, from local linear to non-local nonlinear. The \emph{linear} mapping applies a shared linear transform to single feature vectors as a local baseline. The \emph{mlp} adds one hidden layer to introduce non-linearity while remaining local. The \emph{cnn} operates on the full feature map but stays relatively local due to its finite receptive field. Finally, the \emph{transformer} uses four stacked transformer layers on the flattened feature map to model non-linear, non-local dependencies. Tab.~\ref{tab:mapping_models} summarizes architectures and key hyperparameters. In Appendix~\ref{subsec:Ablation_new_models}, we additionally investigate several diagnostic and closed-form baselines (additive offset, spatially non-shared linear, orthogonal Procrustes, and ridge regression) that probe the necessity of our design choices.
We train with a weighted combination of feature-space MSE and a cosine-similarity loss. For the latter, we compute cosine similarities per spatial location, take the median per sample for robustness, and average over the batch. Empirical evaluation on a data subset showed that $\lambda_\text{MSE}=0.3$ and $\lambda_\text{COS}=0.7$ yield the best trade-off between reconstruction fidelity and feature alignment. Alternative loss formulations change quantitative metrics but do not produce consistent visual or semantic gains (App.~\ref{subsec:lossfunction_choice}), so we retain the weighted MSE--cosine objective for all main experiments. 

\begin{table}[ht]
\centering
\caption{Mapping architectures, implementation identifiers, and key hyperparameters. Input/output dimensions match the channel dimension (128,128), ..., (1024,1024) of the feature layer under consideration in both backbone models.}
\label{tab:mapping_models}

\begin{tabular}{l l l}
\toprule
model & Key hyperparameters \\
\midrule
 \emph{linear}  & Optional spatial transform (rotation/mirror) \\
 \emph{mlp}         & 1 hidden layer (512), ReLU, dropout=0.2 \\
 \emph{cnn}         & 2 Conv layers (3×3), hidden=512, BatchNorm, ReLU, Dropout2d=0.2 \\
 \emph{transformer} & 4 layers, 8 heads, hidden=512, input/output projections, dropout=0.1 \\
\bottomrule
\end{tabular}
\end{table}

\paragraph{Transformations modifying the global composition} 
For geometric transformations that alter the global image composition, such as mirroring and rotation, we reorder feature vectors to match the transformed spatial arrangement. This reordering is essential: without it, the relevant input feature vector is not available to the mapping. Once spatial alignment is established, the task reduces to learning the local feature transformation at each corresponding position. The procedure is illustrated in App. Fig.~\ref{fig:rot_mirror}.

\paragraph{Metrics}
We evaluate mapping quality through feature reconstruction, visual similarity, and semantic fidelity. Feature reconstruction is measured using the median cosine similarity (MdnCS) between original and mapped feature vectors; for localized manipulations, binary masks restrict this computation to the affected regions. Visual quality is assessed using LPIPS and SSIM \citep{DBLP:journals/corr/abs-1801-03924,1284395}. As these metrics may not capture semantic errors in open-ended tasks, we additionally use Qwen2.5-VL-7B-Instruct with a tailored prompt to compare original and target samples and identify editing failures. Mathematical definitions, prompts, and implementation details are provided in App.~\ref{sec:Appendix_metrics} and \ref{app:vlm-eval}.

\paragraph{Datasets}
We evaluate mappings across fine-grained object and scene-level datasets: \emph{Stanford Cars}~\citep{KrauseStarkDengFei-Fei-3D201},\emph{CUB-200-2011}~\citep{WahCUB_200_2011}, and \emph{LSUN Bedroom}~\citep{yu2016lsunconstructionlargescaleimage}. Stanford Cars contains 16,185 images from 196 car classes, while CUB-200-2011 contains 11,788 images from 200 bird subspecies. LSUN Bedroom complements these object-centric benchmarks with scene-level images exhibiting greater variation in viewpoint, layout, illumination, and background clutter. This selection evaluates mappings under both fine-grained semantic variation and
less constrained spatial and photometric conditions.

\paragraph{Backbones}
We use \emph{ConvNeXt} \citep{liu2022convnet} and the \emph{Swin Transformer V2} \citep{liu2022swin} as backbone models, initializing them with their official pretrained ImageNet weights, and additionally include \emph{DINOv3} \citep{siméoni2025dinov3} as a self-supervised backbone. These three architectures cover two prominent and complementary design paradigms, convolutional and transformer-based, as well as two training regimes: supervised classification (ConvNeXt, Swin V2) and modern self-supervised pretraining on large-scale unlabeled data (DINOv3). All models are strong, general-purpose backbones that have demonstrated state-of-the-art performance on a wide range of vision tasks \citep{goldblum2023battle}. Including DINOv3 allows us to test whether the feasibility and characteristics of feature-space mappings extend beyond supervised, classification-oriented backbones to self-supervised models.

\paragraph{Leveraging features extracted at different layers}
For ConvNeXt, we extract representations at all four stages before the final pooling layer to assess how transformations affect different depths of the network. For SwinV2 and DINOv3, we similarly use the four public stages defined in Section~\ref{sec:feature_extraction}. Feature maps range from high-resolution, low-dimensional representations in early layers to low-resolution, high-dimensional representations in deeper layers (see Tab.~\ref{tab:convnext_features} and Section~\ref{sec:feature_extraction} for details). We use normalized feature vectors as input for earlier layers and unnormalized vectors for the deepest layer, where normalization significantly degraded reconstruction quality (App.~\ref{subsec:effect_norm}).

\section{Results}
\label{results}

\paragraph{Mapped samples}
\label{sec:mapped_samples}

\begin{figure}[ht]
    \centering
    \includegraphics[width=1\linewidth]{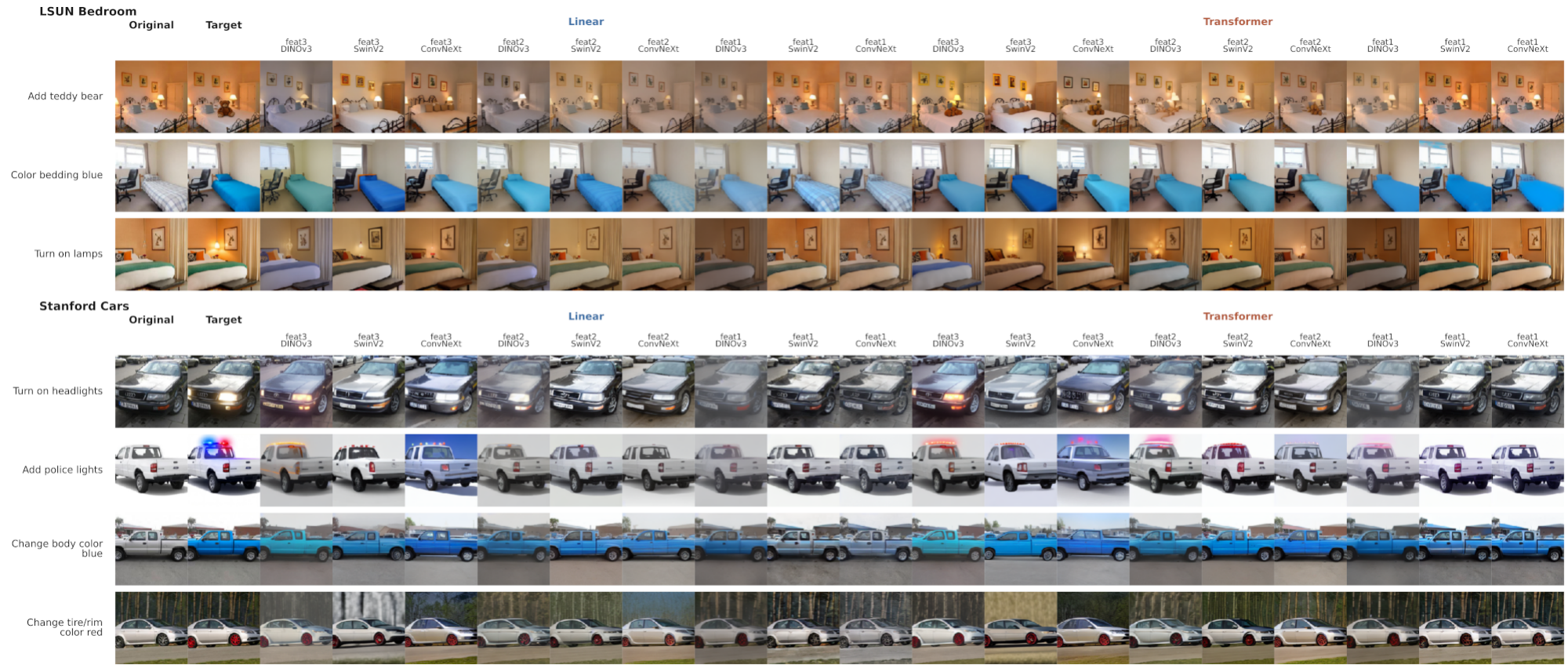}
    \caption{Examples of semantic manipulations on LSUN Bedroom and Stanford Cars. Each row shows the original, target, and mapped reconstructions obtained with linear and transformer mappings from three feature stages of DINOv3, SwinV2, and ConvNeXt.}
    \label{fig:semantic_manipulations_main}
\end{figure}

We evaluate the semantic fidelity and visual quality of images generated by our mapping models across a wide range of manipulations, backbones, and feature stages.
Fig.~\ref{fig:semantic_manipulations_main} shows representative mapped and reconstructed samples for selected semantic manipulations on the LSUN Bedroom and Stanford Cars datasets. We present results for both the linear and transformer mapping models across all three feature stages (\emph{feat1}, \emph{feat2}, \emph{feat3}) and for each backbone (DINOv3, SwinV2, ConvNeXt). Additional examples covering more mapping models, manipulations, and backbones are provided in App.~\ref{subsec:appendix_mapped_samples}. 

Across most settings, the mappings successfully realize the intended transformation. Even the simple linear model yields recognizable manipulations that reflect the target semantic direction. However, both image quality and the correctness of the applied manipulation vary with the choice of backbone and  feature stage. Global photometric manipulations (\emph{Grayscale}, \emph{Hue Shift}) are reliably achieved by most models across all layers. Semantic color edits that target specific object parts (\emph{Body color blue}, \emph{Tire rim red}) are also among the most robust, despite the need to localize the correct region. Lighting-related semantic manipulations (\emph{Turn on lamp}, \emph{Turn on headlights}) are generally feasible in later feature stages for all backbones, but are less reliable at \emph{feat1}, where the effect is often inconsistent.

The most challenging manipulations are those that require adding objects at specific locations or applying precise masks (\emph{Add teddybear}, \emph{Police lights}, and masking). In these cases, CNN and transformer models often produce more successful samples than local mappings, likely because their spatial receptive fields allow them to exploit surrounding context. For local object additions, we observe reliable successes primarily in the deepest feature layers of ConvNeXt and SwinV2. Mappings based on the DINOv3 backbone exhibit higher variability: while some samples show plausible edits, others suffer from noticeable quality degradation, including overall dimming and unintended color shifts on objects. Experiments on the CUB-200-2011 dataset (Figs.~\ref{fig:lin_tf_map_samples_birds} and~\ref{fig:geom_samples_birds}) reveal a consistent pattern: photometric and geometric manipulations are highly reliable, whereas masking and semantic manipulations are feasible but more sensitive to image content and scene conditions.

We also confirmed that learned mappings can be applied sequentially to compose multiple transformations; see App.~Figs.~\ref{fig:manip_compositions} and~\ref{fig:lsun_manip_compositions}.

\begin{figure}[ht]
    \centering
    \includegraphics[width=\linewidth]{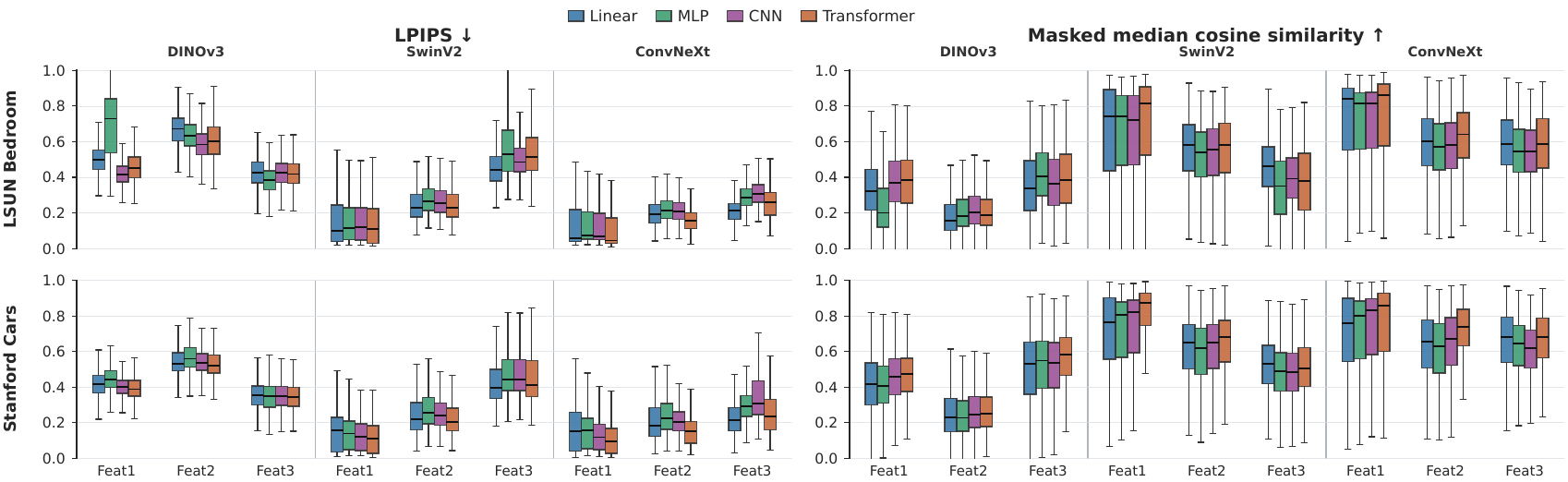}
    \caption{Reconstruction quality across feature layers, backbones, and mapping methods on LSUN Bedroom and Stanford Cars. Boxplots show per-sample LPIPS and masked median cosine similarity; colors indicate the mapping method. Lower LPIPS and higher cosine similarity indicate better alignment.}
    \label{fig:boxplot_reconstruction_metrics} 
\end{figure}

\paragraph{Post-reconstruction mapping quality}
Fig.~\ref{fig:boxplot_reconstruction_metrics} summarizes reconstruction metrics across all manipulation types for the LSUN Bedroom and Stanford Cars datasets. Overall, model performances within a given backbone and feature-depth combination are similar. The transformer-based model yields the best results across all metrics and feature stages. Among backbones, ConvNeXt consistently outperforms the others on both metrics, whereas DinoV3 exhibits lower performance at the \emph{feat2} stage. Although dataset-specific differences exist, the main trends with respect to backbone choice and feature depth are consistent across both datasets.

A more fine-grained breakdown by manipulation type is provided in App.~\ref{subsec:app_metrics}. Across all backbones and mapping models, the \emph{masking} manipulation group is the most challenging. For ConvNeXt, masking-related metrics improve monotonically from \emph{feat1} to \emph{feat3}, consistent with the qualitative examples in Fig.~\ref{fig:mapping_samples_convnext_photometric_masking}, where successful masking is only evident in the deepest layer. In contrast, photometric and geometric manipulations are reliably reconstructed at all feature depths for SwinV2 and ConvNeXt, as reflected by generally high metrics that show the largest gains in the earliest layers. DinoV3 again displays mixed behavior in the heatmap analysis, with a pronounced performance drop at \emph{feat2}. We also exemplarily confirmed the semantic fidelity of the mapping by investigating its effect on the output distribution of a finetuned classifier, see App.~\ref{sec:semantic}. 

We further assessed how faithfully the learned mappings reproduce manipulation-induced shifts in feature space (App.~\ref{sec:geom_accuracy}). Directional alignment is high for global transformations but much lower for localized edits. Although the transformer improves magnitude preservation over the linear model, both mappings exhibit similar directional patterns and compress the magnitude scale, indicating that manipulation geometry is primarily shaped by the underlying feature representation rather than the choice of mapping.

\begin{figure}[ht]
    \centering
    \includegraphics[width=0.8\linewidth]{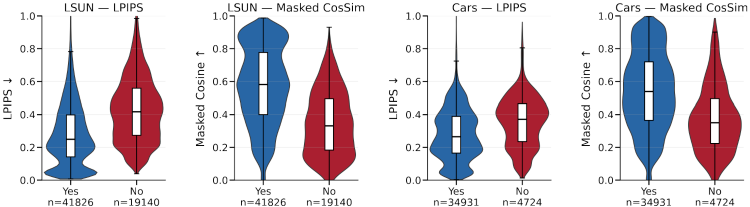}
    \caption{Metric distributions by dataset and VLM verdict. Violin plots show the distributions of LPIPS and masked MdnCS for samples classified as Yes or No by the VLM evaluator on LSUN Bedroom and Cars. }
    \label{fig:VLM_metrics}
\end{figure}

\paragraph{VLM semantic validation}
Fig.~\ref{fig:VLM_metrics} shows the metric distributions for samples assessed as (un)successfully mapped by the VLM. Although the groups overlap for both metrics, successful manipulations tend to have lower LPIPS and higher masked MdnCS, with the clearest separation observed for masked MdnCS, which evaluates the manipulated region rather than the full image. We quantified this association using within-configuration Cliff's $\delta$, stratified by backbone, mapping method, feature stage, and manipulation group. The median $\delta$ was higher for masked MdnCS than for LPIPS: $0.221$ versus $0.081$ on LSUN Bedroom and $0.131$ versus $0.072$ on Cars. VLM-Yes samples had higher masked MdnCS in $78.5\%$ and $75.9\%$ of eligible configuration cells, respectively, supporting its closer association with VLM-assessed manipulation success.

\paragraph{Bias analysis} To gain further insights into the linear mapping, we compare the weighted input $Wx$ with the bias term $b$. The input dominance ratio remains above $0.99$, while the MdnCS between $Wx$ and $Wx+b$ exceeds $0.9992$, indicating that the bias has negligible influence on the output magnitude and direction (see Sec.~\ref{sec:bias}). Consequently, these metrics show no clear separation between manipulation groups. Despite its limited effect on the output, the bias norm varies across architectures, with SwinV2 and ConvNeXt exhibiting values between $0.6$ and $1.4$ across manipulation groups. Thus, the architectures use the bias differently at the parameter level, even though the output remains dominated by $Wx$. 

\paragraph{SVD analysis}
\label{par:svd_analysis}
We analyze the linear mappings via the SVD of their weight matrices for the last feature layer before the classifier. We focus on the ConvNeXt backbone and provide the SwinV2 results in App.~Fig.~\ref{fig:full_svd}. ConvNeXt mappings exhibit higher peak singular values at early ranks than SwinV2. Across all models and backbones, spectral entropies exceed 5.4, indicating that singular values are spread over many dimensions. For ConvNeXt feat3 (1024-dimensional mapping), the maximum entropy is $\ln(1024) \approx 6.93$. Observed values between 5.4 and 6.5 imply non-uniform singular values and thus anisotropic scaling rather than pure rotations. The \emph{Mirror h} manipulation consistently shows the highest entropy, while other geometric transformations remain below 5.8. Later stages (feat2/feat3) exhibit broader spreads across manipulation groups, mirroring the reconstruction-quality gains for linear models in Figs.~\ref{fig:semantic_manipulations_main}, \ref{fig:truncated_svd_all_feats_manip_grayscale}, and \ref{fig:truncated_svd_all_feats_manip_Turn_on_headlights}. SwinV2 shows a similar pattern, with a few large early singular values followed by rapid decay (App.~Fig.~\ref{fig:full_svd}).

\paragraph{Image reconstruction with truncated SVD vs. full linear models}
\label{par:svd_trunc}
Fig.~\ref{fig:svd_vs_full} shows how LPIPS and masked MdnCS change for SVD mappings using only the top-$k$ singular values, relative to the full linear model for the \texttt{feat3} ConvNeXt features. LPIPS improves monotonically with increasing rank, indicating that reconstruction quality does not deteriorate when the full model is used.
The masked MdnCS shows the same trend in the median. Nonetheless, for a small number of instances, a lower-rank mapping yields a higher masked MdnCS than the full model.

\begin{figure}[!htbp]
    \centering

    \begin{subfigure}[t]{0.40\linewidth}
        \centering
        \includegraphics[
            width=\linewidth
        ]{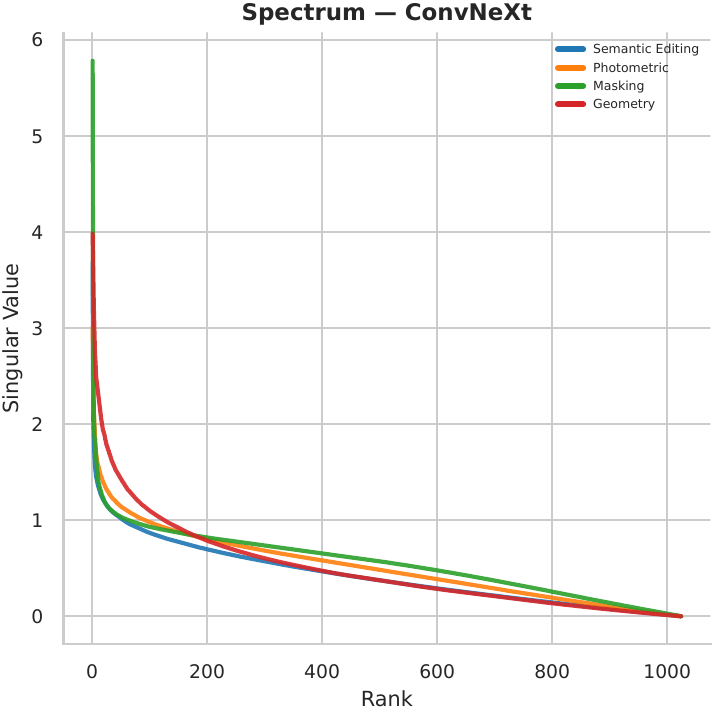}
    \end{subfigure}
    \hfill
    \begin{subfigure}[t]{0.46\linewidth}
        \centering
        \includegraphics[
            width=\linewidth
        ]{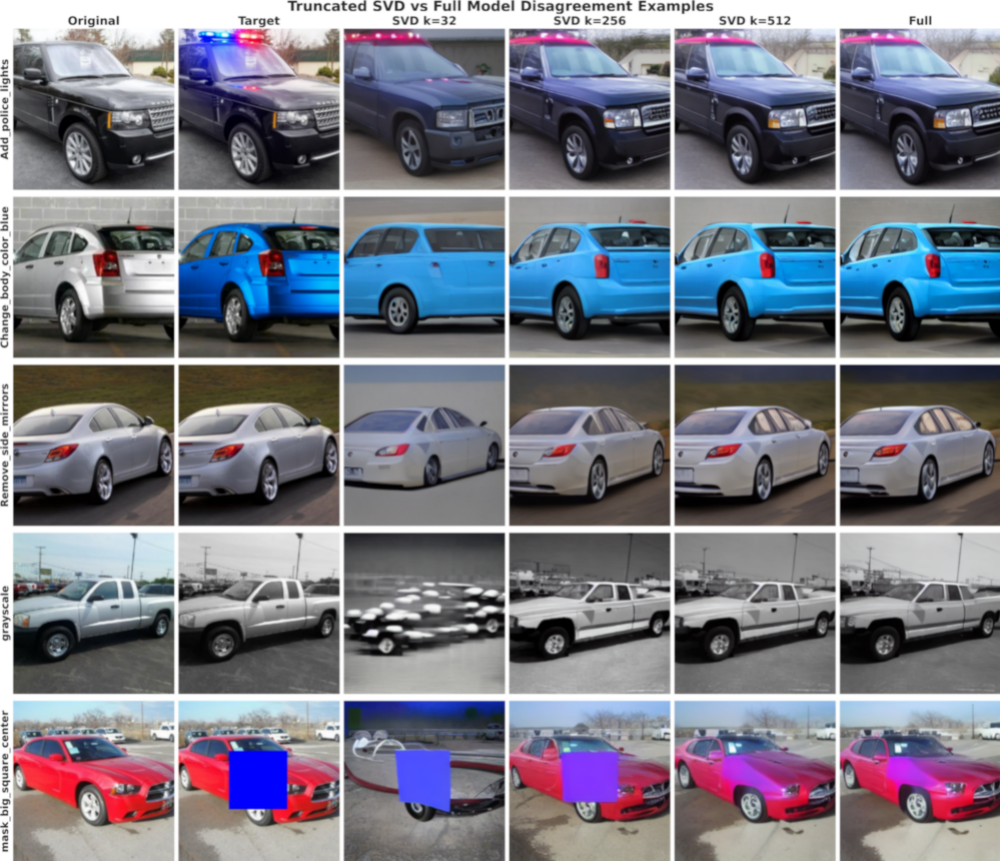}
    \end{subfigure}

    \caption{
        Truncated-SVD analysis using ConvNeXt \texttt{feat3} features.
        \textbf{Left:} Singular-value spectra and their decay as a
        function of rank across manipulations.
        \textbf{Right:} Reconstructed image examples produced by
        truncated-SVD and full-model mappings.
    }
    \label{fig:convnext_svd_analysis}
\end{figure}

Consistent with these trends, Fig.~\ref{fig:convnext_svd_analysis} shows image examples for ConvNeXt \emph{feat3} features. Overall fine-grained details of the car and background improve as the SVD rank approaches the full model, but the semantic intent is already visible at the lowest tested truncation levels. For localized edits such as masking, police lights, and mirror removal, the manipulation is often clearly present at low rank. In a subset of samples, the edit is even more pronounced at low rank and becomes partially attenuated as $k$ increases, so that the full model produces a weaker visual edit. This suggests that the dominant singular directions encode the semantic transformation, whereas higher-rank components mainly add finer image details and can dilute the visual strength of some edits. Image quality changes only marginally between the highest truncation levels (e.g., $k=512$ or $k=768$) and the full model.

For several manipulations, a similar effect appears for features from earlier ConvNeXt layers. App. Figs.~\ref{fig:truncated_svd_all_feats_manip_grayscale} and~\ref{fig:truncated_svd_all_feats_manip_Turn_on_headlights} show mapping examples for all feature depths, with truncation ranks as fixed fractions of each depth’s maximum rank. Although image quality degrades for the earliest ConvNeXt layer, the \emph{Grayscale} manipulation is already apparent at low truncation levels, e.g., through removal of the blue sky. In contrast, the \emph{Turn on headlights} manipulation succeeds only for the last two feature depths. Where it succeeds, the semantic content of the edit is already present at the lowest SVD ranks, although car and background details remain less developed. For all other manipulations, this behavior occurs only for the last ConvNeXt layer.

\section{Discussion}
\label{discussion}

\paragraph{Local mapping models}
The strong performance of local mappings indicates that many manipulations can be predicted without explicit spatial interaction. In particular, a shared linear operator acting independently on each feature vector often performs competitively with substantially more expressive models. This suggests that the relevant information is frequently available in local feature vectors, although it does not imply that all manipulations are purely local or that spatial context is unnecessary. The distinction between local and global mappings becomes most relevant for manipulations requiring precise localization or object insertion. For these cases, the transformer mapping often produces more successful samples, consistent with a benefit from global context. Thus, local linearity should be understood as a frequent sufficiency condition, rather than a universal property of all manipulations and representations.

\paragraph{Linear mapping models}
Beyond the distinction between local and global mappings, the results show that the local linear operator is often sufficient even for semantic transformations. This demonstrates practical predictive sufficiency for the manipulations and feature distributions considered here, but does not establish that the underlying representations are intrinsically linear. The SVD and bias analyses provide further insight into the learned operators. Their outputs are dominated by the weight contribution rather than the bias, and their singular values deviate substantially from one, indicating anisotropic scaling and feature mixing rather than pure rotations or translations. Consistent with this observation, the additive-only baseline does not provide a competitive mapping. The low-rank SVD results further suggest that a small number of operator directions often account for the dominant visible edit, while higher-rank components refine visual details. These findings are compatible with linear or affine subspace structure, but do not distinguish it from more complex nonlinear manifolds that are well approximated by linear subspaces in the regions sampled by our experiments.

\paragraph{Effect of truncated SVD in image space}
The truncated-SVD analyses show that the main visible edit is often recovered from a small number of singular components, while higher-rank components progressively refine background and fine-grained details. For localized edits, truncation can even improve visual results, suggesting that some higher-rank components reduce feature-space error while weakening or interfering with the intended manipulation. Thus, the singular spectrum provides a useful decomposition of the learned operator's observable effects: leading components contribute strongly to the primary edit, whereas later components tend to refine its context and appearance. This interpretation is based on image-space reconstructions and does not imply that individual singular directions correspond to independent semantic concepts.

\paragraph{Layer-dependence}
The relative performance of linear and nonlinear mappings changes with feature depth, particularly for ConvNeXt and SwinV2. Qualitative inspection and VLM-based semantic evaluation indicate that successful semantic manipulations occur more frequently in later stages, with some edits emerging reliably only in the deepest layers. The reconstruction metrics show a less consistent depth trend, however, and absolute values are not directly comparable across layers because reconstruction quality and sensitivity vary intrinsically with feature resolution and depth. We therefore interpret the results as evidence that deeper features often provide more suitable representations for semantic editing, while the extent to which local linear mappings become more sufficient with depth is best assessed through within-layer comparisons to nonlinear mappings.

\paragraph{Architectural generality}
The comparison of ConvNeXt, SwinV2, and DINOv3 shows that the main qualitative findings generalize across architectures, although DINOv3 exhibits greater variability. ConvNeXt and SwinV2 generally support reliable local linear mappings for all manipulations at their deepest feature stage, whereas DINOv3 is less consistent across stages, with pronounced degradation at \emph{feat2}. Nevertheless, DINOv3 also enables plausible mappings in deeper features for many manipulations. Overall, these results suggest that the predictive sufficiency of simple mappings is not specific to convolutional or attention-based inductive biases, but reflects a general property of learned feature spaces, modulated by backbone and stage. Across datasets, successes and failures are largely consistent for a given backbone and stage, whereas changing the representation produces larger differences. This indicates that semantic editing performance is driven primarily by backbone features and depth, with smaller effects from mapping architecture and dataset in our setting. An important exception concerns manipulations requiring precise localization or object insertion (e.g., masking, object addition), where global models often outperform local mappings by exploiting spatial context.

\paragraph{Limitations}
Although the reconstruction metrics and VLM evaluation provide useful indicators of semantic correctness, the metric distributions still overlap and the semantic quality of some samples remains ambiguous. Exhaustive human evaluation would be preferable but is not feasible at the scale of our study. In addition, we investigate three representative vision backbones from two architecture families: ConvNeXt as a CNN, and SwinV2 and DINOv3 as ViTs. While these cover major contemporary design paradigms, future work could extend this analysis to additional architectures and further disentangle architectural effects from those of training strategy. Finally, the individual contributions of the mapping, the FeatInv model, and their interaction remain unclear. Although matched-magnitude perturbation experiments show that learned mapping directions generally yield more robust reconstructions than random directions, see App.~\ref{app:featinv_robustness}, the extent to which this advantage is attributable to the mapping itself or to the FeatInv reconstruction model remains unclear.

\section{Conclusion}
We address three research gaps in feature-space manipulation. First, shared linear mappings extend beyond geometric transformations to many photometric, masking, and semantic edits (G1). Second, a spatially shared operator is often sufficient, although global mappings improve manipulations requiring precise localization or object insertion (G2). Third, these findings extend across the tested CNN and ViT backbones, including the self-supervised DINOv3 model, while remaining dependent on the backbone and feature stage (G3). These results establish predictive sufficiency under the tested conditions, rather than intrinsic linearity of the representations. More broadly, our work demonstrates how semantic image-editing models can be used to probe feature representations beyond conventional geometric transformations.

\clearpage
\bibliography{iclr2027_conference}
\bibliographystyle{iclr2027_conference}

\clearpage

\appendix
\section{Appendix}

\subsection{Details on manipulations}
\label{sec:Details_manip}

\begin{figure}[ht]
    \centering
    \includegraphics[width=0.8\linewidth]{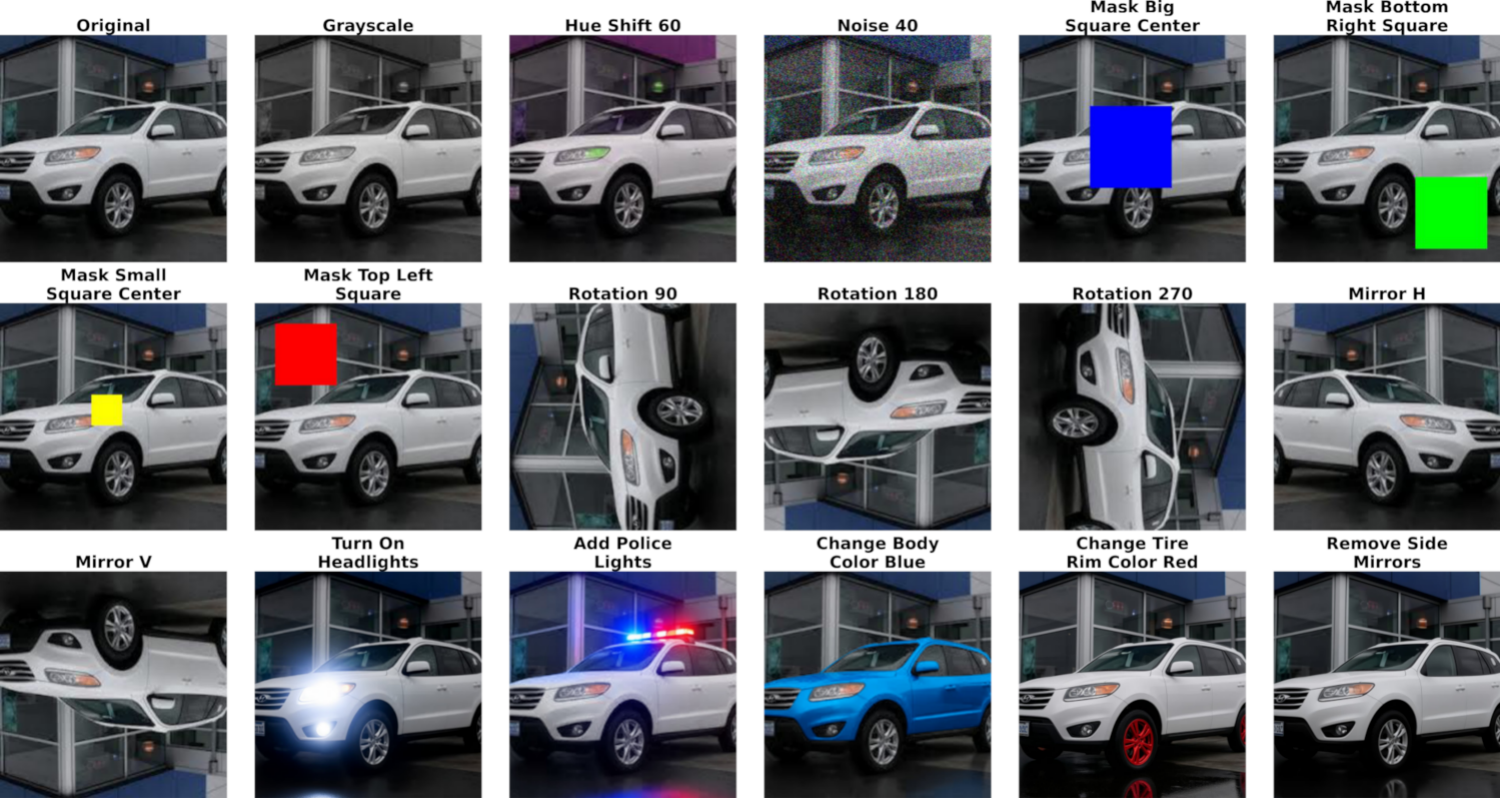}
    \caption{Examples of the manipulations applied to the dataset}
    \label{fig:example_manips}
\end{figure}

Fig.~\ref{fig:example_manips} shows examples of the manipulations, we use in our experiments, applied to the dataset. The following subsections describe each manipulation in more detail.

\subsubsection{Geometric \& Photometric Manipulations}
  \begin{itemize}
  \item Rotation (angles: $90^\circ$, $180^\circ$, $270^\circ$)
  \item Mirroring (horizontal, vertical)
  \item Added Gaussian noise (std deviation: 40)
  \item Color adjustments (hue shift: $+60^\circ$; grayscale conversion)
  \end{itemize}

\subsubsection{Local Masking Manipulations}
  \begin{itemize}
  \item Top-left square (polygon: $[(26,26),(103,26),(103,103),(26,103)]$, fill: red $[255,0,0]$)
  \item Bottom-right square (polygon: $[(180,180),(270,180),(270,270),(180,270)]$, fill: green $[0,255,0]$)
  \item Large center square (polygon: $[(90,90),(193,90),(193,193),(90,193)]$, fill: blue $[0,0,255]$)
  \item Small center square (polygon: $[(116,116),(154,116),(154,154),(116,154)]$, fill: yellow $[255,255,0]$)
  \end{itemize}

\subsubsection{Semantic Manipulations via prompted image editing} 
Using Qwen/Qwen-Image-Edit-2511 
\label{sec:qwen_details}
\citep{wu2025qwenimagetechnicalreport}, 10 inference steps, true\_cfg=4.5, seed=42
  \begin{itemize}
  \item \textbf{Stanford Cars:}
  \begin{itemize}
       \item Color change: ``Change the car's body color to bright blue, while maintaining the exact same lighting, reflections, background, and all other details.''
      \item Mirror removal: ``Remove the car's side mirrors in a realistic photo style, while keeping the car's body, lighting, reflections, and background identical.''
      \item Headlights on: ``Turn on the car's headlights so they appear bright and illuminated, while preserving the car's position, lighting, color, and all background details exactly as in the original image.''
      \item Police lights: ``Add flashing blue and red police lights to the top of the car, making them appear illuminated, while keeping the car's color, lighting, reflections, position, and all background details exactly the same.''
      \item Rim color change: ``Change the car's tire rim color to red in a realistic photo style, while keeping the car's body, lighting, reflections, and background identical.''
  \end{itemize}
  \item \textbf{CUB-200-2011:}
  \begin{itemize}
    \item Change plumage color blue: ``Change only the birds plumage color to bright blue, without altering the birds species, pose, head orientation, body shape, texture, or expression. Keep the original background, lighting, shadows, camera angle, and image composition exactly the same.''
    \item Open beak: ``Edit the bird to show its beak open in a calling or singing posture, while preserving the exact species, head direction, body pose, and limb positions. Do not change the background, lighting, camera angle, or any other element except the mouth/beak opening.''
    \item Preening wing: ``Show the bird gently preening one wing, in the same pose and background, without changing species or plumage.''
  \end{itemize} 
  \item \textbf{LSUN Bedroom:}
  \begin{itemize}
    \item Add teddybear: ``Add a single brown teddy bear positioned in the middle of the bed. Change only the addition of the teddy bear. Do not alter the bed, bedding, pillows, furniture, walls, floor, ceiling, or any existing objects. Preserve their original colors, textures, materials, positions, lighting, shadows, reflections, room layout, and perspective exactly.''
    \item Make bed unmade: ``Make the visible bed appear naturally unmade and recently used by slightly disturbing the bedding, sheets, blankets, and pillows. Modify only the bedding on the bed. Do not add or remove any objects, and do not change the bed frame, furniture, walls, floor, ceiling, decorations, or any other part of the room. Preserve the original colors, materials, lighting, shadows, room layout, perspective, and composition.''
    \item Color bedding blue: ``Recolor the visible bedding bright blue. Change only the sheets, blankets, and pillows on the bed. Keep the bed frame and all other objects, furniture, walls, floor, ceiling, and decorations unchanged. Preserve their original colors, textures, materials, positions, lighting, shadows, room layout, and perspective.''
    \item Turn on lamps: ``Turn on all visible lamps in the room. Change only the lamp states and the resulting illumination. Keep the lamps, furniture, walls, floor, ceiling, decorations, and all other objects in their original positions, colors, shapes, and materials. Do not add, remove, or modify any objects. Preserve the original room layout, perspective, and composition.''
  \end{itemize}
  \end{itemize}

\subsection{Note on the datasets}
\label{sec:datasets}
In our experiments, we use three image datasets: the Stanford Cars dataset
\citep{KrauseStarkDengFei-Fei-3D201}, the CUB\_200\_2011 dataset
\citep{WahCUB_200_2011}, and the bedroom category of the LSUN dataset \citep{yu2016lsunconstructionlargescaleimage}. These datasets cover a range of visual domains,
including cars, bird species, and indoor bedroom scenes, respectively. Neither dataset is distributed under a clearly stated open license in the original source. The dataset authors indicate that use is permitted with appropriate attribution, and the CUB\_200\_2011 dataset is additionally restricted to non-commercial research and educational use. Stanford Cars is no longer available from its original source but is distributed through third-party implementations such as torchvision.

\subsection{Reorderings for mirroring and rotation}
\label{sec:rot_mirror}
\begin{figure}[htbp]
    \centering
    \includegraphics[width=0.8\linewidth]{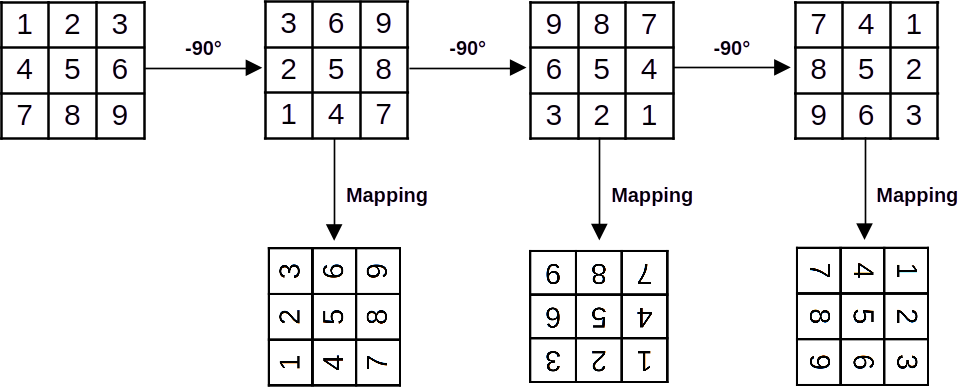}
    \caption{Example how geometric transformations are realized by first reordering feature-vector positions and then applying the local mappings.}
    \label{fig:rot_mirror}
\end{figure}

In Fig.~\ref{fig:rot_mirror}, we illustrate the methodology of reordering feature vectors for the case of rotation/mirroring, which modify the global composition of the image. We reorder feature‑vector positions to enable local mappings for rotation and mirroring. First, we permute the spatial locations of the feature vectors according to the chosen geometric transformation, without modifying the feature values themselves. Then, a mapping model is applied to each feature vector individually to learn the local effect of the same transformation on the features.

\subsection{Feature extraction at different stages}
\label{sec:feature_extraction}
We extract features from three pretrained backbones: ConvNeXt, SwinV2, and DINOv3. For ConvNeXt and SwinV2, we use the three stages before the final pooling layer, following the same procedure as FeatInv for reconstruction~\cite{neukirch2025featinv}. Both models are loaded from the \texttt{timm} library with pretrained ImageNet-22k/1k weights, and inputs are normalized using standard ImageNet statistics. For DINOv3, we extract features from three public stages corresponding to transformer blocks 6, 9, and the final layer (Section~\ref{sec:feature_extraction}), using the official Hugging Face checkpoint. As long as the mapping models preserve feature dimensions, manipulated features can be passed directly to the FeatInv reconstruction pipeline. These feature maps are spatially large (e.g., up to $256\times48\times48 = 589{,}824$ activations per image at layer depth 1), posing significant memory and computational challenges for mapping models. Tab.~\ref{tab:convnext_features} summarizes the input resolutions and feature dimensions for all backbones and stages

\begin{table}[htbp]
\centering
\caption{Feature dimensions for all backbones and layers. For DINOv3, feature keys \texttt{feat1--feat3} correspond to public stages 1--3 (blocks 6, 9, and final).}
\label{tab:convnext_features}
\begin{tabular}{l c l c}
\toprule
Backbone & Input Images & Layer depth & Feature dimensions \\
\midrule
ConvNeXt  & $288\times288\times3$ 
  & feat1 & $256\times36\times36$ \\
  & & feat2 & $512\times18\times18$ \\
  & & feat3 & $1024\times9\times9$ \\
\midrule
SwinV2 & $384\times384\times3$ 
  &  feat1 & $256\times48\times48$ \\
  & & feat2 & $512\times24\times24$ \\
  & & feat3 & $1024\times12\times12$ \\
\midrule
DINOv3 & $224\times224\times3$ 
  &  feat1 (Block 6) & $768\times14\times14$ \\
  & & feat2 (Block 9) & $768\times14\times14$ \\
  & & feat3 (final) & $768\times14\times14$ \\
\bottomrule
\end{tabular}
\end{table}

\subsection{Details on finetuning the ConvNeXt and SwinV2 classification heads}
Tab.~\ref{tab:finetune_setup} presents all configurations we used for finetuning the backbone classifiers for the Stanford Cars dataset.

\begin{table}[ht]
\centering
\caption{Finetune setup for the ConvNeXt and SwinV2 classifier on Stanford Cars.}
\label{tab:finetune_setup}
\resizebox{0.8\textwidth}{!}{%
\begin{tabular}{ll}
\toprule
\textbf{Component} & \textbf{Configuration} \\
\midrule
Backbones & \texttt{convnext\_base.fb\_in22k\_ft\_in1k} \\
& \texttt{swinv2\_base\_window12to24\_192to384\_22kft1k} \\
Frozen stages & All backbone stages \\
Trainable parameters & Classifier head only (\texttt{head.fc}, \texttt{head.norm}) \\
Optimizer & AdamW, \texttt{weight\_decay=0.01} \\
Learning rate & 1e-3 \\
LR scheduler & Cosine annealing warm-restarts \\
Label smoothing & 0.1 \\
Dropout rate & 0.3 \\
Epochs & 30 (original train), 15 (original + augmented train) \\
Batch size & 128 \\
Training augmentation & \texttt{RandAugment}, random horizontal flip, random resized crop \\
Validation & 5\% stratified split \\
Validation transform & Resize/CenterCrop to 288 (ConvNeXt) / 384 (SwinV2) \\
\bottomrule
\end{tabular}%
}
\end{table}

\subsection{Further details on the metrics}
\label{sec:Appendix_metrics}

\paragraph{Reconstruction: Median Cosine-similarity in feature space}
Cosine similarity is defined in equation \ref{eq:cossim} where $\hat{F}_{1,h,w}$ and $\hat{F}_{2,h,w}$ are the two feature vectors and h,w the indices in the spatial feature map. MdnCS is then computed as $\text{median}({s_{h,w}})$ either across the entire feature map or within manipulation-specific masked regions. To address the limitation of global metrics for small, localized manipulations, binary masks are generated from pixel-wise differences between original and manipulated images (resized and Gaussian-smoothed, with Euclidean distances thresholded at 50) and downsampled via any-pooling over $H\times W$ cells to the feature-map resolution. $H$ and $W$ here are the spatial dimensions of each feature map. MdnCS is computed only within the masked spatial regions to enable more accurate evaluation of localized changes.
\begin{equation}
    s_{h,w} = \frac{\hat{F}_{1,h,w} \cdot \hat{F}_{2,h,w}}{\max(\|\hat{F}_{1,h,w}\|_2 \|\hat{F}_{2,h,w}\|_2, \epsilon)} 
    \label{eq:cossim}
\end{equation}

\paragraph{Reconstruction: Perceptual and Structural Similarity}
Perceptual similarity between the reconstructed and target images is measured via the learned perceptual image patch similarity (LPIPS)~\citep{DBLP:journals/corr/abs-1801-03924}, using the AlexNet backbone, which calculates the distance between the images internal feature representations to better capture human-aligned quality. The LPIPS metric is calculated with the formula in \ref{eq:LPIPS}. Here $\hat{f_l}$ are the normalized feature maps for the chosen layer $l$. 

\begin{equation}
    LPIPS(x,y) = \sum_lw_l||\hat{f}_l(x)-\hat{f}_l(y)||_2^2
    \label{eq:LPIPS}
\end{equation}

Additionally we use the structural similarity index (SSIM)~\citep{1284395} to evaluate preservation of local luminance, contrast, and structural patterns. Here $\mu_x$ and $\mu_y$ are the average luminance of the two images, $\sigma_x^2$ and $\sigma_y^2$ are variances of pixel intensities, $\sigma_{xy}$ is the covariance of pixel intensities and $C_1$ and $C_2$ are constants to prevent division by zero.

\begin{equation}
    SSIM(x,y) = \frac{(2\mu_x\mu_y+C_1)(2\sigma_{xy}+C_2)}{(\mu_x^2+\mu_y^2+C_1)(\sigma_x^2+\sigma_y^2+C_2)}
    \label{eq:ssim}
\end{equation}

\paragraph{Semantic quality: Impact on downstream classifier}
Semantic fidelity is assessed using a downstream classifier with frozen ConvNeXt and SwinV2 backbones finetuned on the 196 Stanford Cars classes. We report the classification accuracy of the mapped representations, the fraction of samples for which the Top-1 prediction coincides with that of the manipulated input 
and the Jensen–Shannon divergence (JSD) between original and tested classification probability distributions. JSD  between classification probability distributions $P$ and $Q$ is defined as
\begin{equation}
    \text{JSD}(P\|Q) = \frac{1}{2}D_\text{KL}(P\|M) + \frac{1}{2}D_\text{KL}(Q\|M), \quad M = \frac{P+Q}{2}
    \label{eq:jsd}
\end{equation}
where $D_\text{KL}$ denotes the Kullback-Leibler divergence and $M$ is the mixture distribution.

\subsection{Details on VLM-Based Evaluation}
\label{app:vlm-eval}

We use a vision-language model (Qwen2.5-VL-7B-Instruct) as an automatic judge to determine whether a given semantic or geometric edit was successfully reproduced in a reconstructed (``mapped'') image. Evaluation proceeds in two stages per example:

\begin{enumerate}
    \item \textbf{Original-check}: verify that the edit was validly applicable to the source image (e.g., the object to be modified is actually visible), using the original and target/ground-truth images.
    \item \textbf{Mapped-image judgment}: given the original image (and, depending on configuration, the target image), judge whether the \emph{mapped} prediction exhibits the edit.
\end{enumerate}

Each judgment is elicited with a fixed prompt template plus a one-sentence-then-verdict answer format, described below. All prompt text below is reproduced verbatim from the evaluation script.

\newenvironment{promptbox}{\begin{quote}\itshape\noindent\ignorespaces}{\par\end{quote}}

\subsubsection{Answer Format}

Every query to the VLM (both original-check and mapped-image judgment) ends with the same instruction, to force a parseable verdict:

\begin{promptbox}
First, in one short sentence (max 15 words), describe what you actually observe in the relevant area of the last image. Then, on its own final line, write exactly one of the following two lines and nothing else after it: ``Verdict: Yes'' or ``Verdict: No''.
\end{promptbox}

\subsubsection{Mapped-Image Grading Rubric}

For the main judgment (does the mapped/predicted image show the edit), the model receives the following grading rules before the question. The rubric instructs the model to judge the end state of the image, not how different it looks from the original:

\begin{promptbox}
\begin{itemize}
    \item Yes: the requested property is clearly and substantially present.
    \item No: the property is weak, faint, ambiguous, partial, or altogether absent. If you are not confident the property is clearly and substantially present, answer No -- do not give credit for a borderline or subtle effect. This ``weak/faint'' default is overridden by any per-edit note above that explicitly says a dull, muted, or blurry version of the property still counts -- follow that specific note instead whenever it applies.
    \item Exception: if the scene already had this property somewhat in the original photo and still clearly has it in the last image, answer Yes even if the two images look similar -- do not penalize lack of visible change when the property was already there and remains clearly there.
    \item Never fail the image for being blurrier, duller, or less sharp/vivid than the other image(s) shown to you -- the mapped/predicted image (the last image shown) comes from a lossy compressed representation and is expected to look softer and less saturated than any original or target photo shown alongside it. That expected quality gap is never a valid reason for No; judge the last image on its own terms using the criteria above, not by comparing its overall sharpness, vividness, or brightness to the other image(s) shown before it.
    \item Self-consistency check before you answer: if your one-sentence description says the property is visible (e.g. you describe seeing the target color, a glow, a shape), your verdict must be Yes even if you also note it looks blurry, dim, dull, or not fully sharp -- do not describe the property as present and then answer No anyway.
\end{itemize}
\end{promptbox}

The rationale for this rubric was to correct two failure modes observed empirically: (1)~the model over-penalizing correct-but-blurry reconstructions by comparing them against sharper reference images, and (2)~inconsistency between the model's free-text description and its final Yes/No verdict.

\subsubsection{Original-Check Rubric}

Before grading a mapped prediction, we first verify the edit was validly applicable to the source image:

\begin{promptbox}
\begin{itemize}
    \item Yes: the requested property is clearly and substantially present in Image 2, applied to the correct, relevant part of the image.
    \item No: the edit failed. This includes the property being absent, faint, or ambiguous, and also includes the case where the relevant part (e.g. the tire rims, the headlights, the side mirror mounts) is not clearly visible or present in Image 1 (occluded, cropped out, too small, etc.), which causes the edit to be misapplied to a random or nonsensical location. If the edit is not on the correct part, or you cannot identify the correct part in Image 1 at all, answer No.
    \item Unless the edit explicitly calls for adding a new object, it must be achieved by modifying an object that is already present in Image 1, not by adding a new object that was not there before. For example, ``turn on lamps'' means switching on a lamp already visible in Image 1 -- it does not justify a new lamp appearing in Image 2 that was not present in Image 1. If Image 1 has no such object to modify (e.g. no lamp at all), the edit cannot be correctly applied, so answer No, even if Image 2 shows the desired effect via a newly-added object.
\end{itemize}
\end{promptbox}

Rows failing this check are excluded from the reported mapping-success rate rather than counted against the model. For deterministic augmentations (grayscale, hue shift, noise, mirror/rotation, masking), this check is skipped by construction, since these transforms always apply correctly.

\subsubsection{Prompt Assembly}

The full prompt sent to the VLM combines: an image preamble (identifying Images 1/2/3 as original/target/mapped), an -optional per-manipulation hint, an optional per-category note, the grading rubric, the yes/no question, and the answer-format instruction. For example, without a target image, the assembled prompt reads (in order): an original/edited image preamble naming the requested edit; an optional ``Where to look'' hint; an optional category note; the grading rubric above; the question ``Does Image 2 show evidence of the edit \textless{}edit text\textgreater{}?''; and the answer-format instruction.

When a target/reference image is included, for deterministic transforms it is framed as an exact ground truth to be matched; for semantic edits it is framed as one non-exhaustive example, to avoid over-constraining valid variation:

\begin{promptbox}
Image 2 is a target example showing one valid way the edit \textless{}edit text\textgreater{} can look once applied. Use it only as a rough reference for the intended property, not as an exact pixel-matching template -- different lighting, angle, or exact shade than Image 2 is fine.
\end{promptbox}

\subsubsection{Per-Manipulation Hints}
\label{app:vlm-eval-hints}

For manipulations prone to systematic over- or under-grading, we add a targeted ``where to look'' hint, reproduced in full below.

\smallskip
\noindent\textbf{Stanford Cars: Add police lights.}
\begin{promptbox}
Look at the roof of the car for an added light bar or police lights. Be lenient: it does not need to look like a clean, distinct light-bar shape -- if you can see a colored mark, blob, or protrusion on the roof that is not part of a plain unmodified roofline, and a viewer could reasonably guess `something was added here, probably lights,' answer Yes. Only answer No if the roofline still looks plain and unmodified, with nothing distinguishable added -- general haze or blur spread evenly across the whole car (not concentrated on the roof) does not count as this, since that is just overall reconstruction softness, not an added object.
\end{promptbox}

\noindent\textbf{Stanford Cars: Change body color blue.}
\begin{promptbox}
Look at the main body panels (hood, doors, roof, fenders) for the overall paint color. Judge color only -- a soft/blurry reconstruction can still clearly show a blue body, so do not answer No just because fine detail is missing. But you must genuinely be able to see blue paint -- do not answer Yes just because the edit was supposed to happen; if the panels still basically look like their original color, even softly or blurrily, that is still a No. If you can see any shade of blue on the panels -- however dull, pale, dark, or washed-out -- answer Yes; do not answer No because the blue looks less vivid or less sharply defined than in Image 1 or Image 2, that comparison is not a valid reason for No.
\end{promptbox}

\noindent\textbf{Stanford Cars: Change tire rim color red.}
\begin{promptbox}
Look closely at the wheel rims/hubcaps (the metal part inside the tire) for their color. Judge color only -- a soft/blurry reconstruction can still clearly show red rims, so do not answer No just because fine detail is missing. But you must genuinely be able to see red -- do not answer Yes just because the edit was supposed to happen; if the rims still basically look like their original color, even softly or blurrily, that is still a No. If you can see any shade of red on the rims -- however dull, pale, dark, or washed-out -- answer Yes. Do not answer No just because this photo's red looks less vivid, less clear, or less sharply defined than a cleaner reference photo shown alongside it -- that comparison is not a valid reason for No; judge only whether red (of any shade) is visible on the rims in this one photo.
\end{promptbox}

\noindent\textbf{Stanford Cars: Remove side mirrors.}
\begin{promptbox}
Look carefully at both sides of the car, right next to the front door windows, where side mirrors are normally mounted -- check specifically whether a mirror is present or absent there. Be lenient: if the mirror looks noticeably faded, thinned, smudged-away, or otherwise visibly diminished compared to a solid, intact mirror shape -- to the point a viewer could reasonably guess an attempt was made to erase it -- answer Yes, even if a faint trace remains. Only answer No if a full, solid, clearly-intact mirror shape is still sitting there, undiminished relative to how solidly-formed the rest of the car's similarly-sized details look in this same image.
\end{promptbox}

\noindent\textbf{Stanford Cars: Turn on headlights.}
\begin{promptbox}
Look closely at the headlight housings at the front of the car for any sign of illumination. Be lenient: it does not need to look like a bright, clearly-lit bulb -- if the headlight housing area shows a distinguishable brightening, glow, or whitish/yellowish tint compared to the rest of the car in this one photo, and a viewer could reasonably guess `this looks like it's meant to be lit,' answer Yes. Only answer No if the headlight housings still look like plain, unlit, dark/gray sockets, no different from the rest of the car. Judge this only by comparing the headlight area to the rest of this one photo -- never by judging how bright, sharp, or vivid this photo looks overall compared to any other, cleaner photo shown alongside it (the original, or a reference example of the edit); this reconstruction is expected to look dimmer, duller, or blurrier overall than a clean photo, and that alone is never a valid reason for No. General haze or blur spread evenly across the whole car (not a distinguishable change concentrated at the headlights) does not count on its own, since that is just overall reconstruction softness, not illumination.
\end{promptbox}

\noindent\textbf{LSUN Bedroom: Color bedding blue.}
\begin{promptbox}
Look specifically at the sheets, blankets, and pillows on the bed for their color -- ignore the color of everything else in the room. Be lenient: it does not need to look predominantly or vividly blue -- if the bedding shows a distinguishable bluish tint, cast, or shift away from its original color, even if faint, dull, patchy, or mixed with other colors, and a viewer could reasonably guess `this bedding looks like it's meant to be blue,' answer Yes. Only answer No if the bedding still looks like its plain original color (e.g. cream, white, yellow) with no discernible shift toward blue anywhere on it. Do not answer No just because this photo's blue looks less vivid, less clear, or less sharply defined than a cleaner reference photo shown alongside it -- that comparison is not a valid reason for No; judge only whether blue (of any shade) is visible on the bedding in this one photo.
\end{promptbox}

\noindent\textbf{LSUN Bedroom: Add teddybear.}
\begin{promptbox}
Look on top of the bed for a single object placed roughly in the middle that was not there in the original photo. Be lenient: it does not need to clearly resemble a teddy bear (no need for a distinct head, ears, or ideal silhouette) -- if there is a compact, localized blob or shape sitting on the bed that a viewer could reasonably guess is `some object placed there, maybe a teddy bear,' answer Yes. Only answer No if the bed surface still looks empty or undisturbed, matching the original, with nothing localized added on top of it -- general haze or blur spread evenly across the whole bed (not a compact added shape) does not count, since that is just overall reconstruction softness, not an added object.
\end{promptbox}

\noindent\textbf{LSUN Bedroom: Turn on lamps.}
\begin{promptbox}
Look at any lamp(s) already visible in the original image, in that same spot, for signs they are switched on. Be lenient: it does not need to look like a clearly glowing bulb -- if the lamp area shows a distinguishable brightening, glow, or lightening compared to the rest of the room in this one photo, and a viewer could reasonably guess `this lamp looks like it's meant to be on,' answer Yes. Only answer No if the lamp area still looks just as dim or unlit as the rest of the room. Judge this only by comparing the lamp area to the rest of this one photo -- never by judging how bright, sharp, or vivid this photo looks overall compared to any other, cleaner photo shown alongside it (the original, or a reference example of the edit); this reconstruction is expected to look dimmer, duller, or blurrier overall than a clean photo, and that alone is never a valid reason for No. General haze or blur spread evenly across the whole room (not a distinguishable change concentrated at the lamp) does not count on its own, since that is just overall reconstruction softness, not illumination.
\end{promptbox}

\noindent\textbf{LSUN Bedroom: Make bed unmade.}
\begin{promptbox}
Look at the bedding for a rumpled, creased, recently-slept-in look rather than a smooth, neatly made bed -- specifically, look for visible folds, wrinkles, or an uneven or disturbed surface on the sheets and pillows. Be lenient: it does not need sharp, well-defined creases -- if the bedding's outline or surface looks noticeably uneven, lumpy, or irregular in a way that a viewer could reasonably guess `this looks disturbed or slept-in rather than freshly made,' answer Yes. Only answer No if the bed's overall shape and edges still read as flat, smooth, and rectangular like a neatly made bed -- a generally blurry or hazy look spread evenly across the whole image (not a specifically uneven or irregular bed surface or outline) does not count on its own, since that is just overall reconstruction softness, not messy bedding.
\end{promptbox}

\subsubsection{Category-Level Notes}
\label{app:vlm-eval-category}

For deterministic augmentations, a shared note is added per transform category, in place of a manipulation-specific hint.

\smallskip
\noindent\textbf{Geometric (flip / rotation).}
\begin{promptbox}
This is a whole-image geometric transform (a mirror-flip or rotation), not a localized edit. Pick 2--3 distinctive, recognizable objects in the scene (e.g. the headboard, a lamp, a window, a piece of furniture) and check whether their positions and orientation relative to each other and to the frame match the described transform -- do not judge from a vague overall impression of `does this look sideways or flipped,' since that is unreliable, especially on a blurry image. If a rotation's exact direction (e.g. 90 vs 270 degrees, which look similar) is genuinely hard to tell apart, judge primarily whether a rotation of roughly that magnitude is clearly present at all, and only mark No if the image still looks essentially upright or unrotated.
\end{promptbox}

\noindent\textbf{Occlusion (masking).}
\begin{promptbox}
This transform covers a rectangular region of the image with a solid, uniformly-colored mask -- judge whether a sharp-edged, flat-colored rectangular patch is present in roughly the stated location, with the rest of the image around it still showing normal room detail. General blur, softness, or haze spread across the whole image is not a mask and does not count -- feature-space reconstructions are very often blurry or hazy everywhere, and that alone is not evidence of an occluded region. Only answer Yes if you can point to one specific, compact area that is distinctly flatter or more uniform than its surroundings, not the image as a whole.
\end{promptbox}

\noindent\textbf{Noise.}
\begin{promptbox}
This transform adds random pixel-level noise or grain across the whole image -- judge overall graininess or speckle in otherwise-smooth areas (walls, bedding), not a localized change.
\end{promptbox}

\noindent\textbf{Global color (grayscale, hue shift).}
\begin{promptbox}
This is a whole-image color transform -- judge the overall color cast or saturation of the whole scene, not a localized change. Judge based on color, not fine detail: a blurry or soft image can still clearly show the correct color, so do not answer No merely because detail is missing. But the reverse also matters -- do not answer Yes just because the edit was supposed to happen, or because the image is generically blurry or washed-out. You must be able to genuinely perceive the stated color or desaturation; if the colors still basically look like the original, that is a No, even if the image is soft or hazy overall.
\end{promptbox}

\subsection{Information about compute resources}
\label{sec:compute}
All experiments were conducted on a Linux-based cluster utilizing NVIDIA L40 GPUs. Dataset augmentation using the Qwen-Image-Edit model required a multi-GPU setup, as the model footprint (weights and activation overhead) reaches approximately 57 GB; consequently, we utilized two L40 GPUs (each with 48 GB VRAM) to enable stable inference via accelerate model sharding (device\_map="balanced"). This configuration achieved an inference speed of approximately 30 seconds per image. 
Training mapping models and performing feature extraction were executed on single L40 GPUs, with individual training runs ranging from few minutes to hours depending mainly on the selected model architecture and feature depth. Training mapping models for the earliest ConvNeXt layer especially with the transformer model requires additional time. On one L40 GPU each application of a mapping model, reconstructing to image space with FeatInv and then calculating the evaluation metrics requires approximately 5 seconds per sample. Due to the large amount of manipulation and backbone combinations and the size of the feature dimensions 2 TB of storage space was required.

\subsection{Evaluating semantic preservation of mapped features} 
\label{sec:semantic}
Recent discussions in the field of self-supervised learning \citep{balestriero2024learning} emphasized that features suitable for reconstruction do not necessarily coincide with features suitable for semantic tasks. We therefore also assess the semantic quality of the mapped features as compared to the original features by means of a finetuned downstream classifier. To evaluate the semantic quality of the learned mappings using a downstream classifier, we finetune both backbones on Stanford Cars for the 196 available classes. We freeze all weights in the backbone model and update only the weights of the linear classification head, to preserve the original feature representations and the compatibility with the FeatInv reconstruction. For each of the backbones we train two variants: One for 30 epochs on the original training set alone, and another for 15 epochs on the significantly larger combined set (original + augmented data), which converges much faster due to its scale. This not only boosts classifier robustness \citep{DBLP:journals/corr/abs-1903-12261, liu2023comprehensivestudyrobustnessimage} but ensures fair mapping comparisons by standardizing evaluation across original and manipulated features. We train all models using the AdamW optimizer with a CosineAnnealingWarmRestarts scheduler, employing cross-entropy loss with label smoothing. The full finetuning setup is outlined in App. Tab.~\ref{tab:finetune_setup}.

\begin{figure}[ht]
   \centering
   \includegraphics[width=0.6\linewidth]{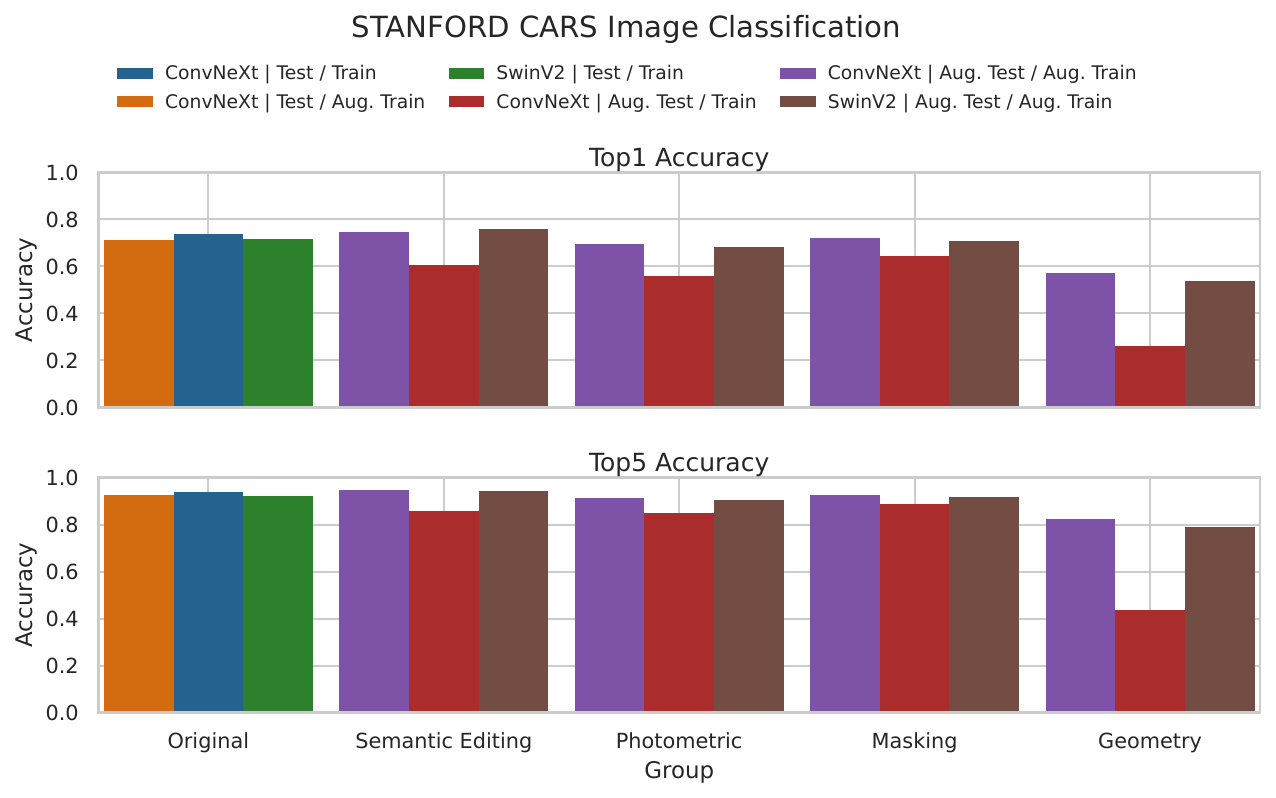}
    \caption{Image classification results using the finetuned Backbone models with both the original and the augmented Stanford Cars image datasets.}
\label{fig:image_classification_top1_top5_acc}
\end{figure}

\begin{figure}[ht]
    \centering
    \includegraphics[width=0.6\linewidth]{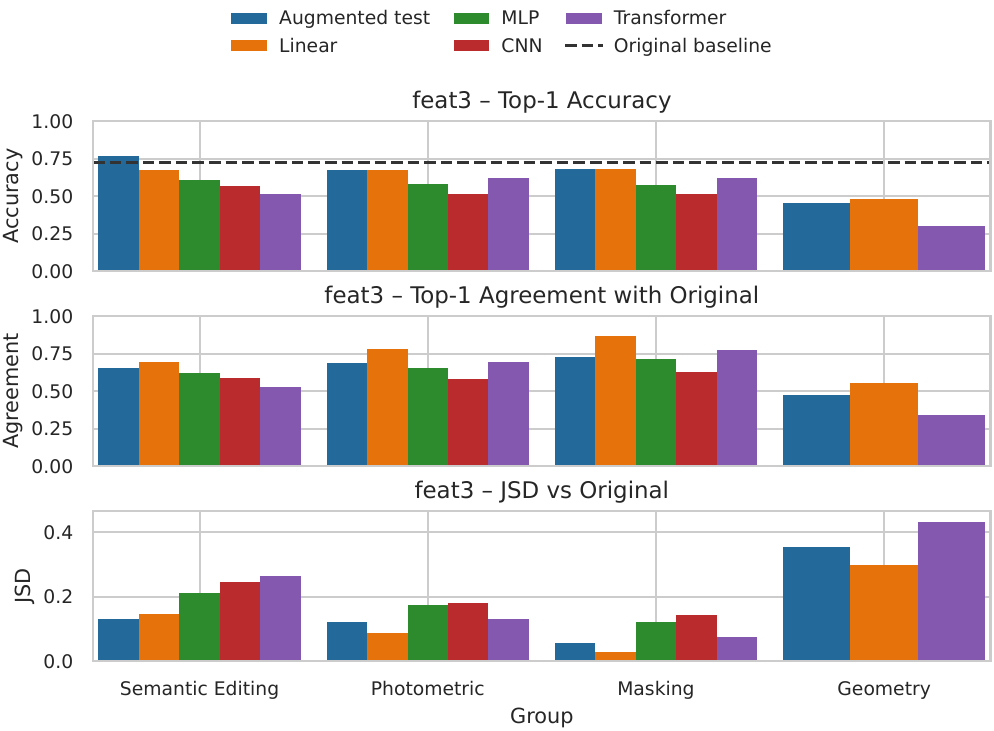}
    \caption{ConvNeXt classification performance on Stanford Cars, measured by Top-1 accuracy, agreement, and JSD. The dashed line indicates performance on the unmodified test set.
    }
    \label{fig:feat3_classifier_metrics}
\end{figure}

\paragraph{Classification results}
Fig.~\ref{fig:image_classification_top1_top5_acc} reports Top-1 and Top-5 accuracy for both backbones on the original and augmented Stanford Cars data. Training on the original data achieves 0.74 Top-1 accuracy, while augmentation slightly reduces performance on the original test set but improves accuracy on augmented sets, particularly under out-of-distribution geometric transformations. Similar results across backbones indicate that performance depends more on the training data than on the architecture. Direct feature classification on the full test set (Fig.~\ref{fig:feat3_classifier_metrics}) shows stable Top-1 accuracy and prediction agreement under semantic, photometric, and masking manipulations, with the largest degradation for geometric transformations. Overall, mapping did not impact classification performance significantly compared to the direct target manipulations.

\FloatBarrier

\subsection{Assessing geometric accuracy of the mapping in feature space}
\label{sec:geom_accuracy}
\begin{figure}[!htbp]
    \centering
    \includegraphics[width=0.8\linewidth]{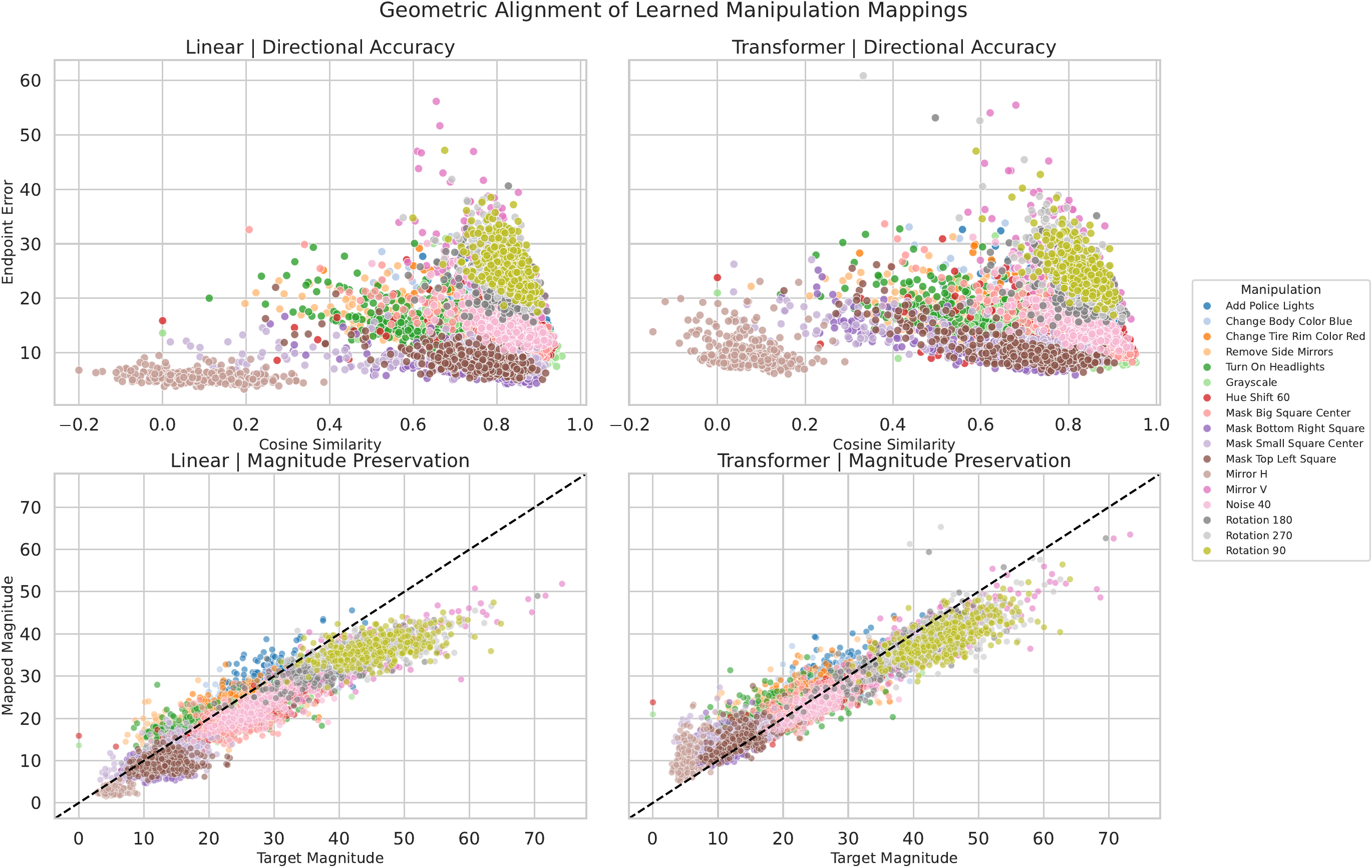}
    \caption{Visualization of geometric evaluation metrics assessing how accurately the learned mappings reproduce manipulation-induced embedding shifts. The top row measures directional alignment between target and mapped displacement vectors, where cosine similarity indicates preservation of movement direction and endpoint error quantifies the remaining displacement mismatch. The bottom row evaluates magnitude preservation, comparing the length of the target manipulation shift against the predicted mapped shift.}
    \label{fig:geom_metrics}
\end{figure}

Fig.~\ref{fig:geom_metrics} summarizes the geometric evaluation metrics for both mapping approaches across all manipulation types for the ConvNeXt feat3 features. The upper row evaluates directional consistency by plotting cosine similarity against endpoint error. For each sample, we define the target displacement as
\[
\Delta_{\mathrm{target}} = x_{\mathrm{manipulated}} - x_{\mathrm{original}},
\]
and the mapped displacement as
\[
\Delta_{\mathrm{mapped}} = x_{\mathrm{mapped}} - x_{\mathrm{original}}.
\]
Directional agreement is measured using cosine similarity between
\(\Delta_{\mathrm{target}}\) and \(\Delta_{\mathrm{mapped}}\), while endpoint error is defined as
\[
E_{\mathrm{endpoint}} =
\left\lVert
\Delta_{\mathrm{mapped}} - \Delta_{\mathrm{target}}
\right\rVert.
\]
Higher cosine similarity indicates that the mapped displacement more closely follows the target manipulation direction, whereas larger endpoint error indicates greater mismatch between the predicted and target feature-space shifts. The lower row compares the magnitudes
\(\lVert\Delta_{\mathrm{target}}\rVert\) and
\(\lVert\Delta_{\mathrm{mapped}}\rVert\); points closer to the diagonal indicate better preservation of manipulation strength.

Directional alignment between target and mapped manipulations, measured via cosine similarity, reveals a highly structured but anisotropic geometry in the feature space. Global transformations such as rotations and grayscale conversion exhibit consistently high alignment (mean cosine $\approx$ 0.82-0.87), whereas localized structural edits such as masking and object removal show substantially lower alignment. Horizontal mirroring produces near-orthogonal displacement directions, suggesting that this transformation is not captured by the primary linear structure of manipulation-induced variations in the learned embedding space, although it remains recoverable through the learned mapping in image space. Importantly, this structure is consistent across both models, suggesting that directional geometry is primarily induced by the underlying feature representation rather than the learned mapping. While the transformer improves magnitude preservation and overall regression accuracy, it does not substantially alter the directional organization of manipulations.

The transformer model preserved displacement magnitudes more accurately than the linear model, achieving a lower prediction error (RMSE = 4.42 vs. 5.99). Despite this improvement in overall accuracy, both models systematically compressed the magnitude scale, as indicated by regression slopes substantially below the ideal value of 1 (0.753 for the transformer and 0.767 for the linear model). This suggests that larger target magnitudes were consistently underestimated by both approaches. The linear model exhibited a smaller intercept (2.11 vs. 6.07), indicating less additive bias and slightly better calibration near the origin, whereas the transformer introduced a larger positive offset. Overall, these results indicate that the transformer more faithfully preserved target magnitudes, although neither model achieved ideal magnitude preservation due to systematic compression of the magnitude scale.

\FloatBarrier

\subsection{Additional results using the truncated SVD mappings}
\label{subsec:additional_results_trunc_svd}
\begin{figure}[!htbp]
    \centering
    \includegraphics[width=0.95\linewidth]{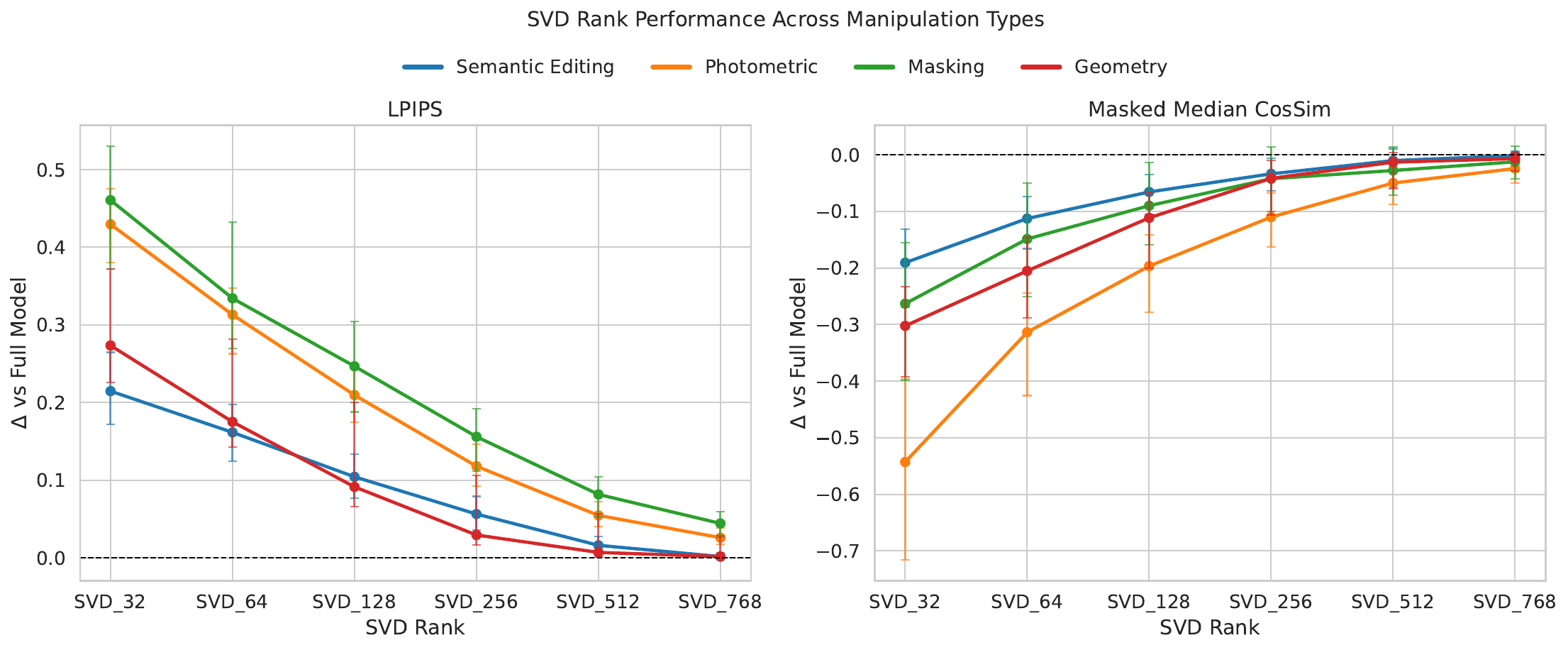}
    \caption{Change of LPIPS and masked MdnCS with SVD mappings using only the top-k singular values against the full linear model for the feat3 ConvNeXt features. Metrics are reported as the median and interquartile range (IQR).}
    \label{fig:svd_vs_full}
\end{figure}

Fig.~\ref{fig:svd_vs_full} shows the progression of metrics for each of the SVD truncation steps against the unaltered full mapping. For all manipulation groups both metrics improve with increasing SVD $k$. However, improvement dimishes with higher $k$ and the resulting differences between the full model and large $k$ mappings is neglible.

\begin{figure}[!htbp]
    \centering
    \includegraphics[width=0.95\linewidth]{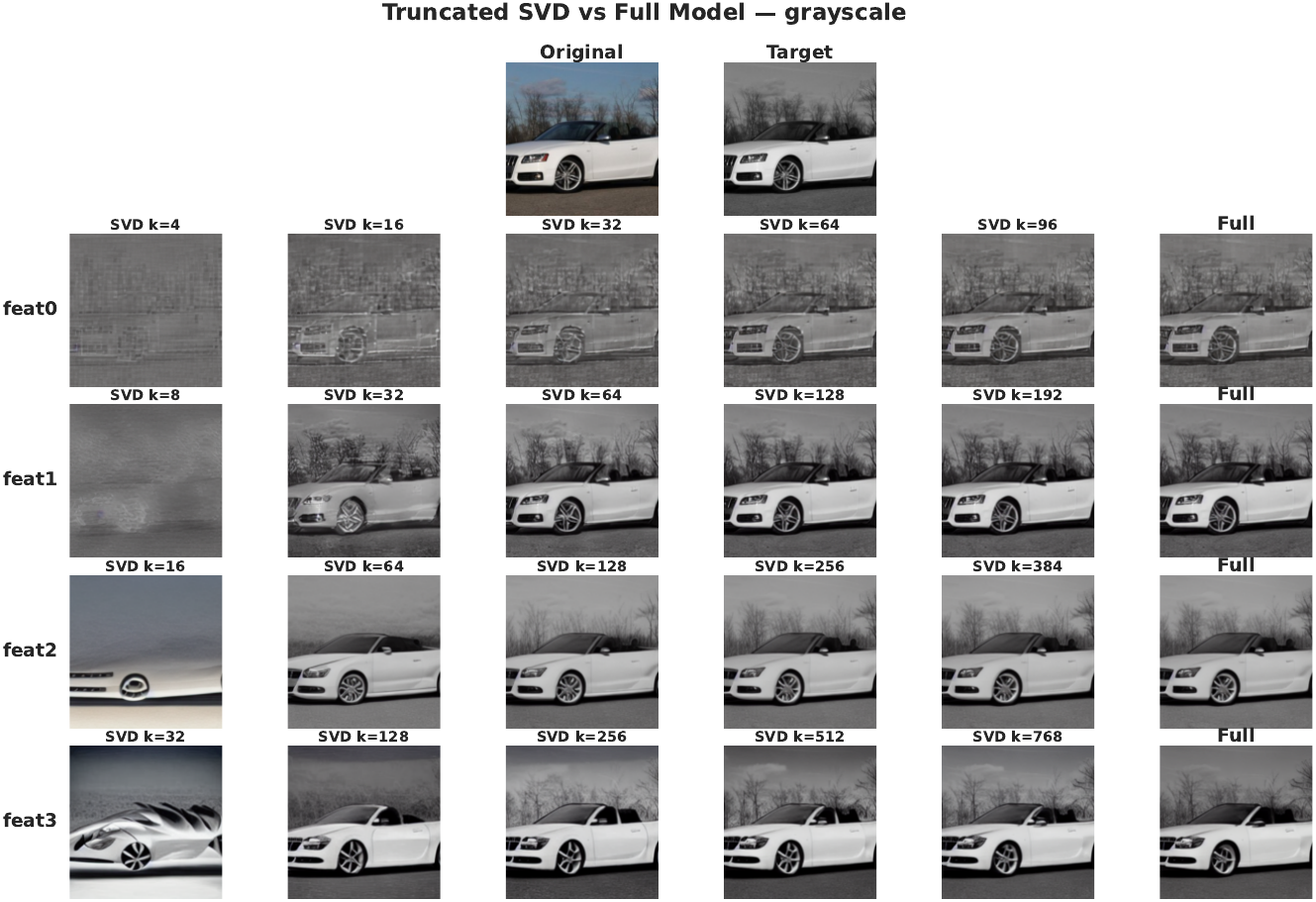}
    \caption{Reconstructed image examples for the \emph{Grayscale} manipulation created with truncated SVD or full model mapping for all feature depths (ConvNeXt).}
    \label{fig:truncated_svd_all_feats_manip_grayscale}
\end{figure}

\begin{figure}[!htbp]
    \centering
    \includegraphics[width=0.95\linewidth]{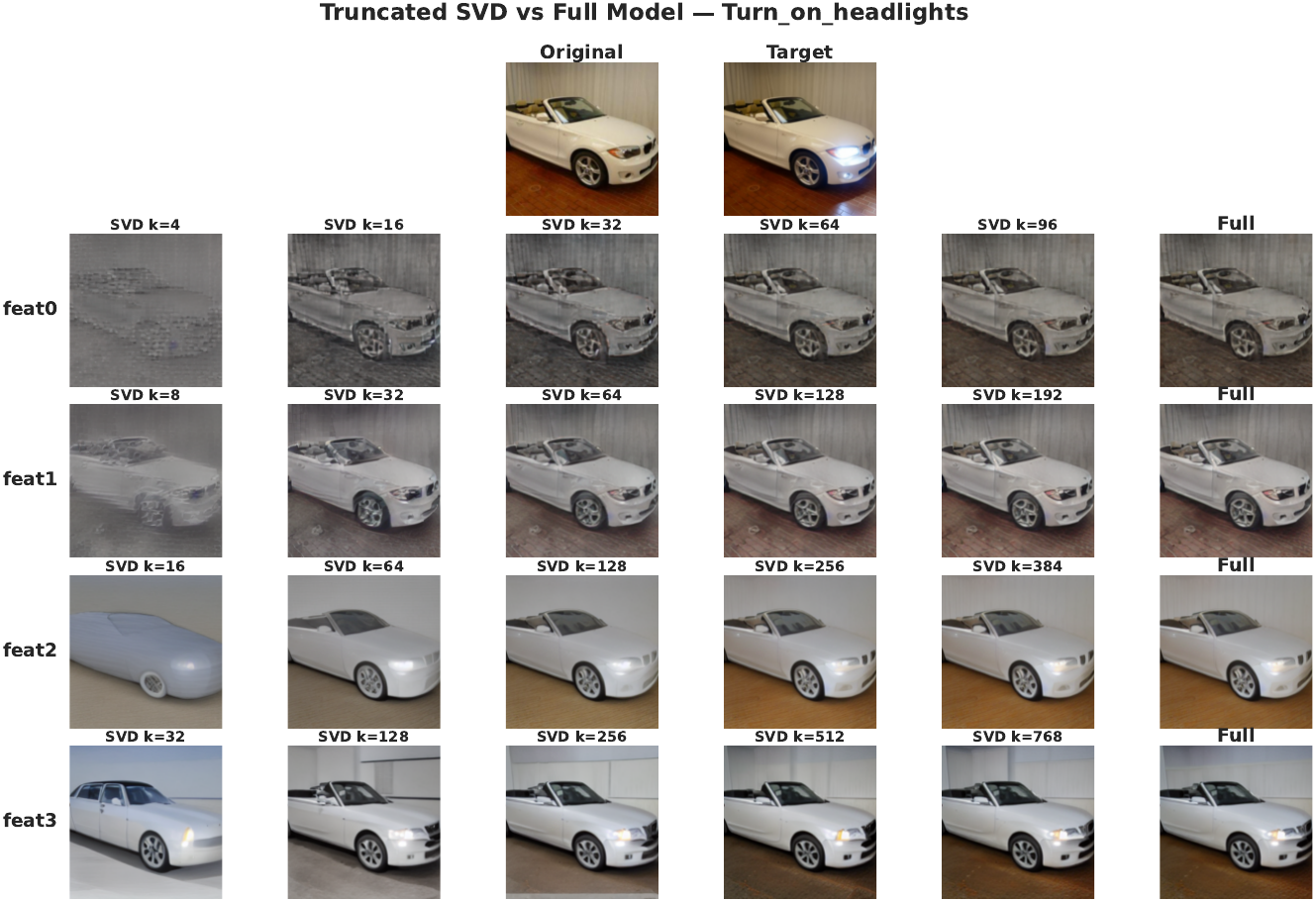}
    \caption{Reconstructed image examples for manipulations created with truncated SVD or full model mapping for the \emph{Turn on headlights} manipulation for all feature depths. (ConvNeXt)}
    \label{fig:truncated_svd_all_feats_manip_Turn_on_headlights}
\end{figure}

To investigate this effect also for earlier feature depths Figs.~\ref{fig:truncated_svd_all_feats_manip_grayscale} and \ref{fig:truncated_svd_all_feats_manip_Turn_on_headlights} demonstrate that manipulations are encoded in the early $k$ but not always also in all feature depths.

\FloatBarrier

\subsection{Effect of Feature Normalization}
\label{subsec:effect_norm}
\begin{figure}[ht]
    \centering
    \includegraphics[width=0.8\linewidth]{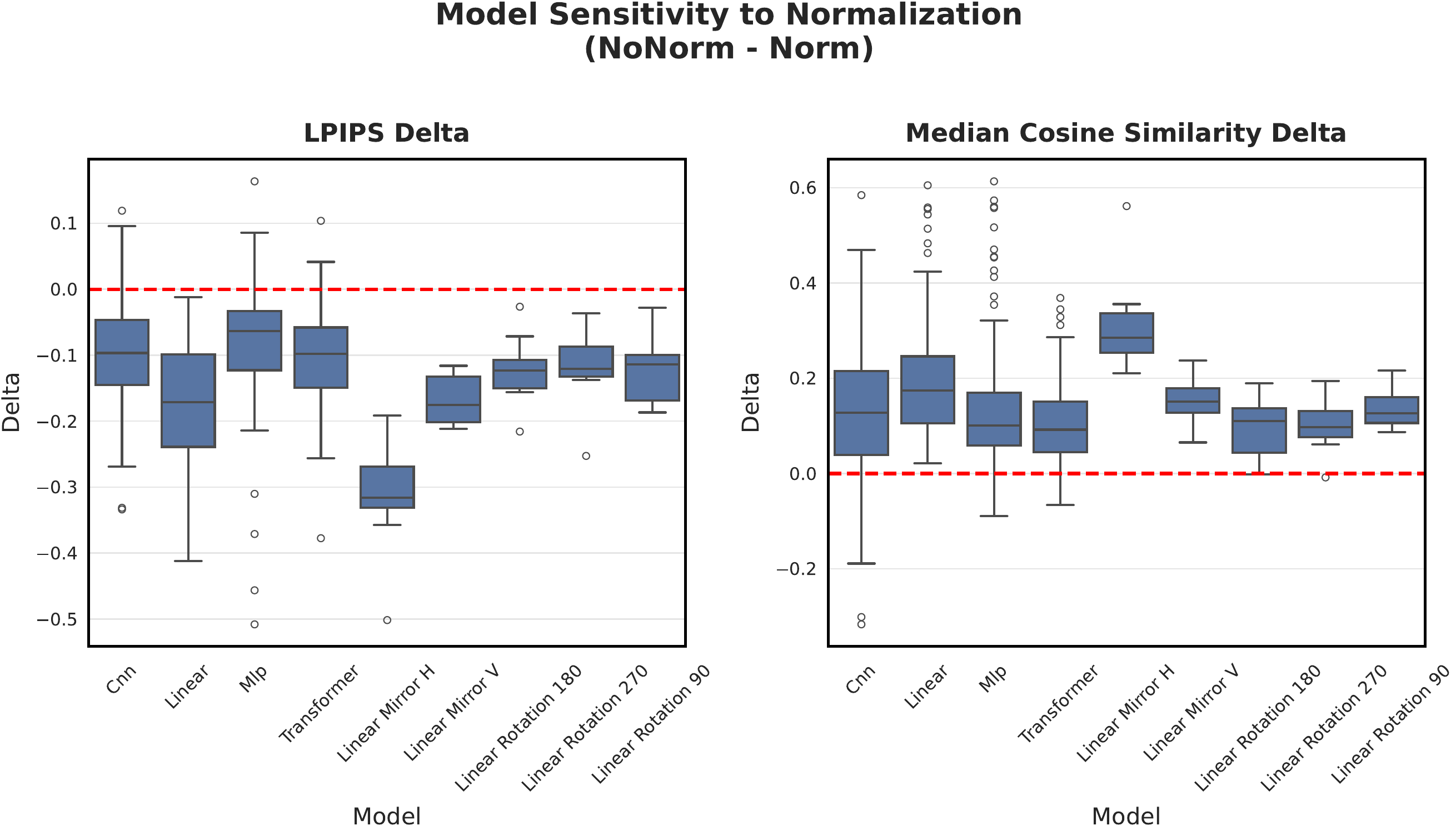}
    \caption{Effect of feature normalization on image quality for various mapping model architectures for the ConvNeXt feat3 features}
    \label{fig:norm_model_sensitivity}
\end{figure}

\begin{figure}[ht]
    \centering
    \includegraphics[width=0.7\linewidth]{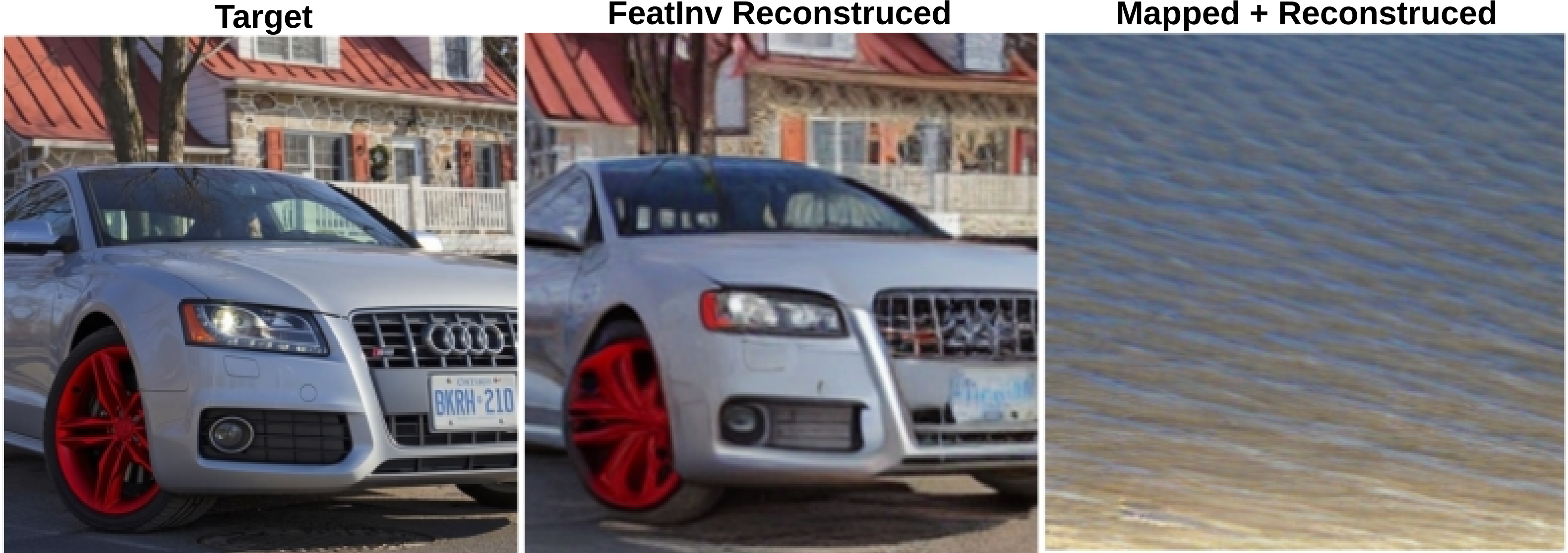}
    \caption{Example for distorted image results when training with unnormalized features in earlier ConvNeXt layers (here feat2)}
    \label{fig:norm_distortion_example}
\end{figure}

During preliminary experiments, we observed that feature normalization substantially changes mapping outcomes across all tested mapping architectures. In our pipeline, features are first optionally normalized before being passed to the mapping model, and the predicted features are subsequently denormalized prior to feature-to-image reconstruction. This ensures that reconstruction with FeatInv operates in the original feature space. Fig.~\ref{fig:norm_model_sensitivity} shows the differences in mapping metrics between original and normalized features. In particular, normalization tends to reduce reconstruction fidelity, likely because it removes magnitude information that is important for accurately recovering image structure. This observation is consistent with \citet{allakhverdov2026feature}, who note that normalization suppresses magnitude cues, although they still adopt normalized features in their approach when applying linear mappings on features.
Given the substantial impact on reconstruction quality we used unnormalized features for training mapping models for the last feature stages of ConvNeXt and SwinV2. During our experiments we also observed that training with unnormalized features was unstable for the earlier feature stages of ConvNeXt, making it necessary to use normalized features. In extreme cases unnormalized features resulted in distorted, wavy patterns as shown in figure \ref{fig:norm_distortion_example}. For all earlier layers, normalization was therefore necessary to ensure stable training.

\FloatBarrier

\subsection{Mapped samples}
\label{subsec:appendix_mapped_samples}
\subsection{Stanford Cars}
\begin{figure}[ht]
    \centering
    \includegraphics[width=0.8\linewidth]{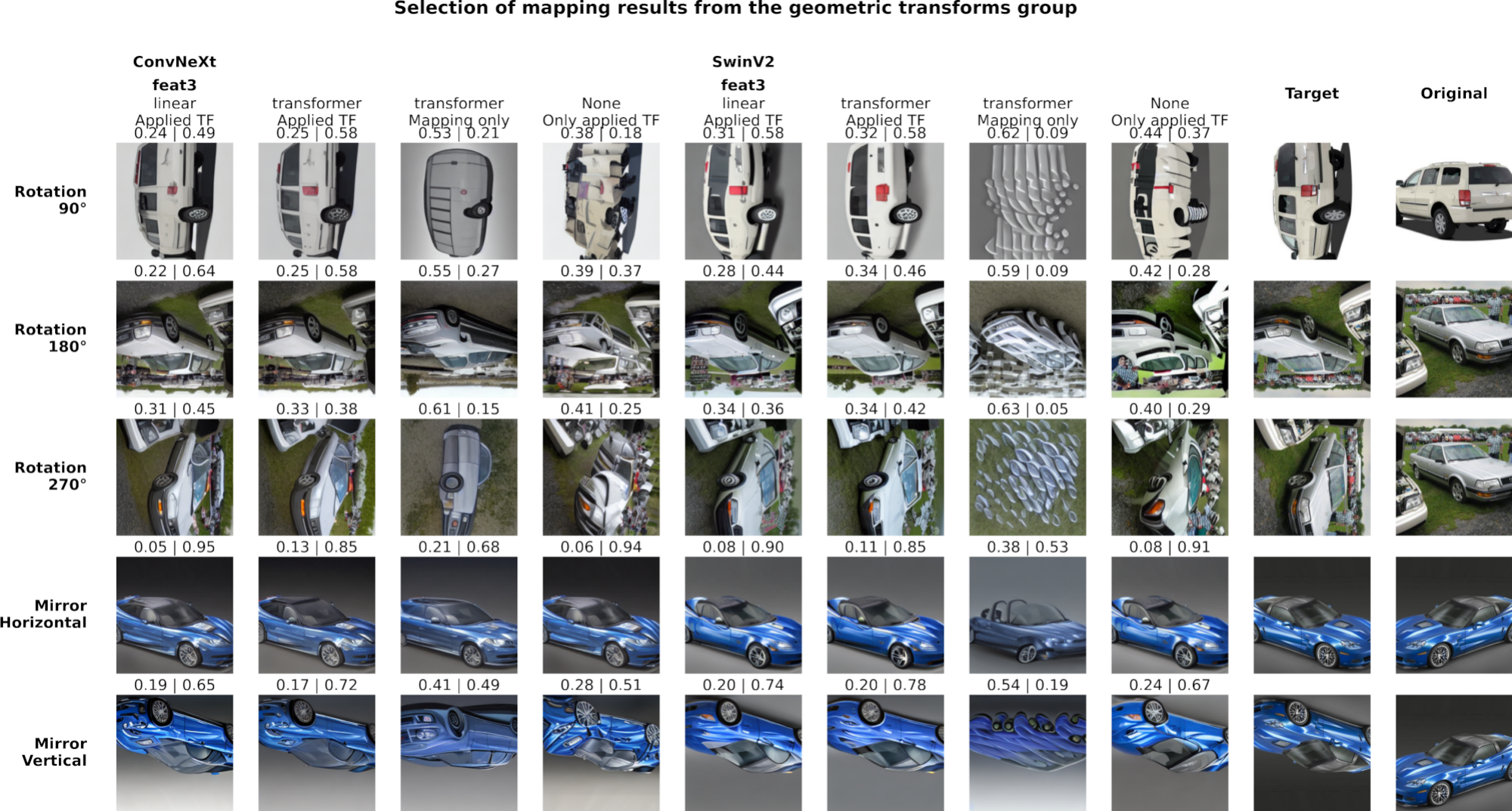}
    \caption{Example mapped and reconstructed images for selected geometric manipulations. They were created using features from both ConvNeXt and SwinV2 stages. For each image the LPIPS and MdnCSmetrics between target image and mapped, reconstructed image are shown. Applied TF: Reordering of the feature vector possitions is applied, Mapping only: No reordering - just the mapping model is applied}
    \label{fig:geom_samples}
\end{figure}

In figure~\ref{fig:geom_samples} we present additonal examples from mapped and reconstructed images for selected geometric manipulations. For comparison also included are samples where just the spatial feature permutations were performed without any mapping models. When the spatial feature permutations are applied generally all tested models are then able to apply the local mapping to the feature vectors. The transformer model is able to apply the geometric manipulation though with clear degradation in image quality and a loss in structure of the cars. Notably the transformer mappings on the SwinV2 samples show higher degrees of image degradation and loss of most car features. Of the tested geometric transformations, only horizontal mirroring could be achieved using spatial feature permutations alone, without noticeable loss in quality.

\begin{figure}[ht]
    \centering
    \includegraphics[width=0.6\linewidth]{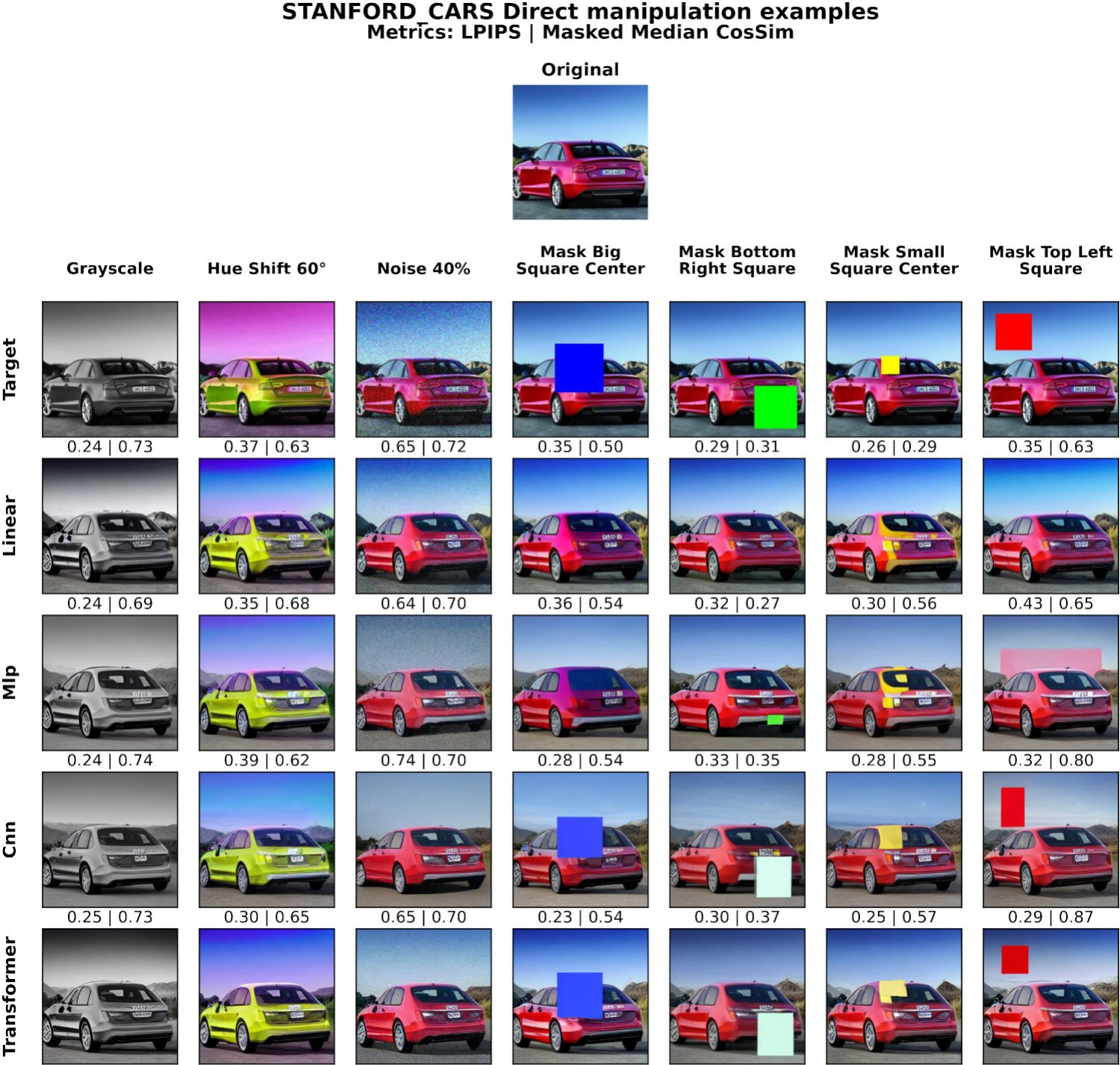}
    \caption{Example mapped and reconstructed images for each direct manipulation type. They were created using features extracted from the last ConvNeXt layer \textbf{feat3} before the classification head. For each image the LPIPS, MdnCS and Masked MdnCS metrics between target image and mapped, reconstructed image are shown.}
    \label{fig:direct_manipulations_id_690}
\end{figure}

Fig.~\ref{fig:direct_manipulations_id_690} shows more examples of manipulation and model combinations on the same original image for comparison using features extracted from the last ConvNeXt layer feat3. Color transformations \emph{Grayscale} and \emph{Hue shift} are shown to be possible with all mapping models. While all models applied were able to apply noise to the image, the target noise strength was not fully reached. Local masking proved to be challenging for the local-only mapping models, while the CNN and Transformer models were able able to apply the masking with occasional loss in exact mask shape and color.

\begin{figure}[!htbp]
    \centering
    \includegraphics[width=0.6\linewidth]{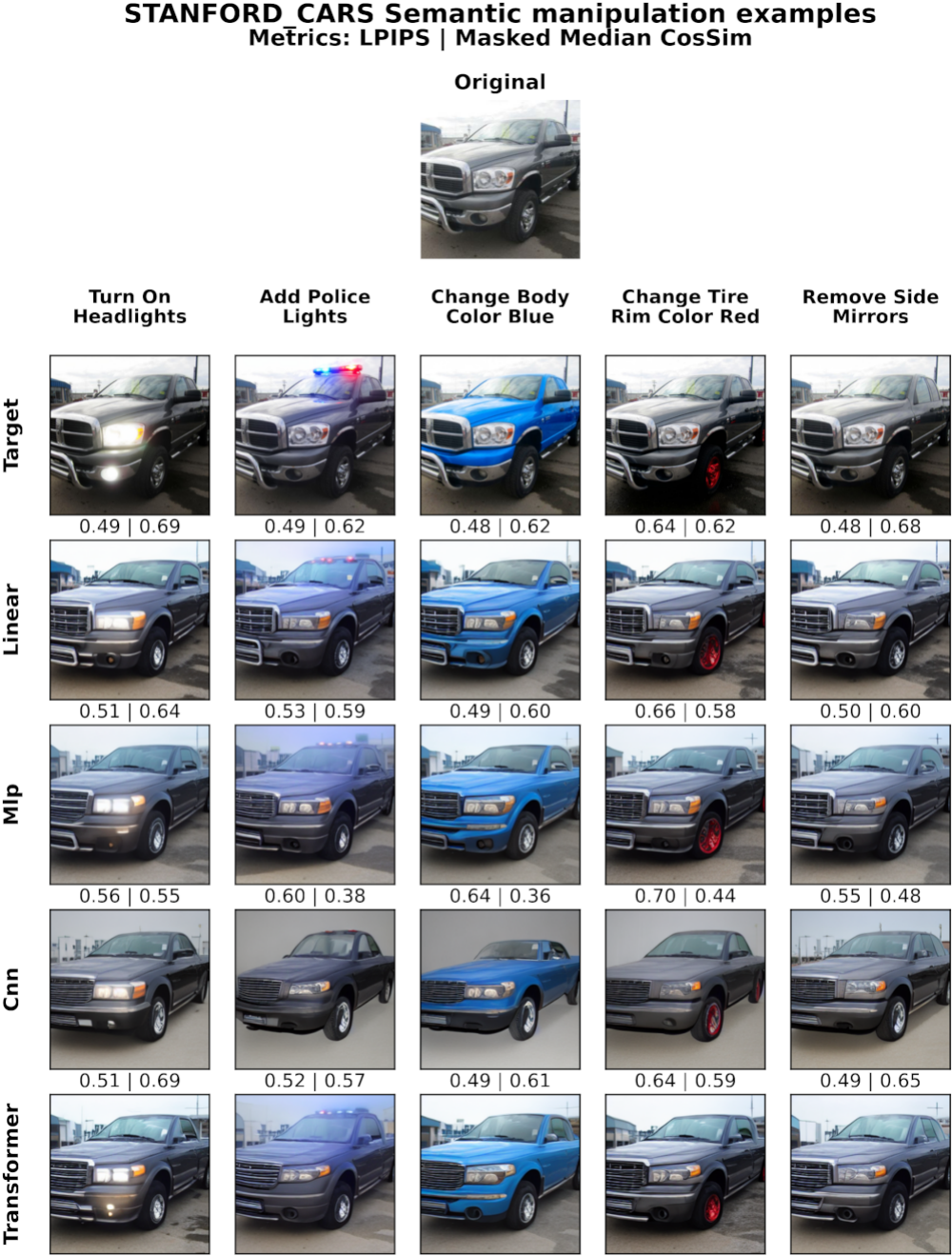}
    \caption{Example mapped and reconstructed images for each semantic manipulation type. They were created using features extracted from the last ConvNeXt layer \textbf{feat3} before the classification head. Target images were created using the Qwen image editing model. For each image the LPIPS, MdnCS and masked MdnCS metrics between target image and mapped, reconstructed image are shown.}
    \label{fig:semantic_manipulations_id_3895}
\end{figure}

Fig.~\ref{fig:semantic_manipulations_id_3895} shows more examples of semantic manipulation and model combinations on the same original image for comparison using features extracted from the last ConvNeXt layer feat3. The \emph{Headlights} manipulation was succesfully applied by all models with varying degrees of light intensity. Adding police lights proved to be another challenging task highly dependent on the car model and background. Semantic localized coloring manipulations (\emph{Change body color blue}, \emph{Change tire rim color red}) were possible to be applied with all mapping models. The \emph{Remove side mirrors} manipulation was again highly dependent on the car type, but generally successful in reducing the size or completely removing the side mirrors. 

\FloatBarrier

\begin{figure}[!htbp]
    \centering
    \includegraphics[width=1\linewidth]{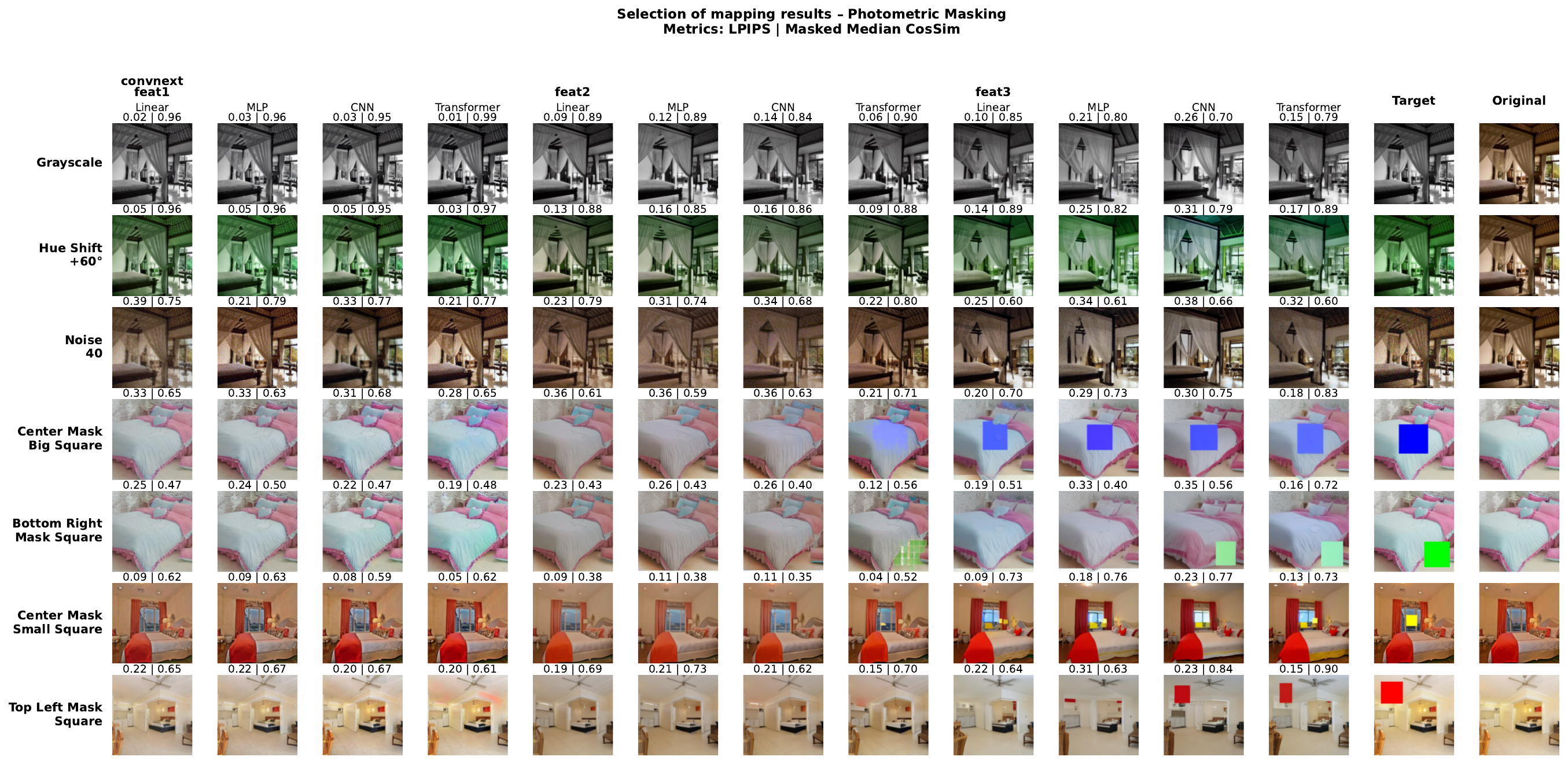}
    \caption{Example mapped and reconstructed images for each photometric and masking manipulation type. They were created using features extracted from the all ConvNeXt layers. Target images were created using the Qwen image editing model. For each image the LPIPS and masked MdnCS metrics between target image and mapped, reconstructed image are shown.}
    \label{fig:mapping_samples_convnext_photometric_masking}
\end{figure}

\begin{figure}[!htbp]
    \centering
    \includegraphics[width=1\linewidth]{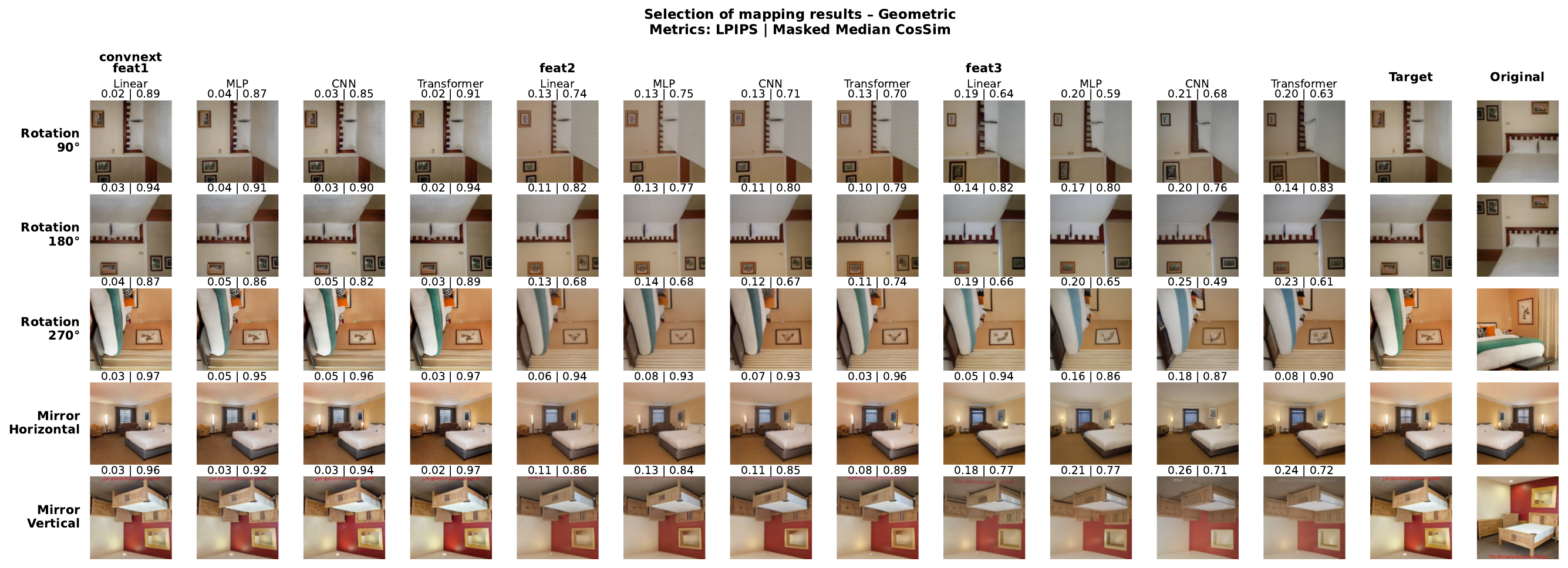}
    \caption{Example mapped and reconstructed images for each geometric manipulation type. They were created using features extracted from the all ConvNeXt layers. Target images were created using the Qwen image editing model. For each image the LPIPS and masked MdnCS metrics between target image and mapped, reconstructed image are shown.}
    \label{fig:mapping_samples_convnext_geometric}
\end{figure}

\begin{figure}[!htbp]
    \centering
    \includegraphics[width=1\linewidth]{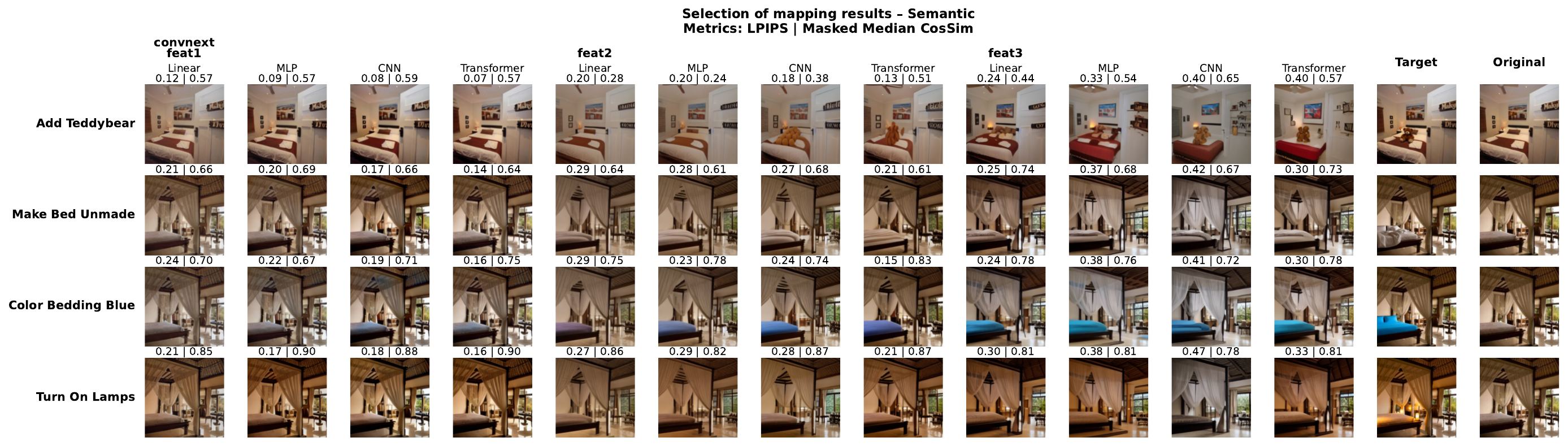}
    \caption{Example mapped and reconstructed images for each semantic manipulation type. They were created using features extracted from the all ConvNeXt layers. Target images were created using the Qwen image editing model. For each image the LPIPS and masked MdnCS metrics between target image and mapped, reconstructed image are shown.}
    \label{fig:mapping_samples_convnext_semantic}
\end{figure}

\begin{figure}[ht]
    \centering
    \includegraphics[width=1\linewidth]{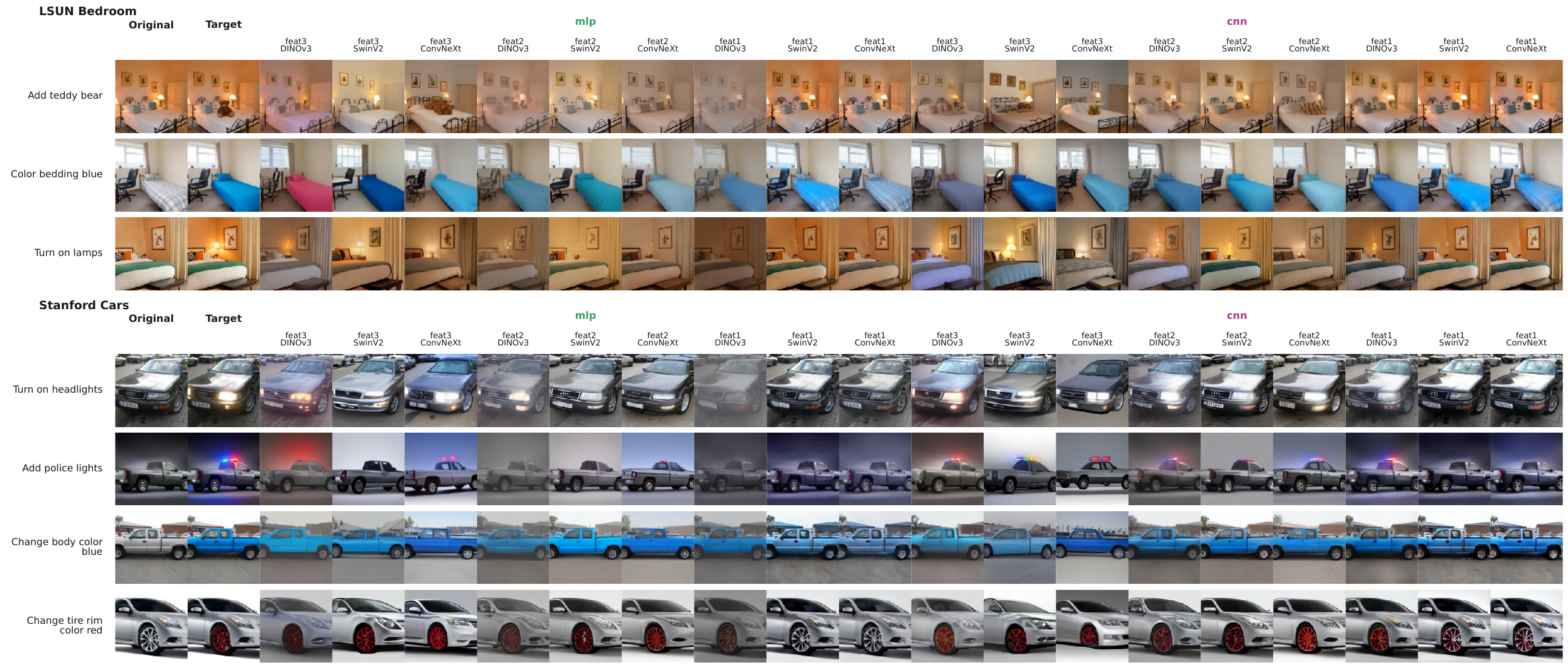}
    \caption{Examples of semantic manipulations on LSUN Bedroom and Stanford Cars. Each row shows the original, target, and mapped reconstructions obtained with MLP and CNN mappings from three feature stages of DINOv3, SwinV2, and ConvNeXt.}
    \label{fig:semantic_manipulations_main_mlp_cnn}
\end{figure}

\paragraph{Lsun Bedroom}
Figs.~\ref{fig:mapping_samples_convnext_photometric_masking}--\ref{fig:mapping_samples_convnext_semantic} present additional ConvNeXt experiments on the LSUN Bedroom dataset, covering photometric, masking, geometric, and semantic manipulations across all feature stages. Fig.~\ref{fig:semantic_manipulations_main_mlp_cnn} provides a complementary comparison of MLP and CNN mappings across backbones and feature stages on
LSUN Bedroom and Stanford Cars. Overall, mappings yield substantially higher reconstruction quality on LSUN Bedroom than on the other datasets, particularly for earlier ConvNeXt layers. Despite this improvement in image quality, the successes and failures of the manipulations are largely consistent with those observed in the other experiments. Both the linear and transformer models reliably apply the \emph{Color Bedding Blue} and \emph{Turn on Lamps} manipulations, correctly identifying the semantic boundaries of the target objects. For more open-ended manipulations, such as \emph{Make Bed Unmade}, the magnitude of the changes produced by the mappings does not always match that of the Qwen Image-Edit outputs. This variation is
unsurprising given the diversity of the training data and does not necessarily indicate semantic failure. The \emph{Add Teddybear} manipulation is generally applied at the correct position and with the correct color, although neither model consistently reproduces the target shape. The transformer more frequently produces the expected proportions, but still lacks fine-grained details.

The results in Fig.~\ref{fig:semantic_manipulations_main_mlp_cnn} are consistent with those obtained using the linear and transformer mappings. The same manipulations generally succeed or fail across mapping architectures, suggesting that outcomes are influenced more strongly by the underlying backbone features than by the choice of mapping model.

\FloatBarrier

\begin{figure}[!htbp]
    \centering
    \includegraphics[width=0.95\linewidth]{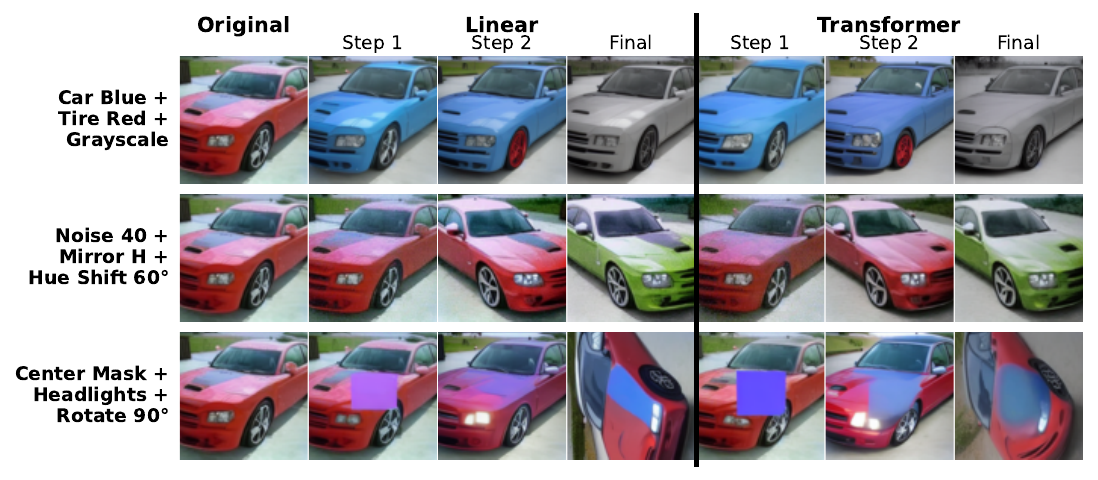}
    \caption{Three examples of reconstructed images obtained through sequential compositions of manipulation mappings for each model for feat3 ConvNeXt features for samples from the Lsun Bedroom dataset.}
    \label{fig:manip_compositions}
\end{figure}

\paragraph{Compositions of multiple mappings}
The examples in Fig.~\ref{fig:manip_compositions} show reconstructed images obtained from compositions of manipulations, where the output of one mapping model serves as the input to the next. In general, photometric and semantic manipulations remained applicable when applied sequentially. Across successive manipulations, object-specific details were largely preserved, indicating that the transformations did not substantially disrupt the semantic consistency of the underlying object. This suggests that the manipulations operate with limited interference and preserve core semantic structure even after repeated application. For manipulations with overlapping target regions (e.g., masking and headlights), both transformations were typically still realized, although often with some loss of fine-grained detail. By contrast, combinations of geometric manipulations and masking frequently introduced distortions, affecting either the masked region or the image more broadly. Both models displayed similar behavior across the tested manipulation combinations, with comparable successes and failures.

\begin{figure}[!htbp]
    \centering
    \includegraphics[width=0.95\linewidth]{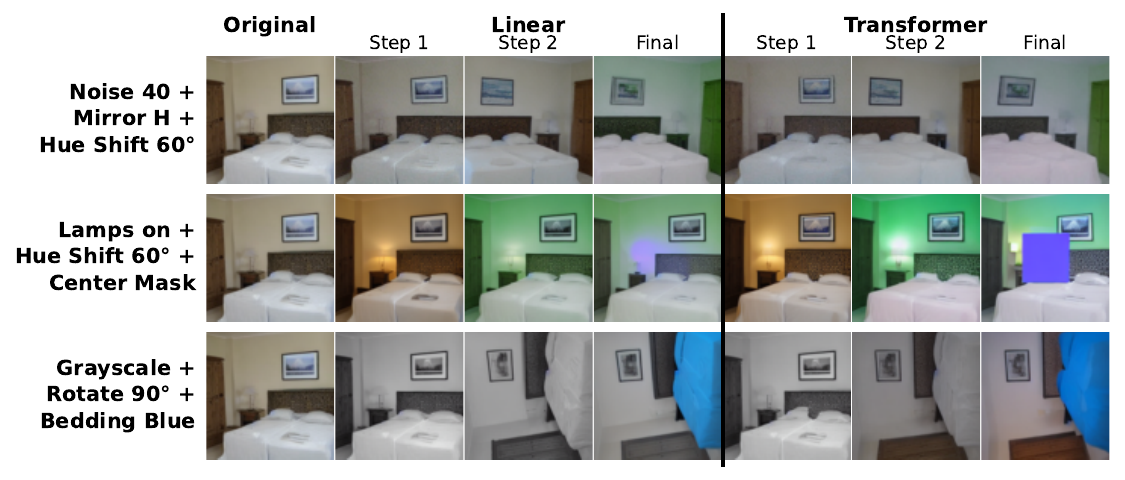}
    \caption{Three examples of reconstructed images obtained through sequential compositions of manipulation mappings for each model for feat3 ConvNeXt features for samples from the Lsun Bedroom dataset.}
    \label{fig:lsun_manip_compositions}
\end{figure}

Fig.~\ref{fig:lsun_manip_compositions} presents additional qualitative examples of composed mappings applied to samples from the LSUN Bedroom dataset. The reconstructions appear comparable in quality to those obtained in the Stanford Cars experiment and illustrate the potential of chaining different manipulation types. Consistent with the previous results, however, the linear mapping cannot reliably apply a masking manipulation. Notably, both mapping models successfully applied the \emph{Bedding Blue} manipulation after a \(90^\circ\) rotation, despite the absence of such rotations in the training set.

\begin{figure}[!htbp]
    \centering
    \includegraphics[width=0.8\linewidth]{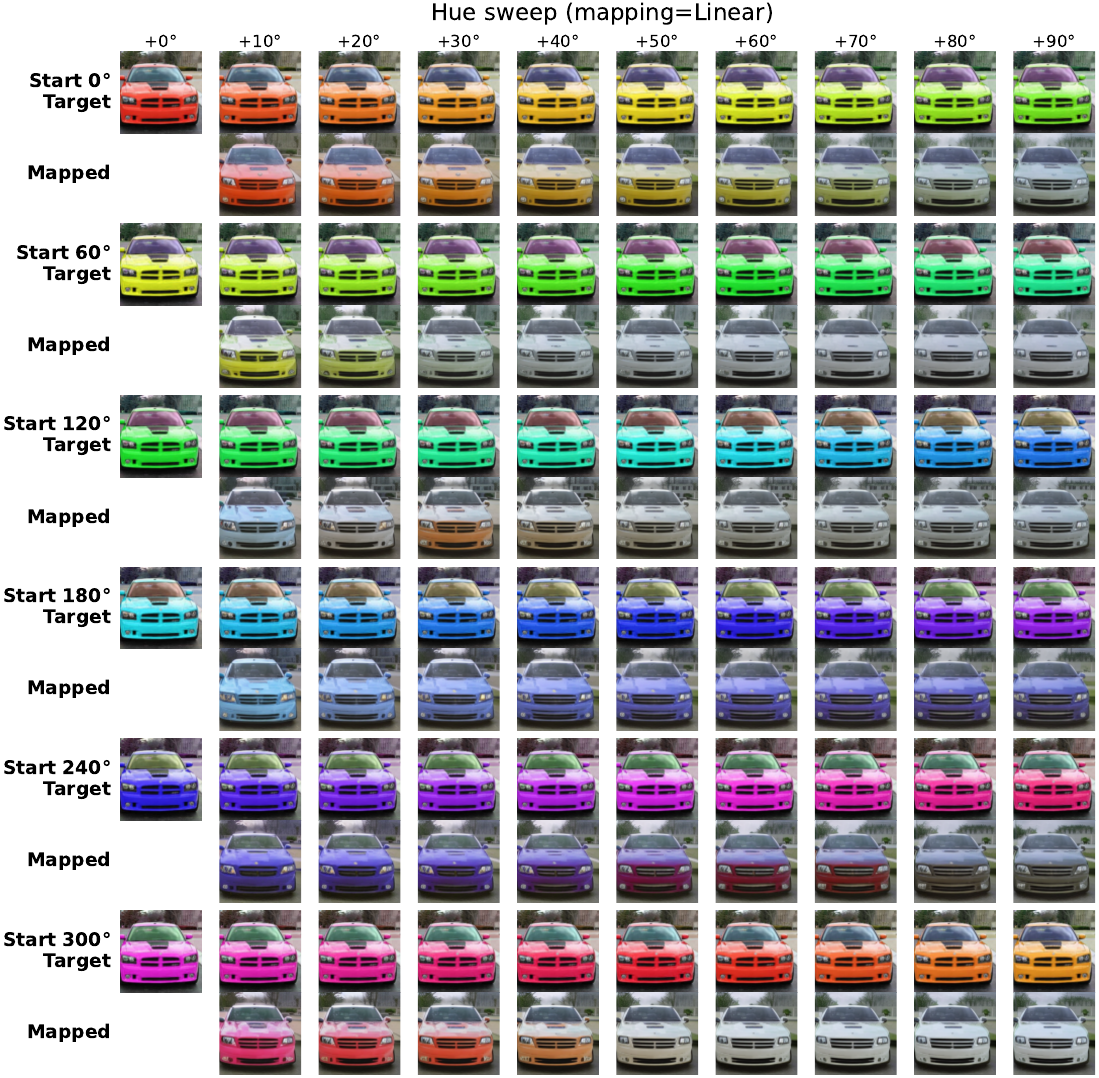}
    \caption{Qualitative comparison of target and recursively mapped hue transformations for a ConvNeXt feature representation. Each row starts from a different base hue, shown in the target row, and progresses in 10\textdegree~ increments from left to right. The target images depict the corresponding hue-shifted inputs, whereas the mapped images are obtained by repeatedly applying the learned linear feature-space mapping. }
    \label{fig:hue_sweep}
\end{figure}

To evaluate whether the linear mapping approach remains stable over longer repeated sequences, we trained the model on images with a 10\textdegree~ hue shift and then applied it recursively 36 times to traverse the full hue circle, starting from multiple different initial hues. Fig.~\ref{fig:hue_sweep} shows the first nine steps of this procedure for representative samples. In all cases, the mapping eventually failed after a varying number of steps, depending on the starting image. We attribute this breakdown to two main factors: (i) the training set did not cover all hues in the circle, leaving parts of the color space unseen during training; and (ii) errors accumulated with each recursive application, leading to progressive drift and instability.

\FloatBarrier

\begin{figure}[ht]
    \centering
    \includegraphics[width=0.9\linewidth]{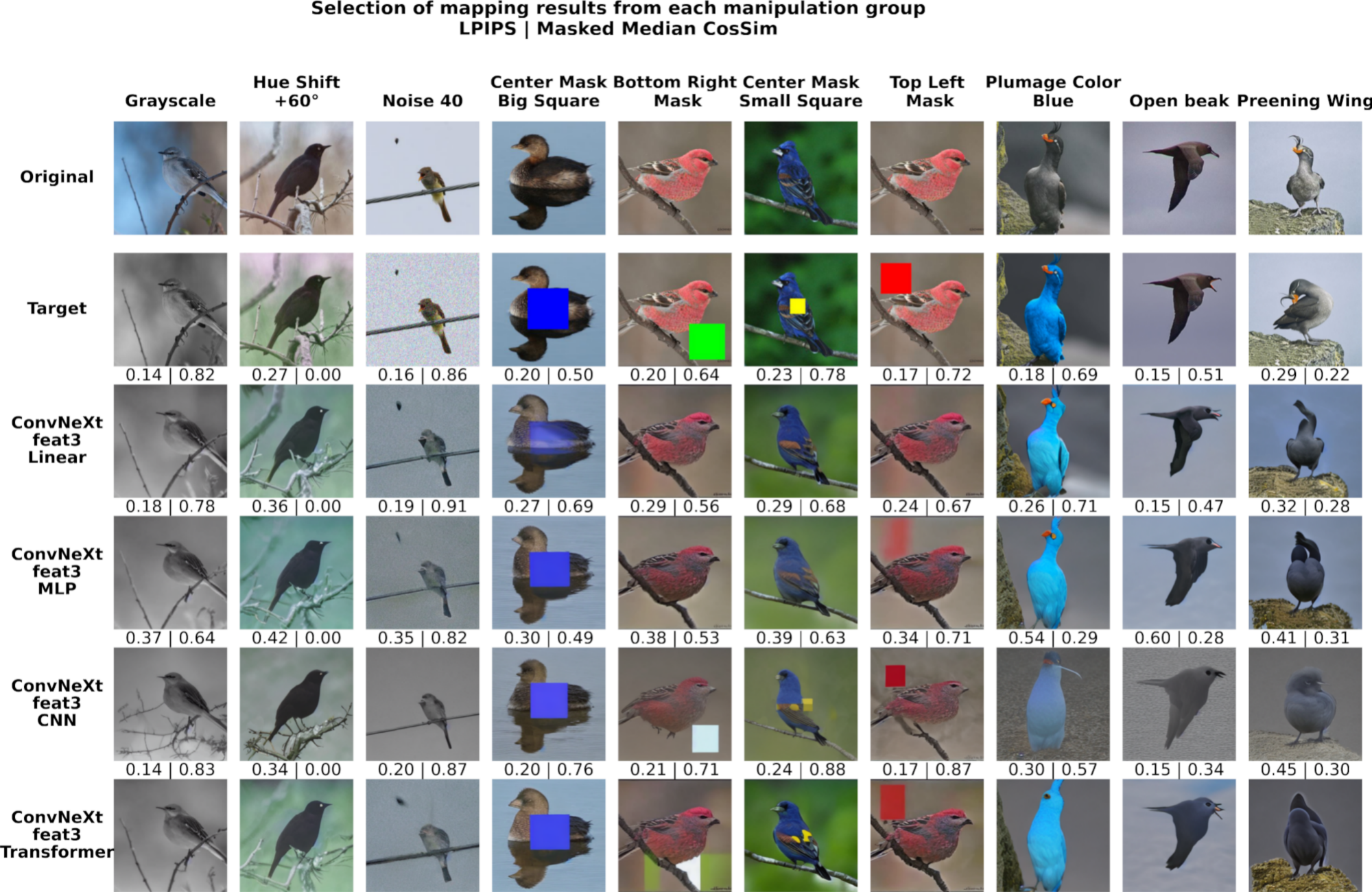}
    \caption{Example mapped and reconstructed images for selected manipulation types. They were created using the last feature stage from ConvNeXt. For each image the LPIPS and masked median MdnCS metrics between target image and mapped, reconstructed image are shown.}
    \label{fig:lin_tf_map_samples_birds}
\end{figure}

\begin{figure}[ht]
    \centering
    \includegraphics[width=0.8\linewidth]{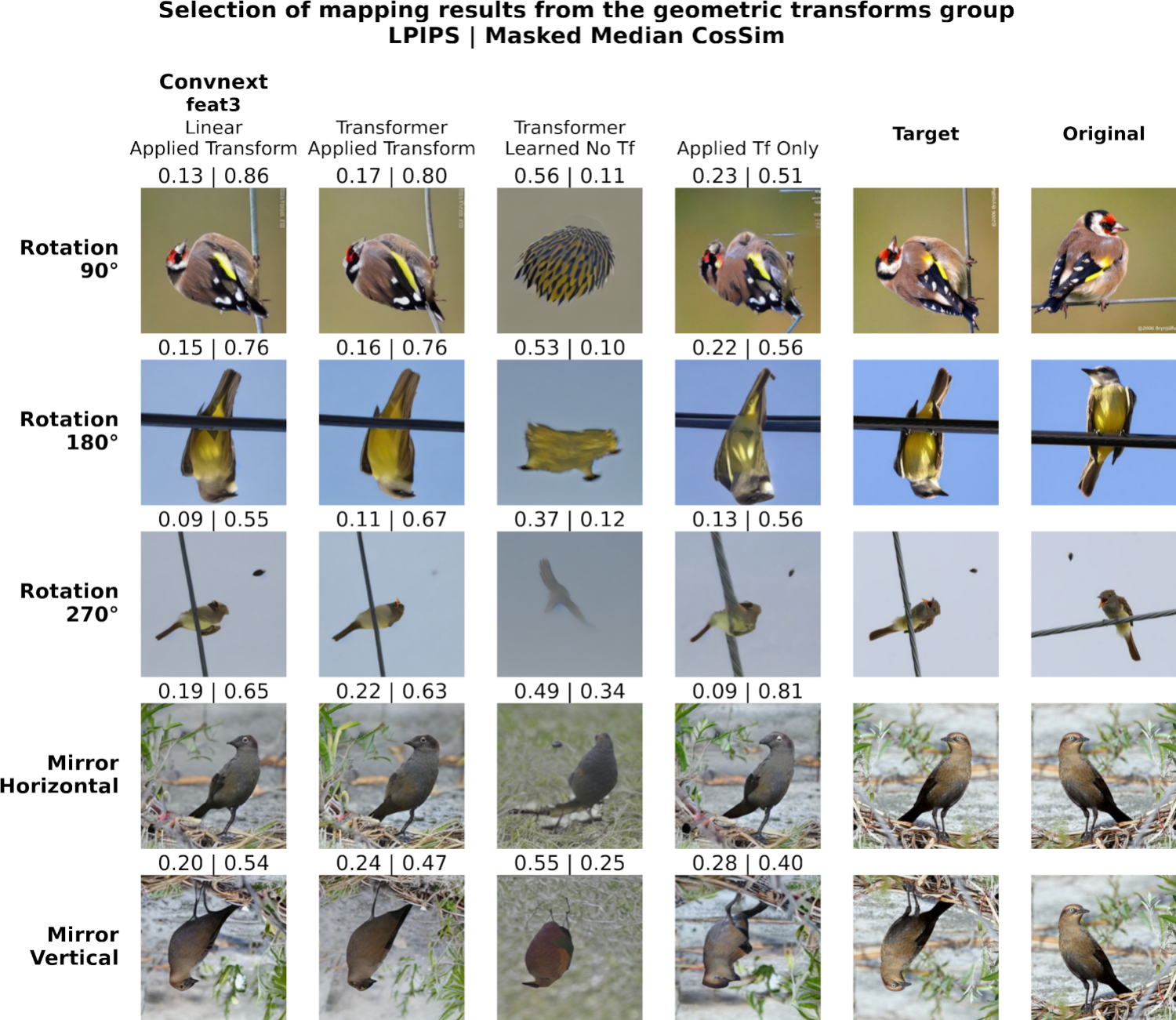}
    \caption{Example mapped and reconstructed images for selected geometric manipulations. They were created using the last feature stage from ConvNeXt. For each image the LPIPS and masked median MdnCS metrics between target image and mapped, reconstructed image are shown.}
    \label{fig:geom_samples_birds}
\end{figure}

\paragraph{Validation of our methods with the CUB\_200\_2011 birds dataset}
We conducted additional experiments on the CUB-200-2011 birds dataset \cite{WahCUB_200_2011} to evaluate how well our methods generalize to a different image domain. Figs.~\ref{fig:lin_tf_map_samples_birds} and \ref{fig:geom_samples_birds} show representative post-reconstruction mapping results across the tested mapping models and manipulation types. Overall, we observe trends similar to those seen on the Stanford Cars dataset: photometric and geometric manipulations transfer reliably, while masking remains comparatively more challenging. Applying semantic manipulations in this setting introduces additional complexity due to the wide variability in bird species and the presence of varied and detailed backgrounds. Among the evaluated operations, plumage color modification proved to be the most consistent, although it still depends on accurate localization of relevant object regions; in most cases, sensitive areas such as the beak and eyes were appropriately preserved. Other semantic manipulations showed more variability in their outcomes, which can be attributed in part to factors such as partial object visibility. Furthermore, semantic manipulations in this domain are inherently more open-ended and less tightly defined than the corresponding target attributes, which makes direct comparison to the targets generated by the Qwen image editing model less straightforward.

\begin{figure}[ht]
    \centering
    \includegraphics[width=0.8\linewidth]{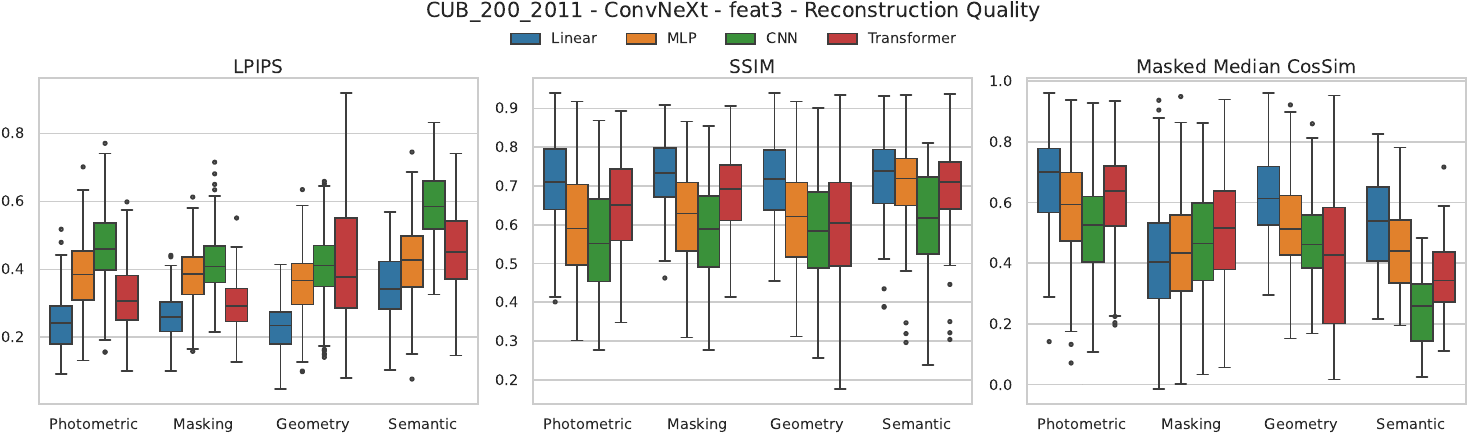}
    \caption{Reconstruction quality for the different manipulation groups for the CUB\_200\_2011 dataset.}
    \label{fig:reconstruction_quality_birds}
\end{figure}
\FloatBarrier

\subsection{Ablation on additional mapping model options}
\label{subsec:Ablation_new_models}
\paragraph{Additive and non-shared linear mapping} We add two diagnostic baselines to the existing spatially shared linear mapping, which applies the same affine transformation \(Wx+b\) to every feature vector in the spatial feature map. The first is an additive mapping, \(\hat{y}=x+\delta\), where \(\delta\) is a learned global offset that tests whether the representation change can be explained by a simple translation. The second is a spatially non-shared linear mapping, \(\hat{y}_{h,w}=W_{h,w}x_{h,w}+b_{h,w}\), which learns a separate affine transformation for each spatial position. Although this local model is more expressive, its parameter count, memory requirements, and computational cost per optimization step scale linearly with the number of spatial feature vectors \(H \cdot W\). Moreover, for spatially non-aligned input data it is questionable whether learning a position-specific mapping is an appropriate design choice. Comparing these models with the shared linear mapping therefore separates additive from matrix-based transformations while testing whether spatial weight sharing sacrifices predictive accuracy in exchange for improved efficiency and interpretability. Although the additional flexibility of the position-dependent mapping does not necessarily require more training epochs to reach a low training loss, it makes each epoch more expensive and becomes increasingly inefficient as the spatial resolution increases.

\paragraph{Procrustes and ridge regression} In addition, we include two closed-form baselines that operate on flattened feature vectors and apply the same transformation at every spatial location. The orthogonal Procrustes mapping solves \(\min_{R} \|XR - Y\|_F\) subject to \(R^\top R = I\) via a single SVD of the cross-covariance \(X^\top Y\), yielding an orthogonal linear map \(\hat{y} = xR\) with no learnable parameters. The ridge regression mapping fits an affine transform \(\hat{y} = (x - \bar{x})W + \bar{y}\) by solving the regularized normal equations \((X_c^\top X_c + \gamma I)W = X_c^\top Y_c\), where \(\gamma=1\) controls the strength of \(\ell_2\) regularization. Both methods are computed in closed form and share weights across spatial positions.

\begin{figure}[!htbp]
    \centering
    \includegraphics[width=0.8\linewidth]{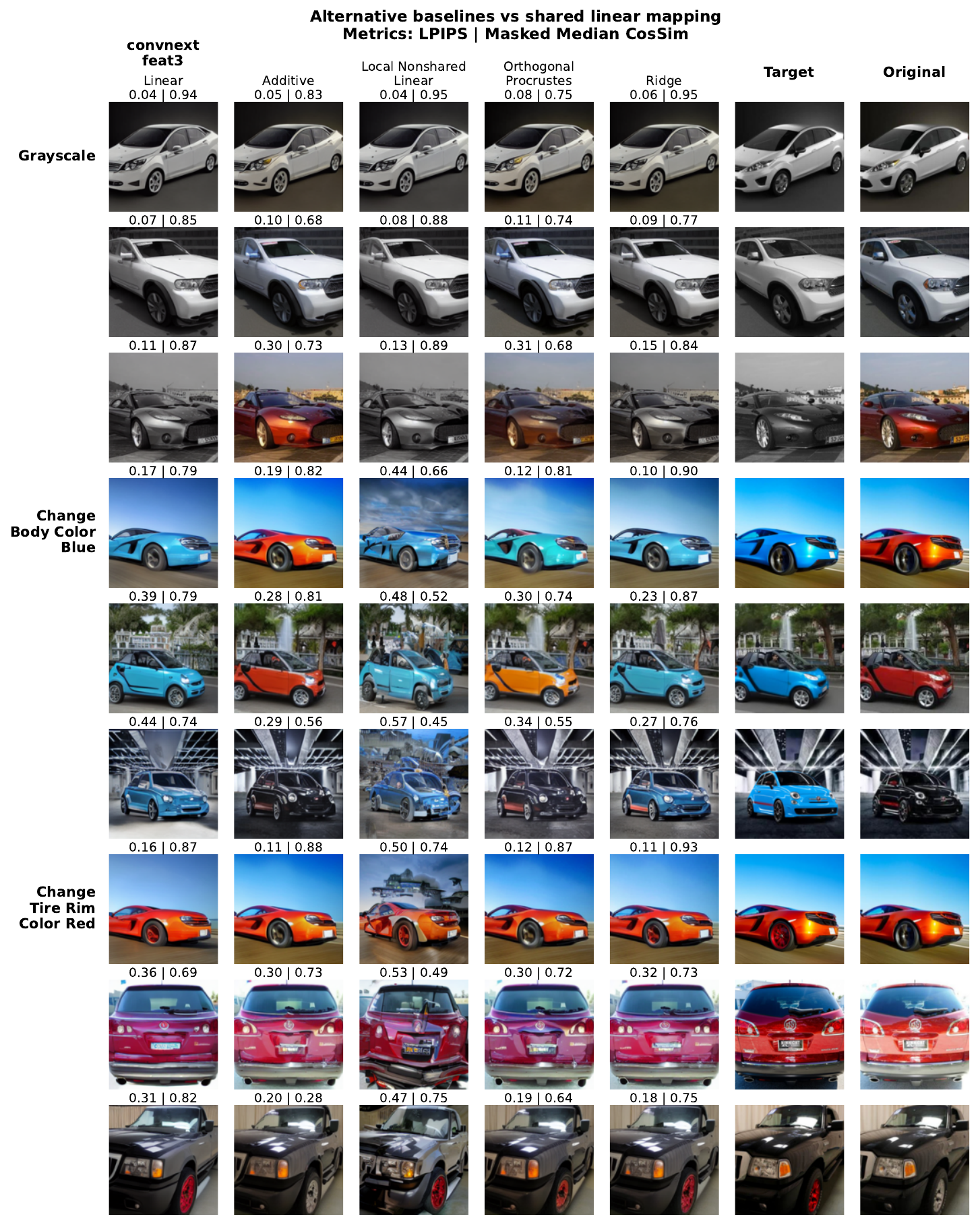}
    \caption{Example mapped and reconstructed images for selected manipulation types using the additive, local non-shared linear, orthogonal Procrustes, and ridge regression baselines compared to the shared linear mapping.}
\label{fig:mapping_samples_convnext_baselines_vs_shared_weight_linear}
\end{figure}

\paragraph{Results: Additive and non-shared linear mapping} As illustrated in Fig.~\ref{fig:mapping_samples_convnext_baselines_vs_shared_weight_linear}, the additive mapping fails to reproduce the target representations in several manipulation types, indicating that a global offset alone cannot capture the required feature-space transformation. The spatially non-shared linear mapping, while able to match the target manipulations in all cases and thus ``understand'' the manipulations as well as the shared model, yields noticeably lower image quality and spatial coherence: adjacent spatial locations often appear misaligned. Although the global loss (MSE + cosine similarity over the full image) in principle allows the model to learn consistent behavior, there is no explicit inductive bias or parameter sharing that enforces smoothness across neighboring positions. Consequently, the optimizer tends to find solutions that fit the target features well but exhibit local inconsistencies between adjacent feature vectors. This degradation is most pronounced for object- or location-specific manipulations (e.g., changing the color of a specific part), which exhibit structured artifacts and local warping, whereas global manipulations such as \emph{Grayscale} remain visually acceptable. This behavior is consistent with the fact that each position-specific linear map is optimized independently, without any constraint enforcing spatial consistency across neighboring locations. 

\paragraph{Results: Procrustes and ridge regression} In contrast, the two closed-form baselines show distinct behavior: the orthogonal Procrustes mapping exhibits high variance even within the same manipulation type—succeeding on some samples while failing completely on others from the same class—suggesting that its orthogonal constraint is too restrictive for the heterogeneous transformations induced by the manipulations. When it does succeed, however, it produces reconstructions of comparable quality to the learned models. Ridge regression, by contrast, performs consistently and competitively with the spatially shared linear model across all manipulation types and samples, reliably producing high-quality reconstructions despite having no iterative training. These results suggest that neither a purely additive transformation nor an unconstrained per-position linear map provides an adequate trade-off between fidelity, efficiency, and interpretability, while the strong and stable performance of ridge regression indicates that a simple, closed-form affine map with \(\ell_2\)-regularization can match learned methods in most cases. This motivates our choice of a spatially shared linear mapping for the main experiments, which offers comparable performance to ridge regression while retaining the flexibility of end-to-end optimization and task-specific loss functions.

\FloatBarrier

\subsection{Effect of Loss Function choice on the mapping quality}
\label{subsec:lossfunction_choice}
\begin{figure}[!htbp]
    \centering
    \includegraphics[width=0.9\linewidth]{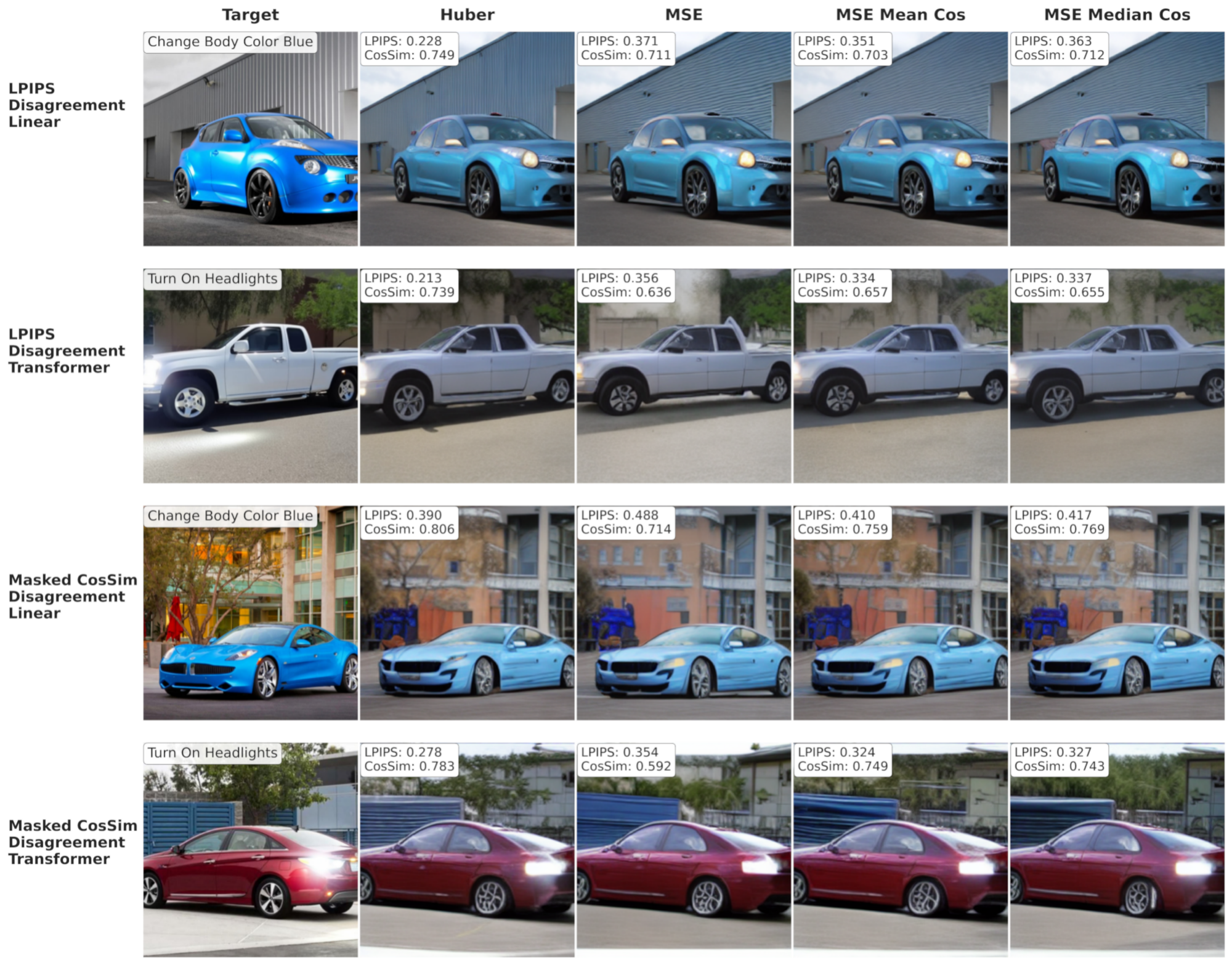}
    \caption{Selected mapping examples produced by models trained with
    different loss functions.}
    \label{fig:loss_function_img_comparison}
\end{figure}

This section evaluates alternative loss functions for the two manipulations
\emph{Change car body blue} and \emph{Turn on headlights}. Further details
on the loss functions are provided in the main text.
Fig.~\ref{fig:loss_function_img_comparison} shows selected reconstructions
produced by models trained with the four loss functions.

Across all examples, the requested semantic manipulations are successfully
performed. The main qualitative difference concerns the preservation of
background details. Models trained with MSE alone more frequently remove or
degrade background structure, although the magnitude of this effect depends on
the input scene. Apart from this tendency, the differences between the loss
functions are not visually apparent in the examples shown. In particular, we
observe no visible difference in the semantic correctness of the
manipulations. The transformer mappings exhibit larger numerical differences
between loss functions than the linear mappings, with differences of up to
approximately 0.2 in masked MdnCS. These numerical differences do not,
however, correspond to a clearly visible or semantically distinguishable
improvement in the reconstructed images. Although the Huber loss generally
produces the best reconstruction metrics, its advantage is not apparent in
terms of semantic fidelity.

\begin{figure}[!htbp]
    \centering
    \includegraphics[width=0.7\linewidth]{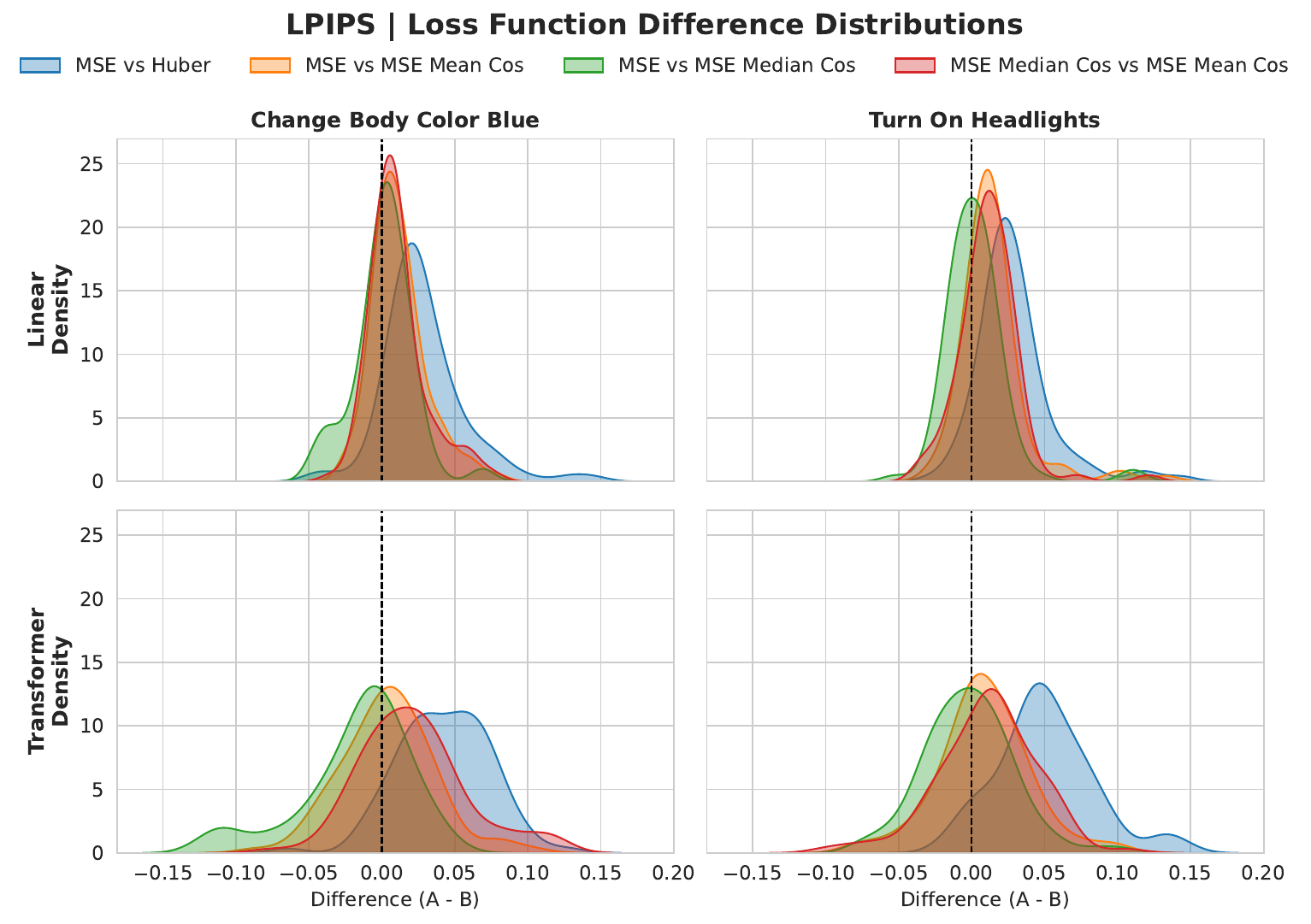}
    \caption{Impact of Loss Function choice on post-reconstruction mapping performance LPIPS for the Stanford Cars dataset. For each comparison, the plotted quantity $\Delta$ denotes the difference in
reconstruction quality between two loss functions evaluated on the same input pair. The vertical dashed line at $\Delta=0$ denotes equal performance.}
    \label{fig:loss_function_comparison_LPIPS}
\end{figure}

\begin{figure}[!htbp]
    \centering
    \includegraphics[width=0.78\linewidth]{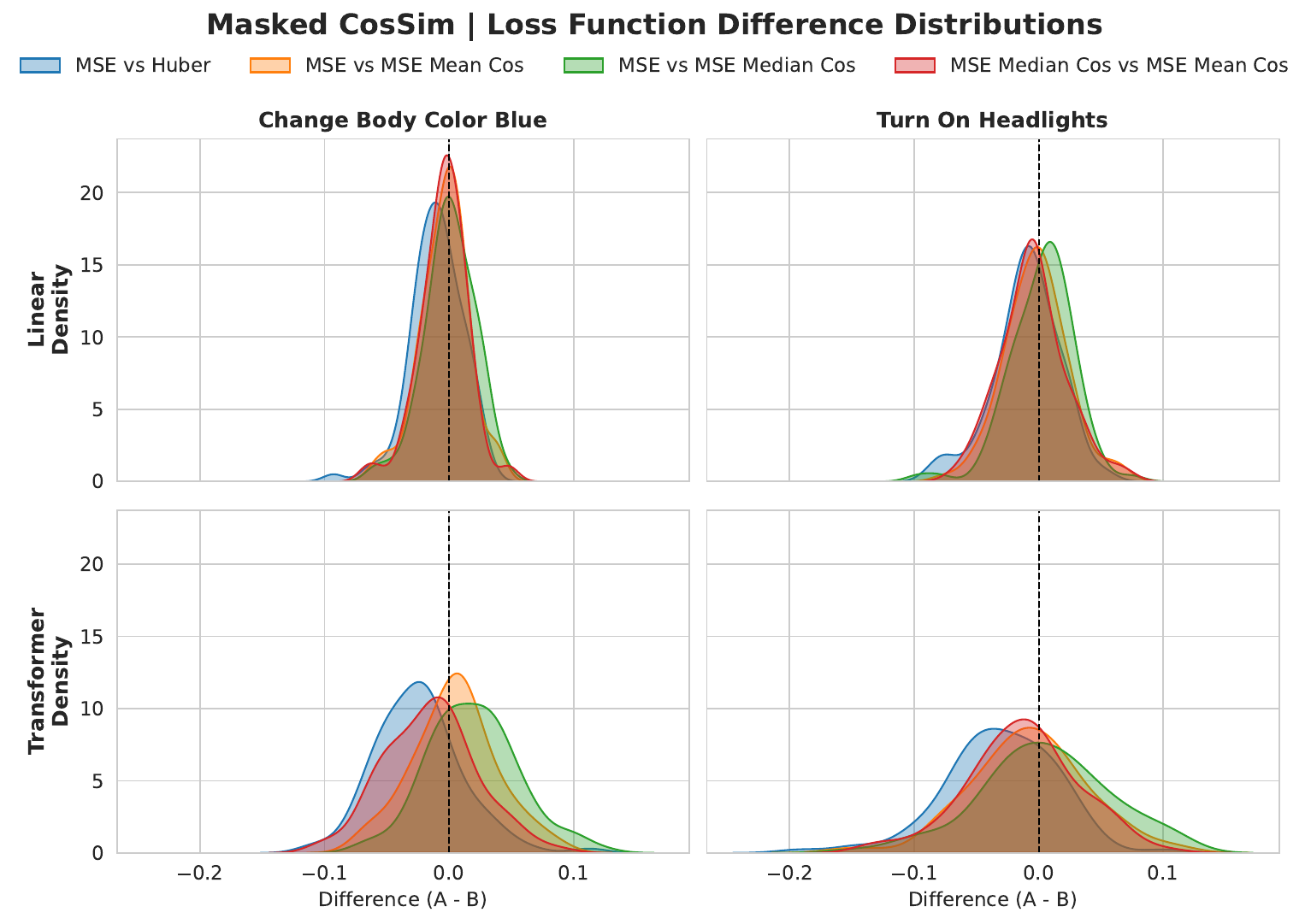}
    \caption{Impact of Loss Function choice on post-reconstruction mapping performance masked MdnCS for the Stanford Cars dataset.  For each comparison, the plotted quantity $\Delta$ denotes the difference in
reconstruction quality between two loss functions evaluated on the same input pair. The vertical dashed line at $\Delta=0$ denotes equal performance.}
    \label{fig:loss_function_comparison_Masked}
\end{figure}

To further investigate the effect of the loss-function choice, Figs.~\ref{fig:loss_function_comparison_LPIPS} and \ref{fig:loss_function_comparison_Masked} show the distributions of paired performance differences in LPIPS and masked MdnCS across matched samples,
separated by target manipulation and backbone architecture.

Replacing the standard MSE objective with a Huber loss produces the largest and most consistent quantitative improvements. The strongest effect is observed for LPIPS, for which the Huber loss yields a relative reduction of approximately 10--12\%. Masked cosine similarity also improves in the same direction, but to a smaller extent of approximately 2--4\%. Paired$t$-tests indicate statistically significant differences for all evaluated metrics ($p<0.001$). These results indicate that the loss functions can be distinguished reliably by the quantitative metrics. However, the magnitude of the differences is small relative to the qualitative behavior of the reconstructions. The metric improvements do not result in a consistently
visible improvement in image quality or in a perceptible improvement in the semantic correctness of the manipulations.

Adding cosine similarity to the MSE objective generally improves the quantitative reconstruction metrics relative to using MSE alone, although the magnitude of the improvement depends on the manipulation and the metric. The MSE-only objective is the main qualitative exception because it more frequently produces reconstructions with missing or degraded background details. Overall, the choice of loss function has a measurable effect on the reported metrics, but only limited qualitative consequences for the visual and semantic properties of the mapped reconstructions.

\subsection{Reconstruction Robustness}
\label{app:featinv_robustness}

\begin{figure}[ht]
\centering
\includegraphics[width=0.6\linewidth]{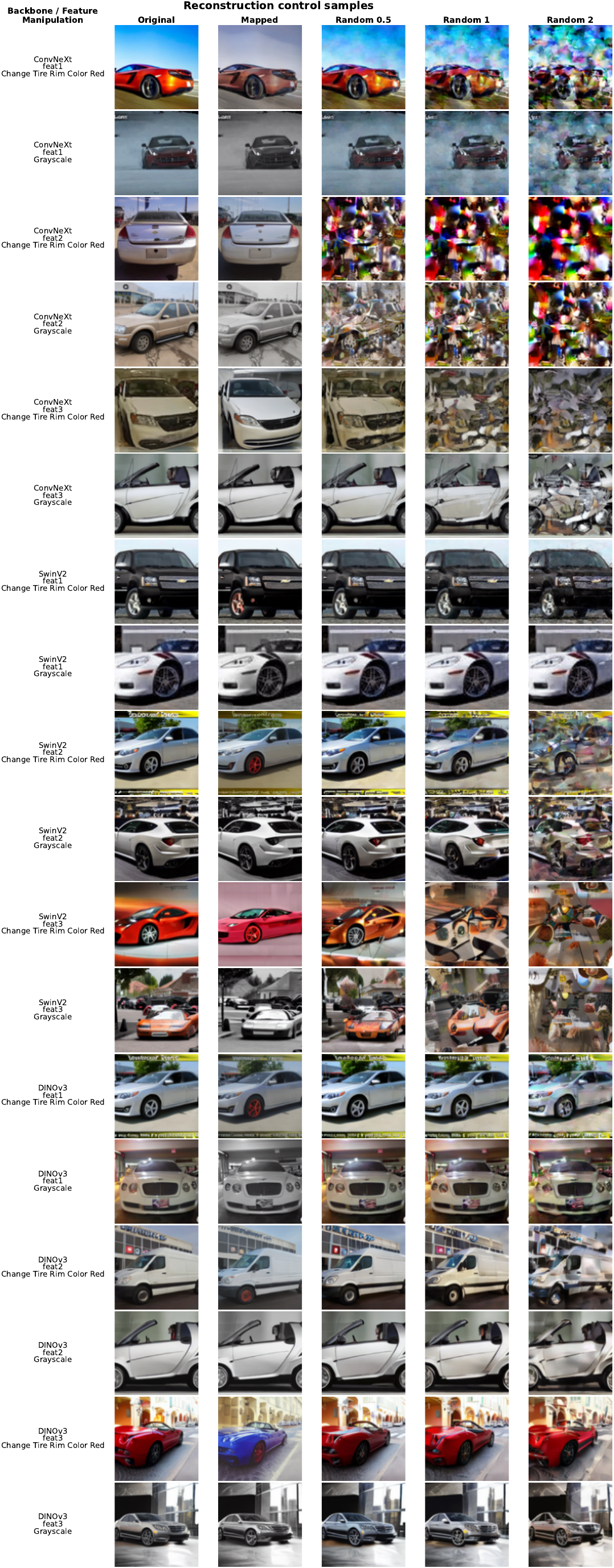}
\caption{Example reconstructions under matched-magnitude feature perturbations.}
\label{fig:recon_robustness_samples}
\end{figure}

\begin{figure}[ht]
\centering
\includegraphics[width=0.9\linewidth]{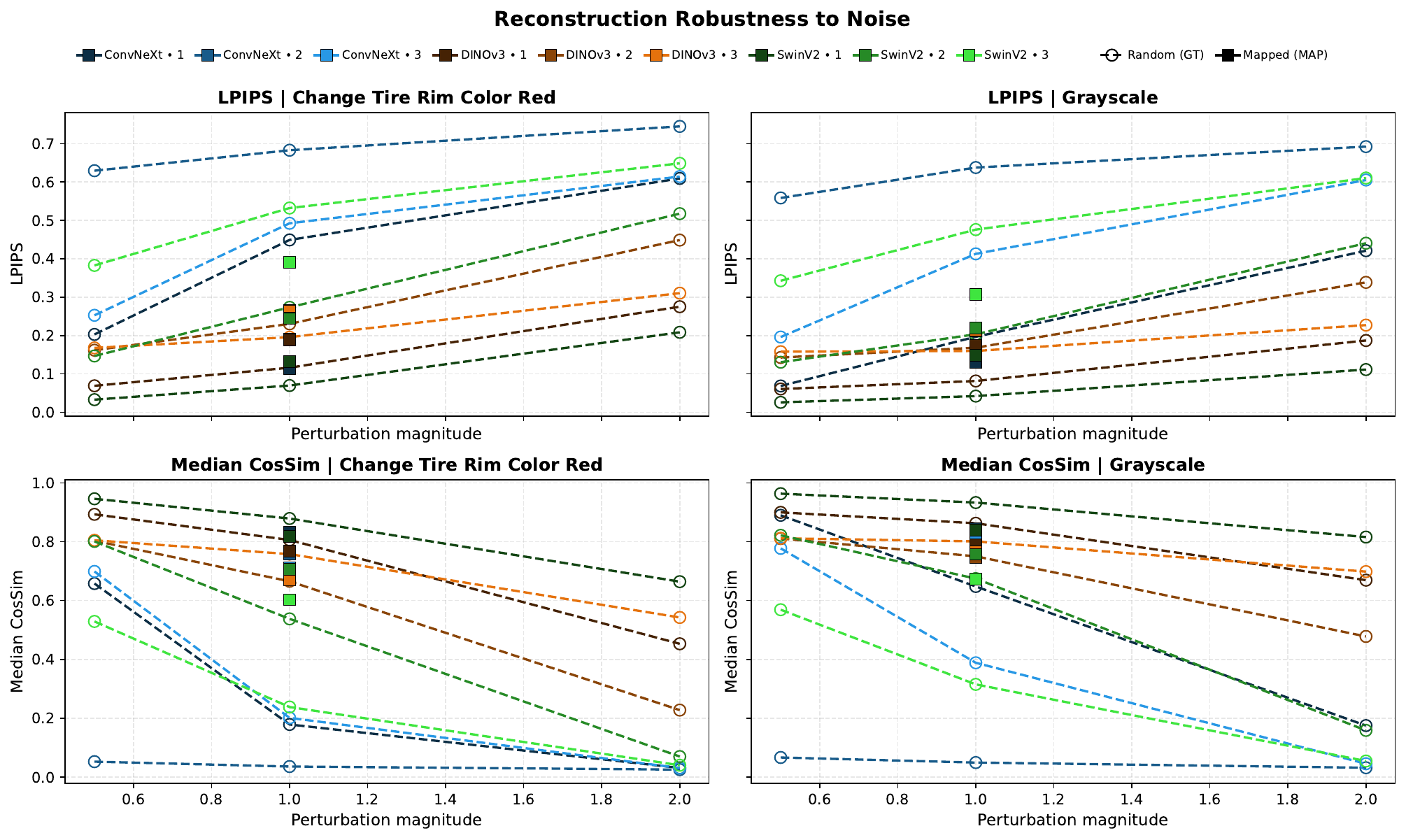}
\caption{Evaluation of reconstruction robustness using LPIPS and median cosine similarity (MdnCS).}
\label{fig:recon_robustness_metrics}
\end{figure}

In the FeatInv paper, robustness to feature-space noise was only assessed implicitly, via out-of-distribution (OOD) evaluation experiments. Here, we explicitly evaluate this robustness using a matched-magnitude perturbation control. For each layer, architecture, and manipulation, we first measure the empirical feature-space shift norm
\begin{equation}
    \delta=\|f(z)-z\|
\end{equation}
induced by the learned mapping model (treated as the ``ground-truth'' direction in feature space). We then generate random-direction perturbations with norm equal to $\delta$, as well as scaled versions ($0.5\delta$, $2\delta$). Both the learned-direction (mapped) perturbations and the random-direction perturbations are passed through the same FeatInv reconstruction pipeline, enabling a direct comparison of reconstruction quality for the learned mapping direction against random directions of matched and scaled magnitude. Example reconstructions are shown in Fig.~\ref{fig:recon_robustness_samples}; aggregate metrics are summarized in Fig.~\ref{fig:recon_robustness_metrics}.

\paragraph{Backbone-wise reconstruction statistics}
We quantify robustness by comparing learned-direction (mapped) perturbations at fixed strength~$\delta$ against random-direction perturbations at $0.5\delta$, $1.0\delta$, and $2.0\delta$, in terms of LPIPS and median cosine similarity (MdnCS; Fig.~\ref{fig:recon_robustness_metrics}). Across backbones, the benefit of using the learned direction differs systematically. For ConvNeXt, mapped perturbations consistently yield better reconstructions than random perturbations at all noise levels (lower LPIPS and higher MdnCS for 0.5--2$\times$ random), indicating that the learned mapping direction is substantially more benign than arbitrary directions of matched or even larger norm. This effect is particularly pronounced for ConvNeXt feat2 features, which are highly sensitive to random perturbations (Fig.~\ref{fig:recon_robustness_samples}).

For DINOv3 and SwinV2, the pattern is more nuanced: at low noise (0.5$\times$), random perturbations are competitive with or slightly better than mapped perturbations (small negative impact in both LPIPS and MdnCS), whereas at medium to high noise (1--2$\times$) the learned direction becomes increasingly advantageous, with growing positive impact as noise strength increases. These results suggest that ConvNeXt representations are particularly sensitive to the direction of feature-space perturbations, whereas DINOv3 and SwinV2 exhibit more direction-robust behaviour at low magnitudes but still benefit from the learned mapping direction under stronger perturbations.

ConvNeXt and SwinV2 exhibit comparable robustness curves (noise strength vs.\ LPIPS and cosine similarity) at their respective top layers, supporting the feasibility of comparing these architectures at those layers. For cross-layer comparisons, reconstruction floors and sensitivity differ substantially by layer. Consequently, absolute metric comparisons across layers can be misleading; within-layer relative comparisons across mapping types are more informative. Importantly, the paper's central claim---that linearity increases with depth---relies on the gap between linear and nonlinear mapping-model performance at a given layer, not on absolute cross-layer comparability. This gap is more reliable because both the linear and nonlinear mappings are reconstructed using the same per-layer FeatInv model, so the layer-specific reconstruction floor is shared and largely cancels out in the comparison.

\FloatBarrier

\subsection{Bias analysis}
\label{sec:bias}
\begin{figure}[ht]
    \centering
    \includegraphics[width=0.65\linewidth]{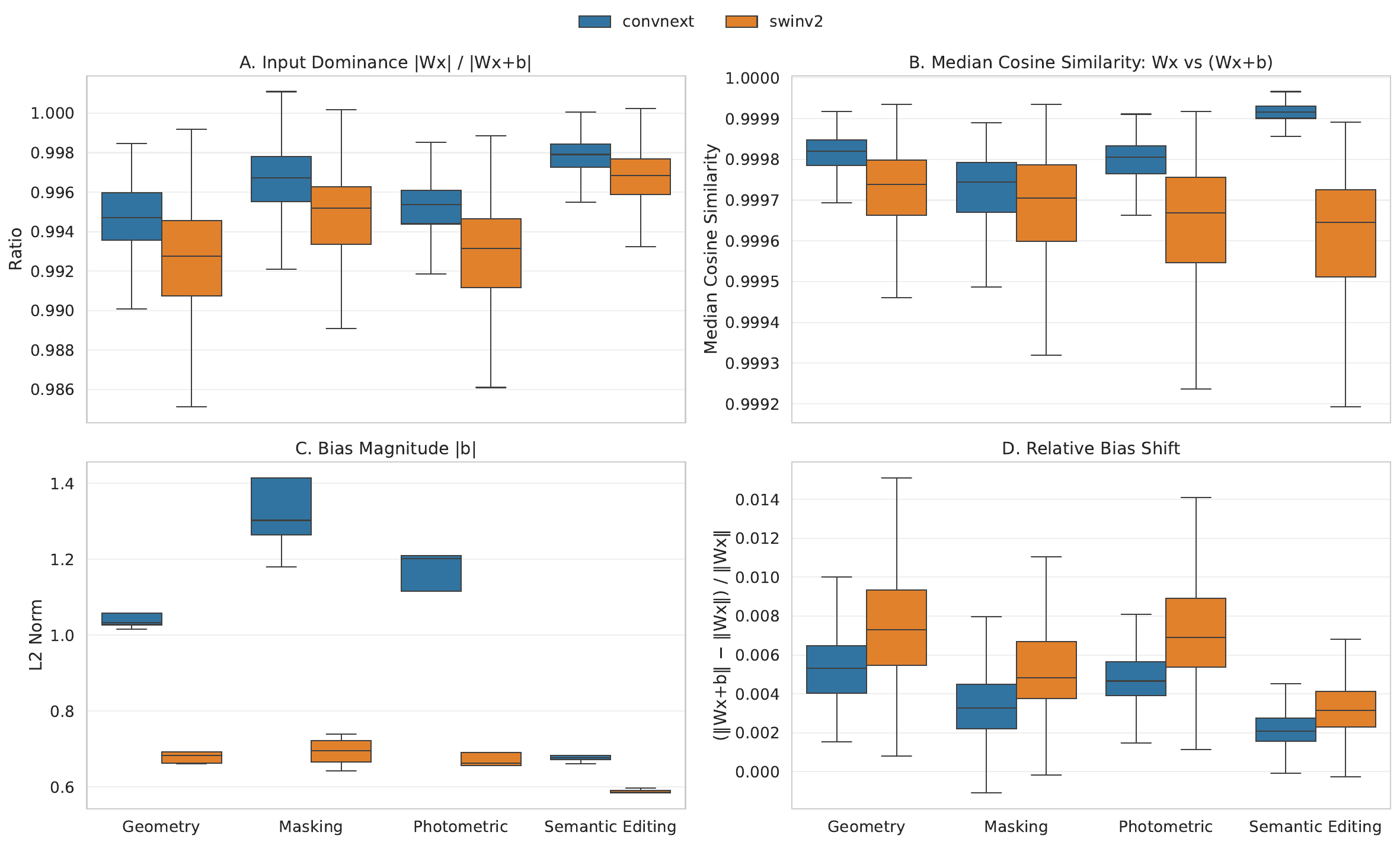}
    \caption{
        Bias dominance metrics across manipulation groups showing magnitude dominance, directional stability, and small relative bias effects.}
    \label{fig:wx_bias_eval}
\end{figure}

Fig.~\ref{fig:wx_bias_eval} shows comparisons between the weighted inputs $Wx$ and the bias term. 
\FloatBarrier

\subsection{SVD analysis over all feature depths}
\begin{figure}[ht]
    \centering
    \begin{subfigure}{0.48\linewidth}
        \centering
        \includegraphics[width=\linewidth]{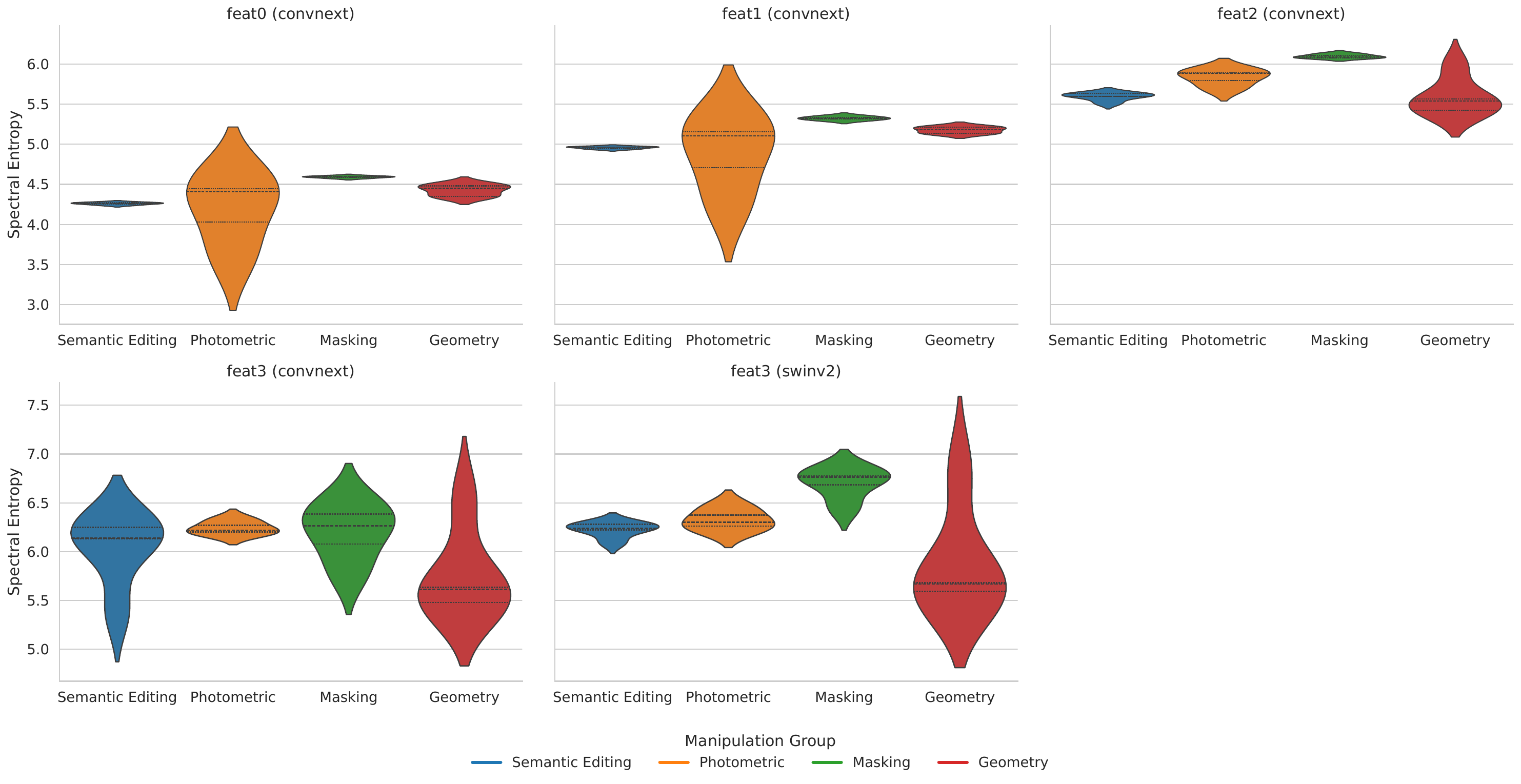}
        \caption{Distribution of spectral entropy for the different manipulation groups}
        \label{fig:entropy_rank}
    \end{subfigure}
    \hfill
    \begin{subfigure}{0.48\linewidth}
        \centering
        \includegraphics[width=\linewidth]{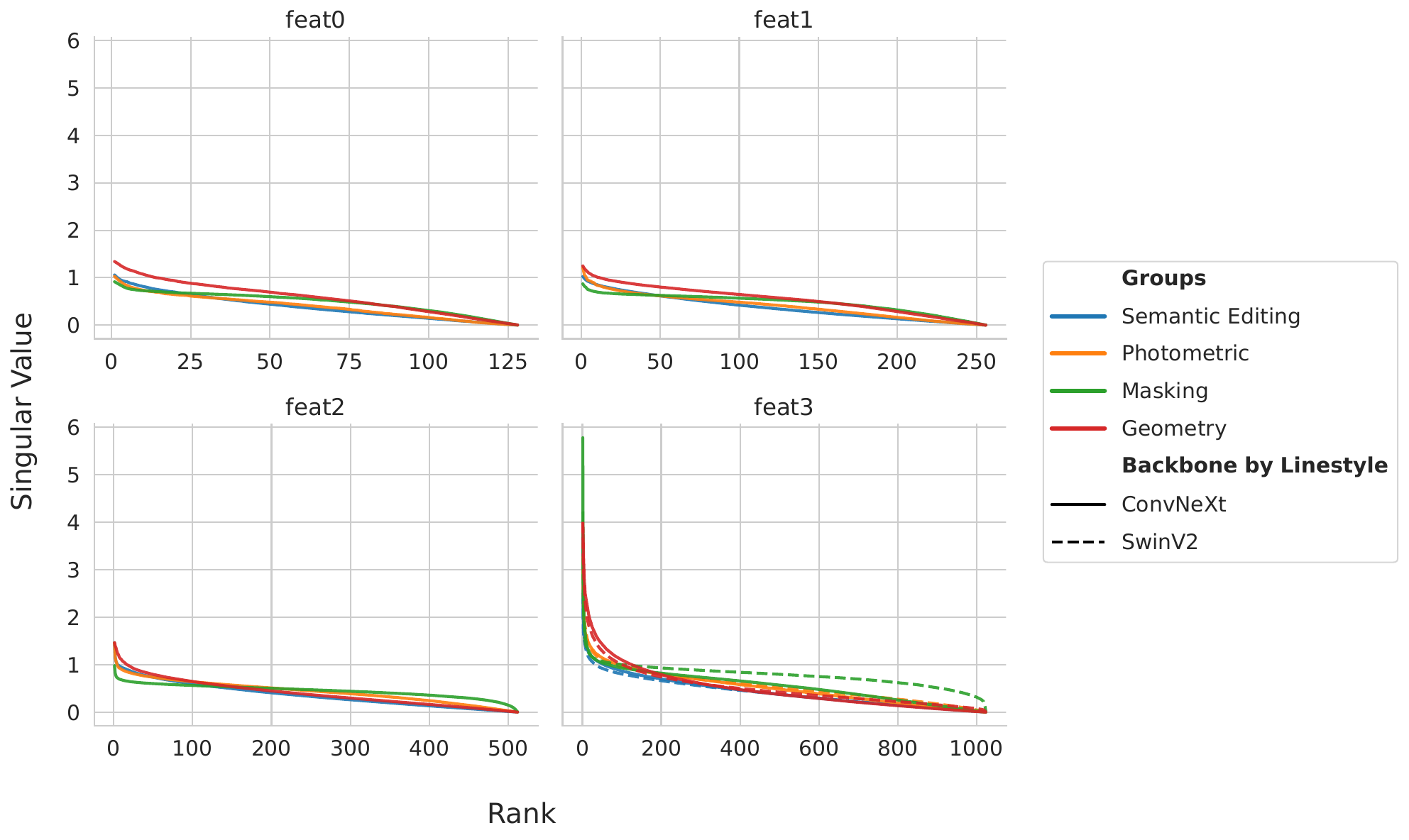}
        \caption{Singular value energy spectrum as a function of rank}
        \label{fig:svd_spectrum}
    \end{subfigure}
    \caption{Comparison of structural complexity and spectral decay across feature representations and image manipulations for the ConvNeXt backbone.}
    \label{fig:full_svd}
\end{figure}

Fig.~\ref{fig:full_svd} shows the SVD analysis over all feature depths of ConvNeXt and for the last feature layer for the SwinV2 backbone. 

\FloatBarrier

\subsection{Reconstruction quality metrics}
\label{subsec:app_metrics}

We evaluate the quality of the post-reconstruction mapping using three image-space metrics: LPIPS, structural similarity (SSIM), and mean squared error (MSE). These metrics quantify how closely the reconstructed target image matches the manipulated target image after mapping through the learned representation. Lower values indicate better reconstruction quality for allthree metrics.

For each metric, we report the mean value over all samples within a manipulation group. The results are stratified by dataset, manipulation group, backbone, feature layer, and mapping model. In each heatmap, rows correspond to backbone--model-type combinations, while columns correspond to the feature layers \textit{Feat1}--\textit{Feat3}. 

\begin{figure}[ht]
    \centering
   \includegraphics[width=\linewidth]{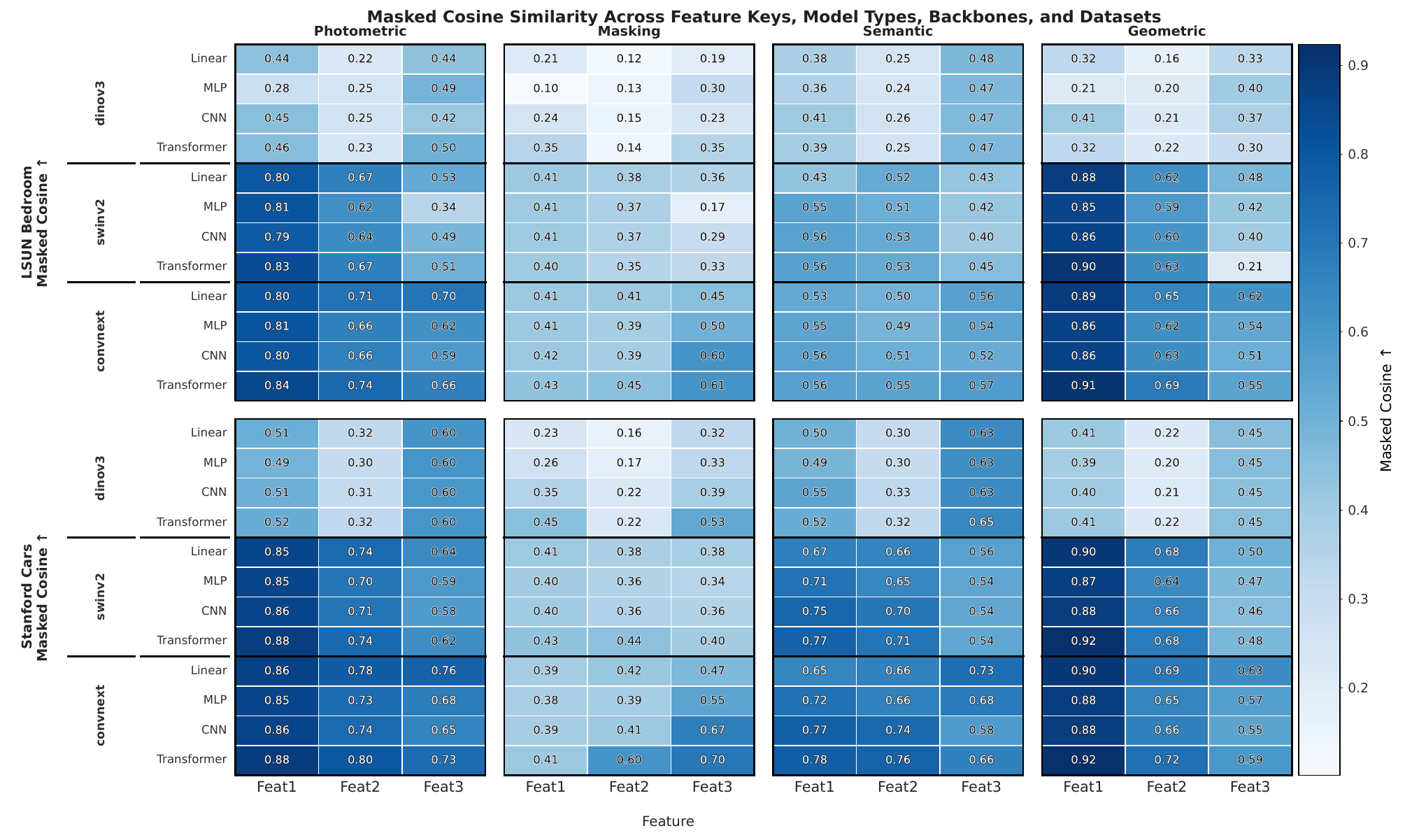}
    \caption{Heatmaps of reconstruction quality (Masked MdnCS) averaged over manipulation groups. Each row corresponds to a backbone × model-type pair, and each column within a panel corresponds to a feature layer (Feat1–Feat3). Panels are grouped by manipulation type and datase. Cell values show the mean metric for that configuration; lower LPIPS and higher masked cosine similarity indicate better alignment between reconstructed and mapped representations.}
    \label{fig:masked_cosine_shared_lsun_bedroom_stanford_cars}
\end{figure}

\begin{figure}[ht]
    \centering
   \includegraphics[width=\linewidth]{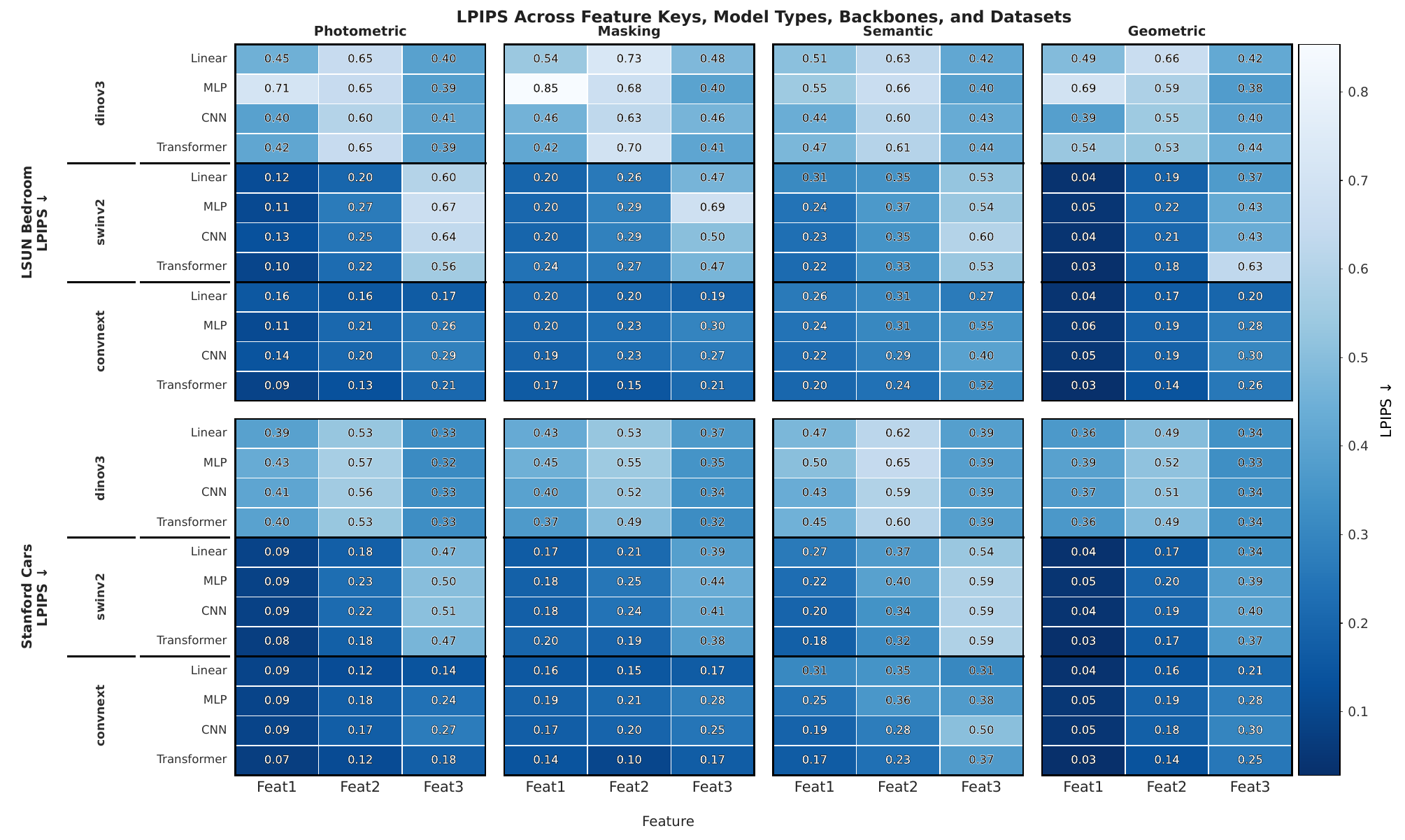}
    \caption{Heatmaps of reconstruction quality (LPIPS) averaged over manipulation groups. Each row corresponds to a backbone × model-type pair, and each column within a panel corresponds to a feature layer (Feat1–Feat3). Panels are grouped by manipulation type and datase. Cell values show the mean metric for that configuration; lower LPIPS and higher masked cosine similarity indicate better alignment between reconstructed and mapped representations.} 
    \label{fig:lpips_shared_lsun_bedroom_stanford_cars}
\end{figure}

\begin{figure}[ht]
    \centering
   \includegraphics[width=\linewidth]{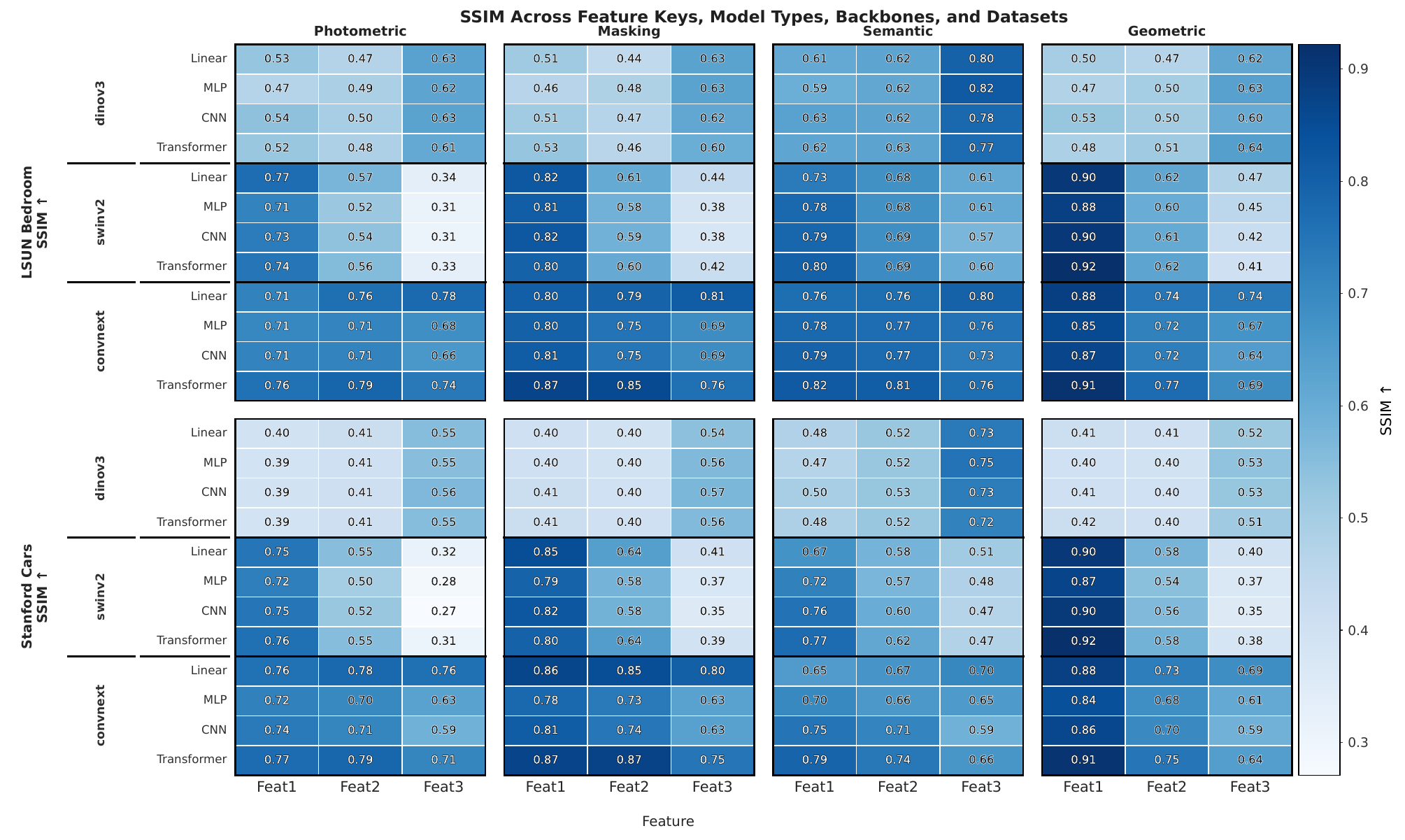}
    \caption{Heatmaps of reconstruction quality (SSIM) averaged over manipulation groups. Each row corresponds to a backbone × model-type pair, and each column within a panel corresponds to a feature layer (Feat1–Feat3). Panels are grouped by manipulation type and datase. Cell values show the mean metric for that configuration; lower LPIPS and higher masked cosine similarity indicate better alignment between reconstructed and mapped representations.} 
    \label{fig:ssim_shared_lsun_bedroom_stanford_cars}
\end{figure}

\begin{figure}[ht]
    \centering
   \includegraphics[width=\linewidth]{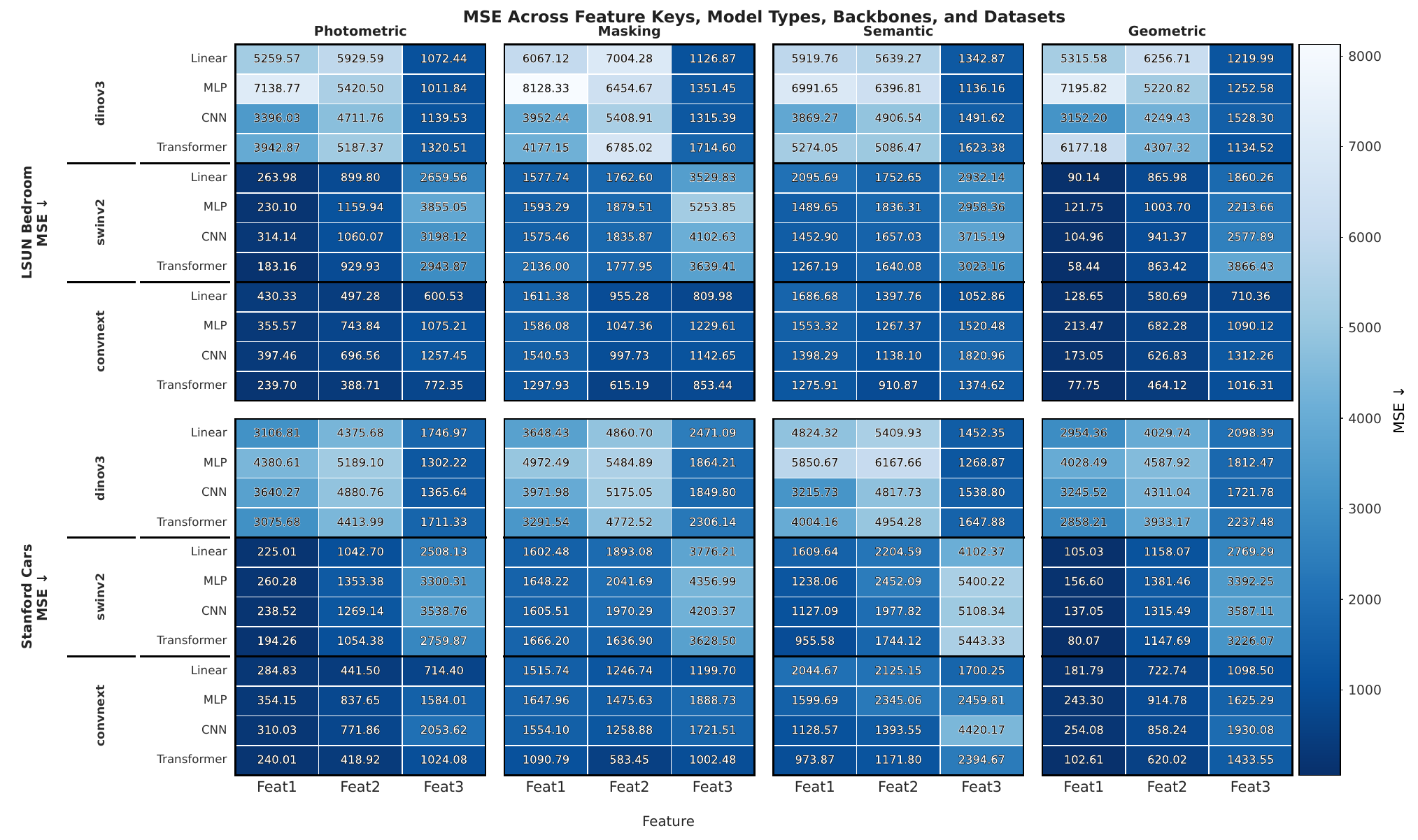}
    \caption{Heatmaps of reconstruction quality (MSE) averaged over manipulation groups. Each row corresponds to a backbone × model-type pair, and each column within a panel corresponds to a feature layer (Feat1–Feat3). Panels are grouped by manipulation type and datase. Cell values show the mean metric for that configuration; lower LPIPS and higher masked cosine similarity indicate better alignment between reconstructed and mapped representations.} 
    \label{fig:mse_shared_lsun_bedroom_stanford_cars}
\end{figure}

The corresponding results for Masked MdnCS, LPIPS, SSIM, and MSE are shown in Figs.~\ref{fig:masked_cosine_shared_lsun_bedroom_stanford_cars}, \ref{fig:lpips_shared_lsun_bedroom_stanford_cars}, \ref{fig:ssim_shared_lsun_bedroom_stanford_cars}, and \ref{fig:mse_shared_lsun_bedroom_stanford_cars}, respectively. 

\end{document}

%% file: math_commands.tex
\usepackage{amsmath,amsfonts,bm}

\def\eqref#1{equation~\ref{#1}}

\def\1{\bm{1}}

\DeclareMathAlphabet{\mathsfit}{\encodingdefault}{\sfdefault}{m}{sl}
\SetMathAlphabet{\mathsfit}{bold}{\encodingdefault}{\sfdefault}{bx}{n}

